%% file: arXiv.tex
\documentclass[10pt,twocolumn,letterpaper]{article}

\usepackage{cvpr}              

\usepackage{graphicx}
\usepackage{amsmath}
\usepackage{amssymb}
\usepackage{booktabs}
\usepackage{stfloats}
\usepackage{bbding}
\usepackage{arydshln}
\input{math.tex}

\usepackage[pagebackref,breaklinks,colorlinks]{hyperref}

\usepackage[capitalize]{cleveref}
\crefname{section}{Sec.}{Secs.}
\Crefname{section}{Section}{Sections}
\Crefname{table}{Table}{Tables}
\crefname{table}{Tab.}{Tabs.}

\def\cvprPaperID{12537} 
\def\confName{CVPR}
\def\confYear{2023}

\begin{document}

\title{Exploring Visual Explanations for Contrastive Language-Image Pre-training}

\author{Yi Li\textsuperscript{\rm1}, Hualiang Wang\textsuperscript{\rm1}, Yiqun Duan\textsuperscript{\rm2}, Xiaomeng Li\textsuperscript{\rm1} \thanks{ Corresponding author. We will release the code upon acceptance at https://github.com/xmed-lab/ECLIP.}\\
\textsuperscript{\rm1}The Hong Kong University of Science and Technology, \textsuperscript{\rm2}University of Technology Sydney\\
{\tt\small \{ylini,eexmli\}@ust.hk}
}
\maketitle

\begin{abstract}
 Contrastive Language-Image Pre-training (CLIP) learns rich representations via readily available supervision of natural language. It improves the performance of downstream vision tasks, including but not limited to the zero-shot, long tail, segmentation, retrieval, caption, and video. However, the visual explainability of CLIP is rarely studied, especially for the raw feature map. To provide visual explanations of its predictions, we propose the Image-Text Similarity Map (ITSM). Based on it, we surprisingly find that CLIP prefers the background regions than the foregrounds, and shows erroneous visualization results against human understanding. This phenomenon is universal for both vision transformers and convolutional networks, which suggests this problem is unique and not owing to certain network. Experimentally, we find the devil is in the pooling part, where inappropriate pooling methods lead to a phenomenon called semantic shift. For this problem, we propose the Explainable Contrastive Language-Image Pre-training (ECLIP), which corrects the explainability via the Masked Max Pooling. Specifically, to avoid the semantic shift, we replace the original attention pooling by max pooling to focus on the confident foreground, with guidance from free attention during training. Experiments on three datasets suggest that ECLIP greatly improves the explainability of CLIP, and beyond previous explainability methods at large margins. The code will be released later.
\end{abstract}

\section{Introduction}
\label{sec:intro}

Pre-training is ubiquitously applied in many computer vision tasks such as image classification \cite{azizi2021big}, object detection \cite{li2019analysis} and semantic segmentation \cite{bao2021beit}. 
Various weakly-supervised pre-training~\cite{mahajan2018exploring} and self-supervised pre-training\cite{doersch2015unsupervised, jaiswal2020survey} methods have been introduced to reduce the high requirements of training labels. 
Recently, the Contrastive Language-Image pre-training (CLIP)~\cite{radford2021learning}  achieved stunning results by pre-training the model on 400 million image-text pairs and learning the representations by directly matching raw text with the corresponding image.   
The success of CLIP improves the model performance on various downstream tasks, such as zero-shot and long tail classification \cite{changpinyo2021conceptual}, domain generalization \cite{cha2022domain}, segmentation \cite{xu2022groupvit,wang2022cris}, retrieval \cite{luo2021clip4clip} and video classification \cite{ni2022expanding}. 
Taking inspiration from CLIP, researchers further extend the work from several perspectives, including training efficiency \cite{zhai2022lit}, prompt design \cite{zhou2022learning} and language type \cite{gu2022wukong}.

\begin{figure}[t]
\centering
\begin{subfigure}{.133\textwidth}
  \rotatebox{90}{\ \ \ \ \ \ \ GT}\hspace{.063cm}\includegraphics[width=1.95cm,height=1.6cm]{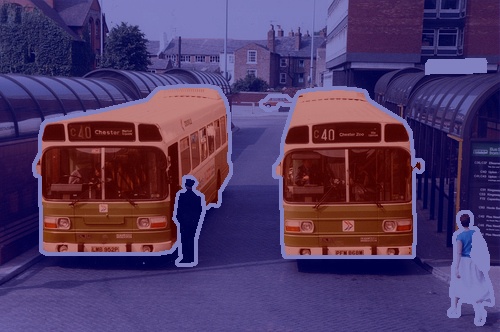} \\
  \rotatebox{90}{\ \ \ \ \ \ CLIP}\hspace{.063cm}\includegraphics[width=1.95cm,height=1.6cm]{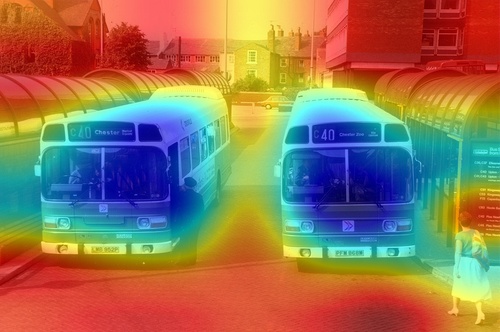} \\
  \rotatebox{90}{\ \ \ \ \ \ ECLIP}\hspace{.063cm}\includegraphics[width=1.95cm,height=1.6cm]{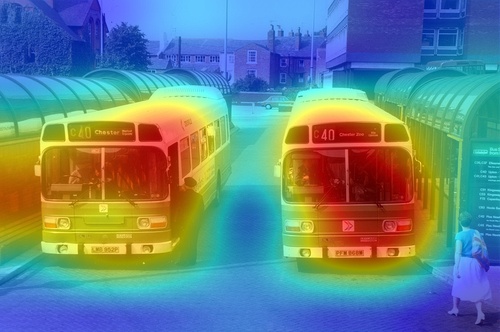}
  \caption{\centering ResNet50}
  \label{prompt_tail}
\end{subfigure}%
\begin{subfigure}{.115\textwidth}
  \includegraphics[width=1.95cm,height=1.6cm]{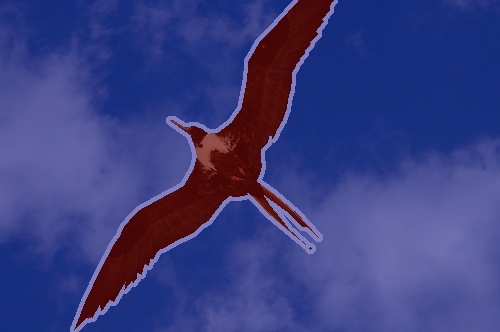} \\
  \includegraphics[width=1.95cm,height=1.6cm]{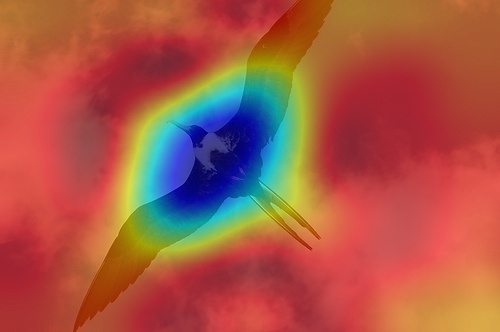} \\
  \includegraphics[width=1.95cm,height=1.6cm]{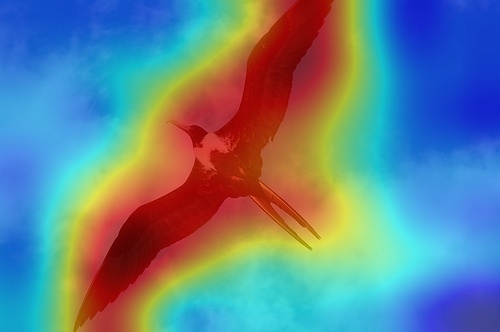}
  \caption{\centering ResNet101}
  \label{prompt_legs}
\end{subfigure}%
\begin{subfigure}{.115\textwidth}
  \includegraphics[width=1.95cm,height=1.6cm]{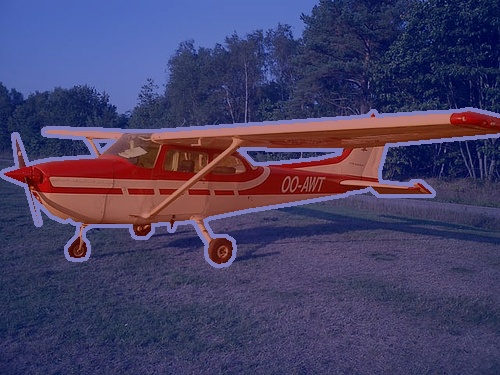} \\ \includegraphics[width=1.95cm,height=1.6cm]{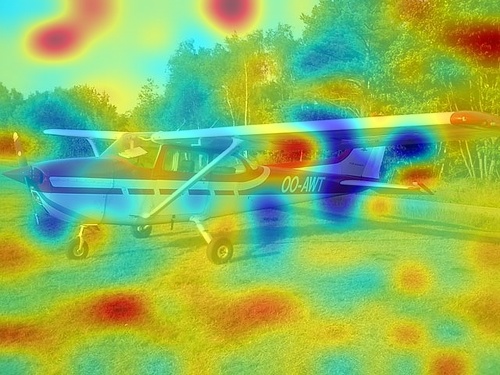} \\
  \includegraphics[width=1.95cm,height=1.6cm]{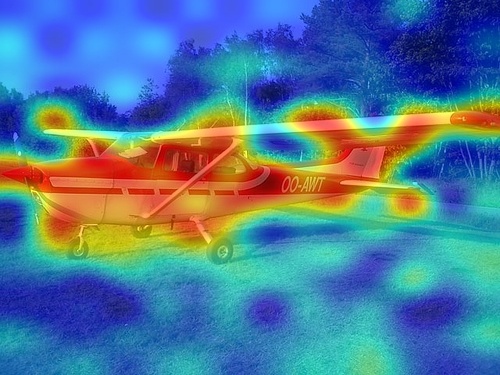}
  \caption{\centering ViT-B/16}
  \label{prompt_body}
\end{subfigure}%
\begin{subfigure}{.115\textwidth}
  \includegraphics[width=1.95cm,height=1.6cm]{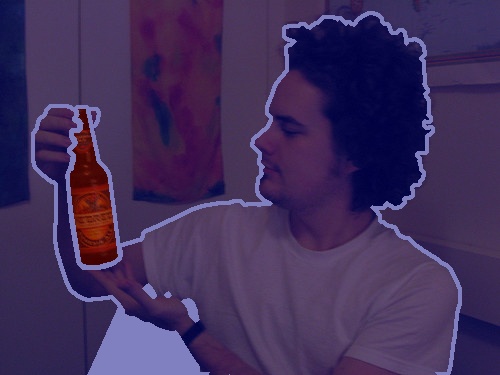} \\ \includegraphics[width=1.95cm,height=1.6cm]{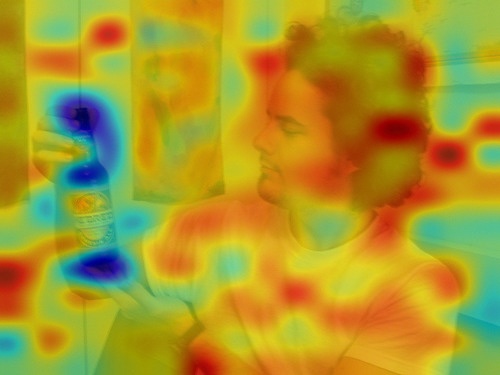} \\
  \includegraphics[width=1.95cm,height=1.6cm]{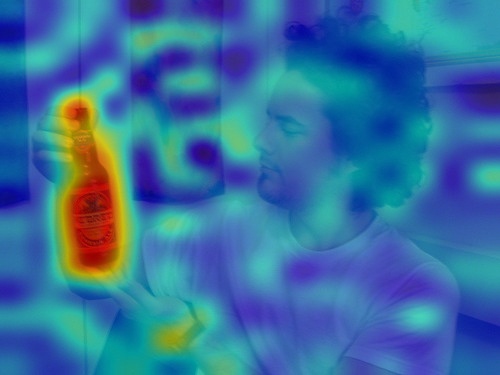}
  \caption{\centering ViT-L/14}
  \label{prompt_head}
\end{subfigure}%
\caption{Visualization of predictions via Image-Text Similarity Map in CLIP~\cite{radford2021learning} and our ECLIP with both CNN (a,b) and ViT backbones (c,d). 
Background: \textcolor{blue}{blue}; Foreground: \textcolor{red}{red}. Compared to CLIP, our ECLIP is more consistent with ground-truth (GT). }
\label{problem}
\end{figure}

Although considerable efforts have been devoted to improving CLIP or using it to enhance various downstream tasks, the explainability of CLIP has been paid 
less attention. An existing explainable method~\cite{goh2021multimodal} interprets the active neurons of CLIP via generating images related to the text. However, it doesn't locate the discriminative regions as most common explainable methods, such as CAM \cite{zhou2016learning}, Grad-CAM \cite{selvaraju2017grad}, Grad-CAM++ \cite{chattopadhay2018grad}, Score-CAM \cite{wang2020score}. 
Another explainable method~\cite{chefer2021generic} treats CLIP as vision transformer~\cite{chefer2021transformer} and relies on the self-attention of ViT~\cite{dosovitskiy2020image} to explain the images, followed by rollout~\cite{abnar2020quantifying}  to expand class-agnostic attention map into class-specific. 
However, the self-attention quality of CLIP is bad, thus leads to confused results and focuses on partial regions; see results in Fig.~\ref{vis_grad_method}. Most importantly, the raw predictions of CLIP, a natural cue for the visual explanation, is ignored by above methods.



When we visually explain the raw predictions of CLIP, via the proposed Image-Text Similarity Map (ITSM), an abnormal phenomenon occurs. Specifically, we use the similarity map between text feature and image features of image tokens to visualize the prediction map. And we surprisingly find that image features from background tokens are more close to the text feature than foregrounds, as shown in Fig. \ref{problem}.  It means CLIP prefers the background regions more, regardless of vision transformers \cite{dosovitskiy2020image} and convolutional networks \cite{he2016deep}, which is opposite to general understanding.

For the reason of this problem, we find better representation ability with high quality self-attention cannot solve this matter. Instead, the devil is in the pooling part. Specifically, we replace the original attention pooling by global average pooling and global max pooling, respectively. This problem only disappears in max pooling. And average pooling behaves like attention pooling, since it can be regarded as a weighted average pooling. We further analyze the reason and find one factor: features are shifted from discriminative locations to opposite semantic regions, owing to average-like operations. This shift makes target text features are matched to background regions instead of foregrounds, leading to the problem of erroneous visualization.

To correct the above semantic shift, we replace the original attention pooling by max pooling to focus on the confident foreground regions. Besides, to better align the text token to foreground image tokens, we apply attention map from self-supervised model as mask to guide the training, which is freely available without manual annotation. Note, this attention mask is not applied in test phase, and all the parameters are locked except an extra pair of linear projection layers. Thus, there is no sacrifice of recognition performance, with great improvements of explainability. We call the pooling as Masked Max Pooling (MMP) and name this framework as Explainable  Contrastive Language-Image Pre-training (ECLIP). Then we set rich experiments on three datasets, to measure the explainability in level of mask and score. Results suggest our ECLIP beyond original CLIP and other conventional explainability methods at large margins without any loss of recognition accuracy. Our main contributions are summarized as follows:
\begin{itemize}
\item We apply the Image-Text Similarity Map to explain the raw predictions of CLIP, and find CLIP prefers background regions than foregrounds, showing opposite visual results against human understanding, regardless of vision transformers and convolutional networks.
\item We further locate the problem at the pooling module, and point out the reason is the semantic shift, which shifts discriminative features to opposite semantic regions, owing to average-like operations.
\item The masked max pooling is proposed to correct the visualization results, also align the text tokens towarfs discriminative image tokens without manual annotation or any sacrifice of recognition performance.
\item We propose ECLIP and a simple version RCLIP, which surpass the original CLIP and other methods at large margins (e.g. mean improvements of mIoU at 27.26\% on three datasets for CLIP ResNet101.).
\end{itemize}

\section{Related Work}
\subsection{Contrastive Language-Image pre-training}
Extensive works have been developed on the interaction of computer vision and natural language processing, such as text-to-image retrieval~\cite{wang2019camp}, visual question answering~\cite{antol2015vqa} and referring segmentation~\cite{wang2022cris}.
Recently, contrastive language-image pre-training (CLIP)~\cite{radford2021learning} has gained substantial attention due to its impressive performance and superior transfer ability over diverse classification datasets. 
The following researchers either further extend CLIP from different perspectives~\cite{zhou2022learning,gao2021clip,zhai2022lit} or use it to enhance the performance of various downstream tasks~\cite{xu2022groupvit,ni2022expanding,cha2022domain,changpinyo2021conceptual}. 
For example, CoOp~\cite{zhou2022learning} and CLIP-Adapter~\cite{gao2021clip} improved the performance by fine-tuning CLIP with adapter or prompt optimization. 
LiT~\cite{zhai2022lit} designed a new pre-training strategy that only tuned the text model CLIP using image-text pairs and froze the image model. 
Lei~\etal~\cite{lei2021less} and Luo~\etal~\cite{luo2021clip4clip} showed that the CLIP model can contribute to the image retrieval task.
Ma~\etal~\cite{ma2021simple} leveraged language modality via CLIP backbone to improve the long-tailed recognition task. Although stunning results have been achieved by CLIP, its visual explainability is still underexplored.   


\subsection{Dense Predictions of CLIP}
Visual explainability of CLIP is related to dense prediction tasks of CLIP in some extent. For example, Rao~\etal~\cite{Rao_2022_CVPR} extend image-text relationships to pixel-text relationships, then fine-tune the pretraining model on detection and segmentation tasks. But this work doesn't focus on the behaviors of CLIP itself. Instead, it fine-tunes down stream tasks with fully supervision. Besides, Zhong~\etal~\cite{zhong2022regionclip} match regions to text with dense predictions, but the localization ability is borrowed from a fully-supervised region proposal network, instead of explainability from the raw predictions of CLIP. Similarly, Xu~\etal~\cite{xu2021simple} and Li~\etal~\cite{li2022language} use manual mask to locate the foregrounds for semantic segmentation. And it utilizes the recognition ability of CLIP, instead of explainability. Besides, Xu~\etal~\cite{xu2022groupvit} use language supervision only for segmentation. However, they abandon the network architecture of the original CLIP, instead,  use a grouping module to merge tokens, which is quite different from the visual explanation task.

\subsection{Explainability for Deep Learning Models}
Most existing methods for visual explainability are designed for convolutional networks~\cite{zeiler2014visualizing,zhou2016learning} or vision transformers~\cite{chefer2021generic,chefer2021transformer}. Specifically, the prior work, CAM~\cite{zhou2016learning}, locates the discriminative regions of CNN to explain the model. Inspired by this work, some researchers explored different methods (\eg, Grad-CAM~\cite{selvaraju2017grad} and Score-CAM~\cite{wang2020score}) to more precisely and effectively generate CAM to reveal visual cues distributed on images. Compared with CNN-based methods, self-attention is an additional visualization target in visual transformers~\cite{dosovitskiy2020image}. Abnar~\etal~\cite{abnar2020quantifying} presented a rollout method and attention flow method to consider the path and max-flow problems along the pairwise attention graph. Then, Chefer~\etal~\cite{chefer2021transformer} expand the class-agnostic self-attention of rollout to class-specific.

As for visual explainability in CLIP, Goh~\etal~\cite{goh2021multimodal} visualizes the activated neurons of CLIP by image generation, while the location of discriminative regions are not provided. Besides, Bi-Model~\cite{chefer2021generic} based on~\cite{chefer2021transformer} treats CLIP as ViT \cite{dosovitskiy2020image} and explains it with self-attention, even the quality of self-attention in CLIP is bad (see Fig. \ref{solution_attn_clip}). Specifically, it uses the similarity scores of CLIP as ``logit'' and interprets CLIP with gradient and self-attention. Note, this ``logit'' is also applicable to Grad-CAM \cite{selvaraju2017grad}. But experiments in Fig. \ref{vis_grad_method} suggest Grad-CAM shows opposite visualization as CLIP, and Bi-Module is confused at multiple classes, focusing on partial regions. Besides the problems of performance, the raw predictions of CLIP, a natural cue for the visual explanation, is ignored by above methods, even it's more straight forward without backpropagation multiple times for each image as gradient-based methods.


\section{Method}

\subsection{Image-Text Similarity Map}
Foremost, we define the involved components of CLIP as below symbols: image encoder $f_i$, text encoder $f_t$, and corresponding pair of linear projections $\phi_i,\phi_t$ (learnable 2-D parameter matrix).
Then we have text features $\mF_t \in \R^{N_t \times C}$ from text input $x_t$ as:
\begin{equation}
    \label{text_porj}
    f_t(x_t) = \mF_t,
\end{equation}
and image features: 
\begin{equation}
\label{img_porj}
f_i(x_i) = \{\mF_c, \mF_i\},
\end{equation}
from image input $x_i$, where $\mF_c \in \R^{1 \times C}$ is the class token for classification, and rest image tokens $\mF_i \in \R^{N_i \times C}$ are the raw feature map ($N_i$ and $N_t$ indicates number of text tokens and image tokens, respectively, C is the embedding size). In CLIP, $\mF_c$ and $\mF_i$ are both belong to the final predictions of $f_i$, but unlike CLIP which skips the image features $\mF_i$, we need it for prediction explanation. 

The acquisition of Image-Text Similarity Map (ITSM) is similar to the generation of confidence score $s \in \R^{N_t}$ from class feature $\mF_c$, text features $\mF_t$ and corresponding linear projections $\phi_i$, $\phi_t$:
\begin{equation}
    s = (\frac{\mF_c \cdot \phi_i}{|| \mF_c \cdot \phi_i ||_2}) \cdot (\frac{\mF_t \cdot \phi_i}{|| \mF_t \cdot \phi_i ||_2})^\top.
\end{equation}
The difference is that we replace the class feature $F_c \in \R^{1 \times C}$ to features of image tokens $\mF_i \in \R^{N_i \times C}$, thus we have intermediate ITSM $\hat{\mM} \in \R^{N_i \times N_t}$ as:
\begin{equation}
    \hat{\mM} = (\frac{\mF_i \cdot \phi_i}{|| \mF_i \cdot \phi_i ||_2}) \cdot (\frac{\mF_t \cdot \phi_i}{|| \mF_t \cdot \phi_i ||_2})^\top,
\end{equation}
Then we reshape $\hat{\mM}$ to final ITSM $\tM \in \R^{H \times W \times N_t}$ with resize operation and min-max normalization:
\begin{equation}
\label{itsm}
    \tM = norm(resize(reshape(\hat{\mM}))),
\end{equation}
where H and W indicates the height and width of original image, and $N_i = \frac{H}{scale_H} \cdot \frac{W}{scale_W}$ ($scale_H$ and $scale_W$ are the resize scales).

After the generation of ITSM over varied backbones of CLIP, we find the behavior of CLIP's ITSM is erroneous and against human understanding, as shown in Fig. \ref{problem}. 

\subsection{The Devil Is in the Pooling Module}

\begin{figure}[ht]
\centering
\begin{subfigure}{.097\textwidth}
  \includegraphics[width=.95\linewidth]{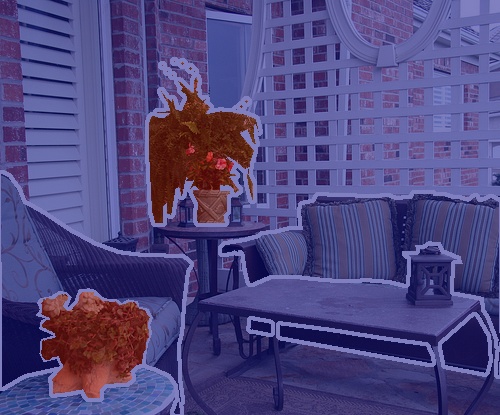} \\
  \includegraphics[width=.95\linewidth]{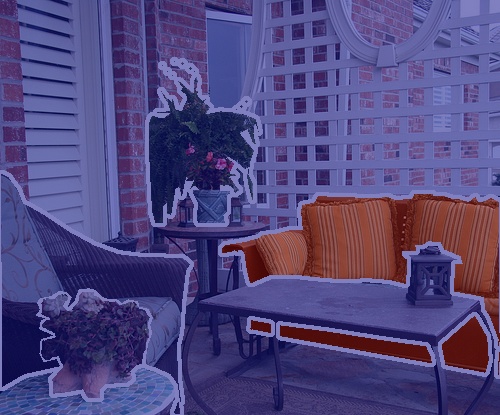}
  \caption{Ground\\Truth\centering}
  \label{solution_gt}
\end{subfigure}%
\begin{subfigure}{.097\textwidth}
  \includegraphics[width=.95\linewidth]{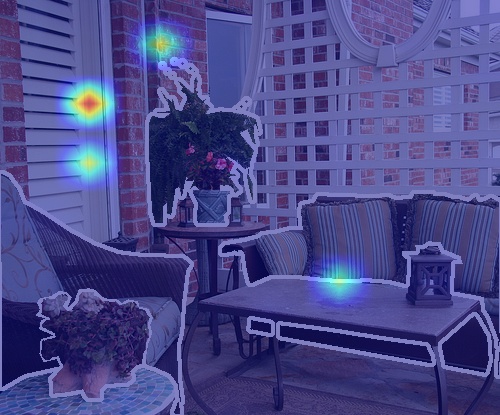} \\
  \includegraphics[width=.95\linewidth]{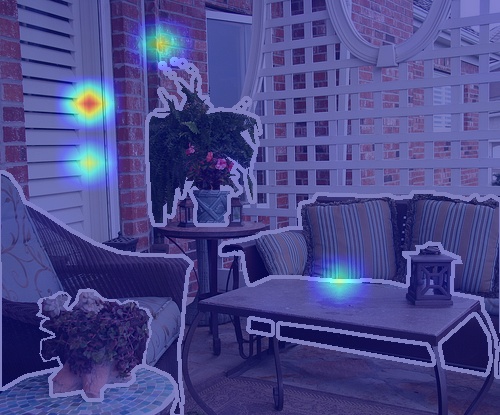}
  \caption{Attention \\ CLIP \centering}
  \label{solution_attn_clip}
\end{subfigure}%
\begin{subfigure}{.097\textwidth}
  \includegraphics[width=.95\linewidth]{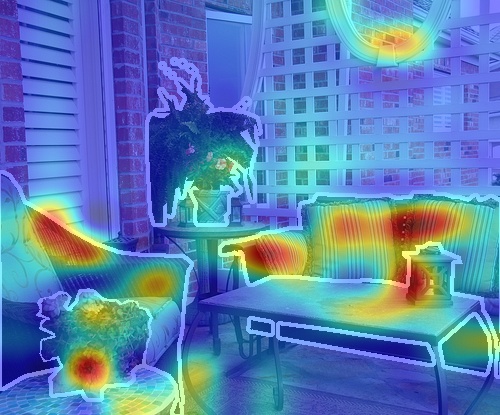} \\
  \includegraphics[width=.95\linewidth]{./figs/solution/2009_001775_potted_plant_dino_attn.jpg}
  \caption{Attention \\ DINO \centering}
  \label{solution_attn_dino}
\end{subfigure}%
\begin{subfigure}{.097\textwidth}
  \includegraphics[width=.95\linewidth]{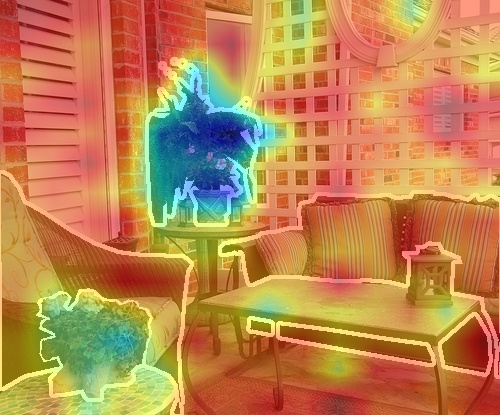} \\
  \includegraphics[width=.95\linewidth]{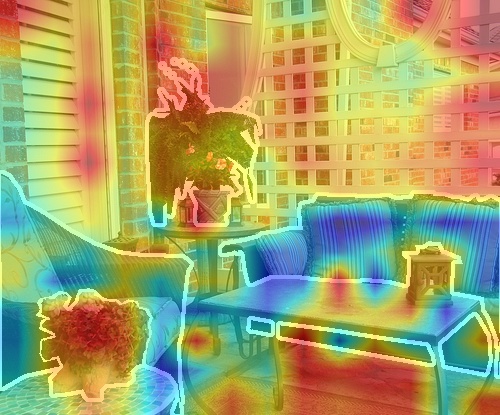}
  \caption{ITSM \\ with CLIP \centering}
  \label{solution_token}
\end{subfigure}%
\begin{subfigure}{.097\textwidth}
  \includegraphics[width=.95\linewidth]{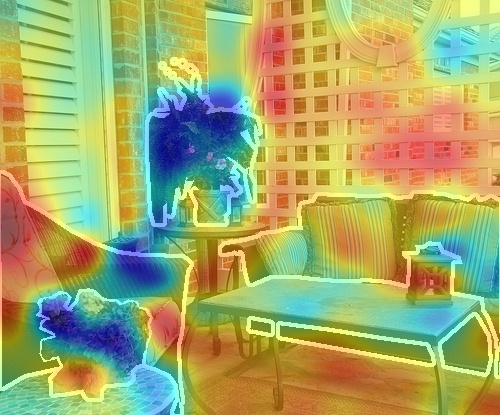} \\
  \includegraphics[width=.95\linewidth]{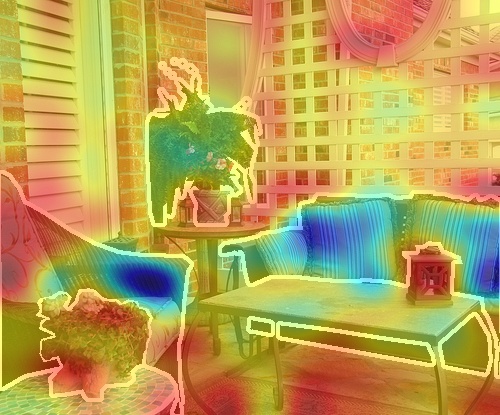}
  \caption{ITSM \\ with DINO \centering}
  \label{solution_token_dino}
\end{subfigure}%

\caption{Representation is not the factor. Attention indicates the self-attention of the last transformer layer. And DINO means the weights of image encoder are initialized by DINO \cite{caron2021emerging} (all other experiments use the locked weights of CLIP). The localization ability of backbone is improved (c vs. b), but the explainability of prediction map (ITSM) is still opposite to ground truth (e vs. d).}
\label{solution}
\end{figure}

\textbf{Representation ability is not the key factor.} This first influential factor is the representation ability of the image encoder. For the image encoder, we experimentally replace it by the self-supervised image encoder, DINO \cite{caron2021emerging}, to introduce better localization ability. As Shown in Fig. \ref{solution_attn_dino}, the attention quality is much better than that of CLIP (Fg. \ref{solution_attn_clip}), when the self-supervised image encoder is applied. However, its ITSM still focuses on background as Fig. \ref{solution_token_dino}. So, it shows the image encoder is not the key of erroneous results. 

\begin{figure}[h]
\centering
\begin{subfigure}{.12\textwidth}
  \includegraphics[width=.95\linewidth]{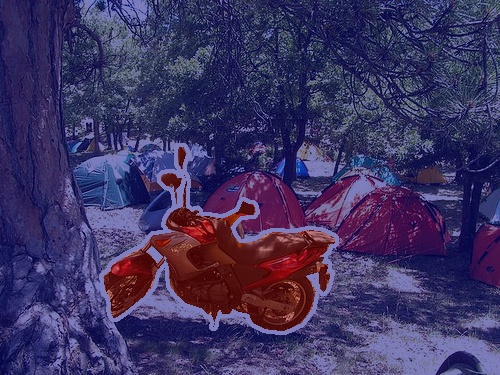} \\
  \includegraphics[width=.95\linewidth]{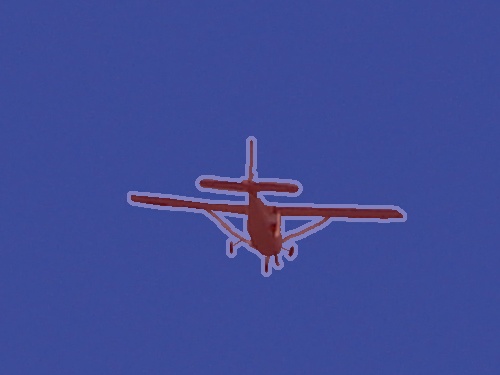}
  \caption{Ground\\Truth\centering}
\end{subfigure}%
\begin{subfigure}{.12\textwidth}
  \includegraphics[width=.95\linewidth]{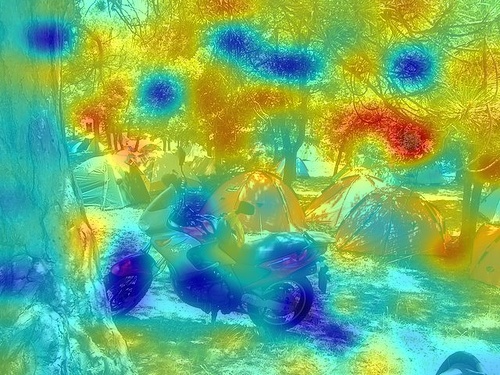} \\
  \includegraphics[width=.95\linewidth]{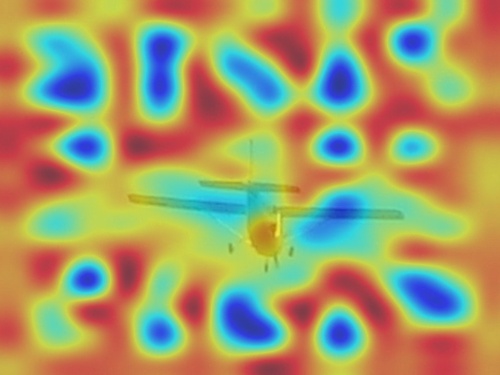}
  \caption{Attention \\Pooling\centering}
\end{subfigure}%
\begin{subfigure}{.12\textwidth}
  \includegraphics[width=.95\linewidth]{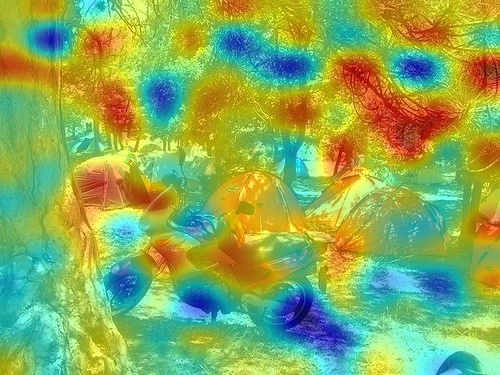} \\
  \includegraphics[width=.95\linewidth]{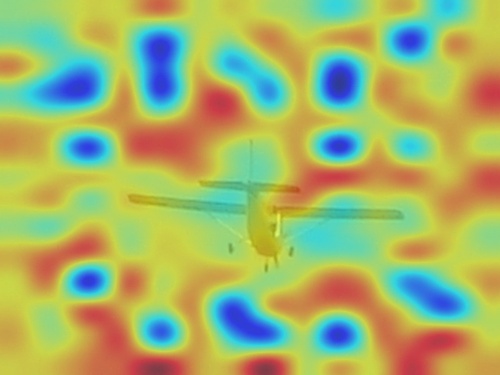}
  \caption{Average \\ Pooling\centering}
\end{subfigure}%
\begin{subfigure}{.12\textwidth}
  \includegraphics[width=.95\linewidth]{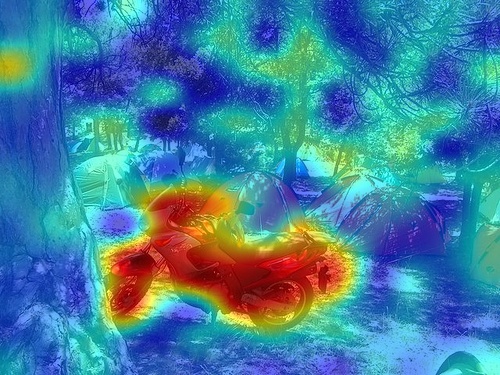} \\
  \includegraphics[width=.95\linewidth]{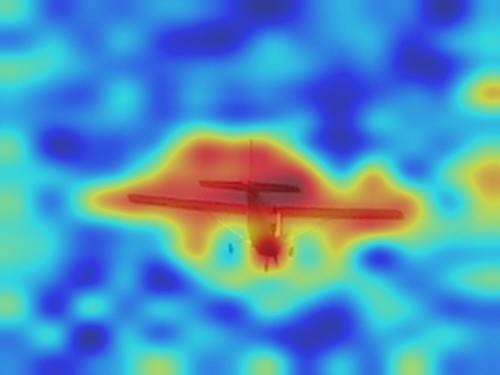}
  \caption{Max \\Pooling\centering}
\end{subfigure}%

\caption{Visualization results shown that max pooling solve the problem. All the models are trained at the same settings on CLIP ViT-B/16. Predicted foreground is color in red.}
\label{pooling}
\end{figure}

\textbf{Max pooling solve the problem}. As Fig. \ref{pooling}, we focus on the second factor: pooling methods. We lock the parameters of CLIP, and add a new pair of linear projections, with different pooling methods to pool the image features $\mF_i$. And all the models use the same training settings. When we replace the original attention pooling layer to global average pooling, the average pooling is similar to attention pooling. \emph{While the max pooling solves the problem, which suggests the devil is in the pooling module}.

\begin{figure}[h]
\centering
\begin{subfigure}{.122\textwidth}
  \includegraphics[width=.95\linewidth]{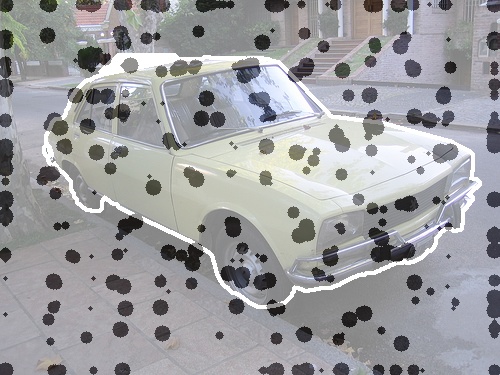} \\ \includegraphics[width=.95\linewidth]{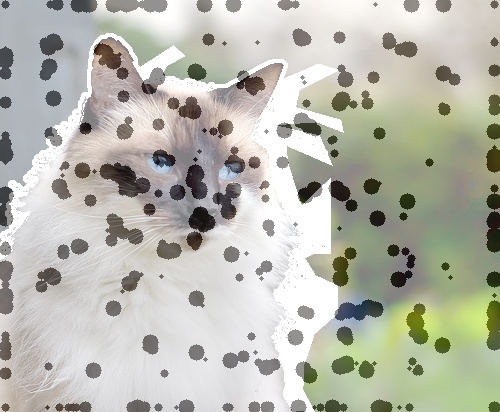}
  \caption{\centering MaxPool}
  \label{reason_max}
\end{subfigure}%
\begin{subfigure}{.122\textwidth}
  \includegraphics[width=.95\linewidth]{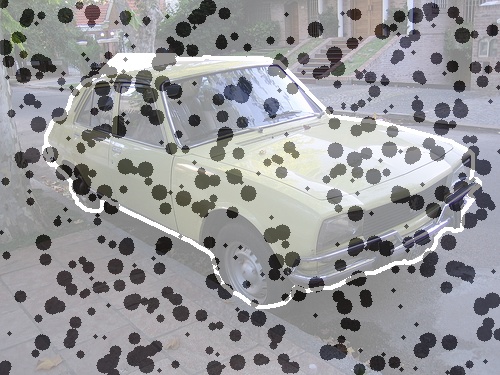} \\ \includegraphics[width=.95\linewidth]{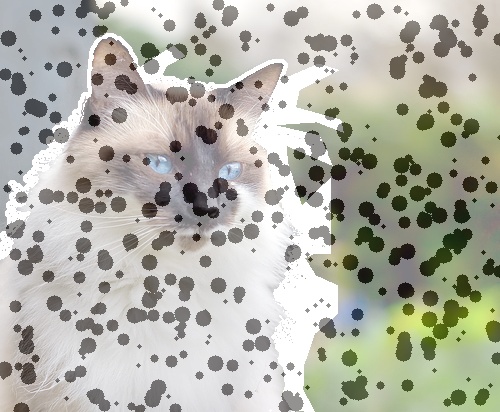}
  \caption{\centering AvgPool}
  \label{reason_avg}
\end{subfigure}%
\begin{subfigure}{.122\textwidth}
  \includegraphics[width=.95\linewidth]{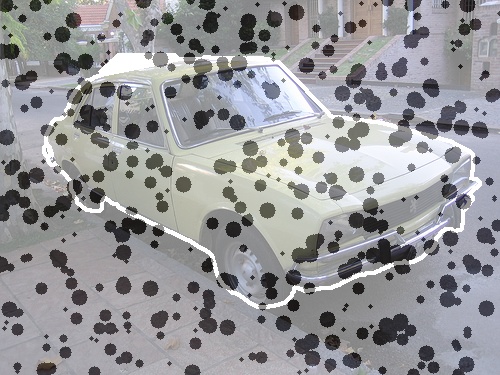} \\ \includegraphics[width=.95\linewidth]{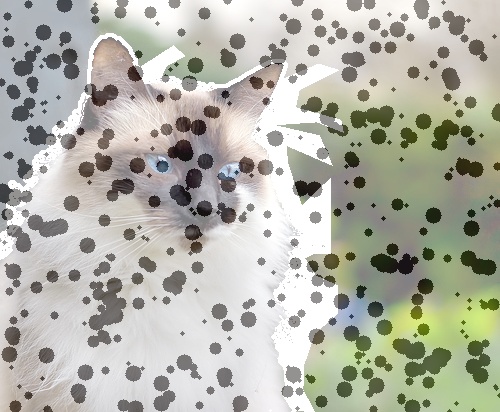}
  \caption{\centering AttnPool}
  \label{reason_attn}
\end{subfigure}%
\begin{subfigure}{.122\textwidth}
  \includegraphics[width=.95\linewidth]{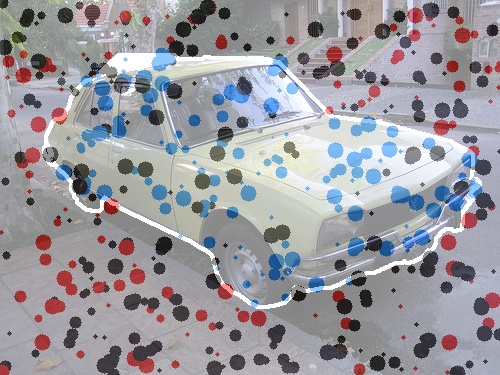} \\ \includegraphics[width=.95\linewidth]{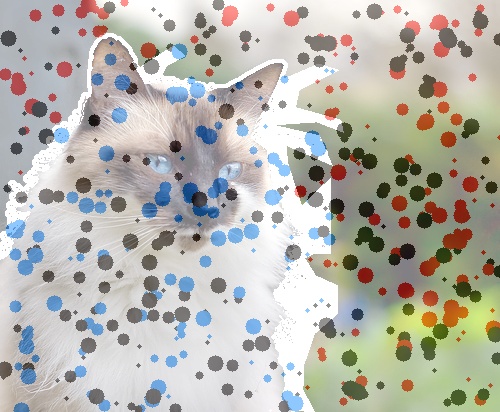}
  \caption{\centering Avg vs. Max}
  \label{reason_shift}
\end{subfigure}%
\caption{Illustration of feature shift (a vs. b\&c) and semantic shift (d) on CLIP ViT-B/16. we draw one point for each channel on the image. And the location of a point is same to the pixel, whose score is most close to the pooled score. Note larger points means higher pooled value. Since the number of points are fixed, points on max pooling (the best localization ability) presents overlapped status with the least area of points. While we can see more points in (b, c), which means the feature are shifted from discriminative locations. This is feature shift. For (d), Blue points indicate features shifted from background to foreground, and red points are opposite. (d) shows feature shift leads points to opposite semantic regions, and explain how erroneous visualization happens.} 
\label{reason}
\end{figure}

\textbf{Reason analyze.} After locating the problem at the pooling module. We analyze the reason by element-wise feature comparison among different pooling methods. Specifically, we draw a point for each channel on the image. And the location of a point is the same as the pixel on this channel map, whose score is most close to the pooled score. As the number of points are fixed, and we can see the points in Fig. \ref {reason_max} (from max pooling) is less than others, which indicates the points are highly aggregated and overlapped. Since max pooling has the best localization ability as Fig. \ref{pooling}, we compare it with average pooling (Fig. \ref {reason_avg}) and attention pooling (Fig. \ref {reason_attn}). We find average pooling and attention pooling disperse the points from overlapped status. We call this phenomenon as \emph{feature shift}. And feature shift between foregrounds and backgrounds is the \emph{semantic shift} as Fig. \ref{reason_shift}. This feature shift leads foregrounds are matched to backgrounds, and background features are turned to foreground regions. Exactly, it behaves as the problem of opposite visualization results, and explain how it happens. Besides, why CLIP is more sensitive than single modality model about pooling method is waiting to explore.

\subsection{Masked Max Pooling}

As analyzed above, the first principle to improve the explainability of CLIP is to avoid semantic shift owing to average-like pooling. Another motivation is to emphasize features of foregrounds to boost the alignment. As shown in Fig. \ref{solution_attn_clip}, the attention map of self-supervised model is better than the original CLIP. While, this attention map is class-agnostic. We aim to correct the explainability of CLIP with it, to get high quality class-aware visualization maps. In this paper, we propose the Masked Max Pooling (MMP) to correct the semantic shift, as well as emphasize the discriminative features.

To remain original recognition ability, we lock all the parameters of CLIP, and correct the explainability with an extra pair of linear projection layers, $\hat{\phi_i}$ and $\hat{\phi_t}$ for image encoder and text encoder, respectively. Besides, we introduce the free available self-attention mask from another offline self-supervised model (e.g. DINO \cite{caron2021emerging}, MoCo \cite{chen2021empirical}). Specifically, the multi-head attention mask $\hat{\mA} \in \R^{N_h,N_i}$ comes from the last transformer layer, where $N_h$ is the number of heads and $N_i$ is the size of image tokens. Before deploying it, we reduce the head dimension by mean operation and expand to the shape of as image features $F_i\in\R^{N_i\times  C}$. Here we have the expanded mean attention map $\mA\in\R^{N_i\times C}$:
\begin{equation}
    \mA = expand(mean(\hat{\mA}))
\end{equation}

Then we apply the offline mean attention map as mask to guide the alignment between text features and salient image tokens. We achieve this step by element-wise production $\odot$ towards image features as:
\begin{equation}
    \label{correct}
    \mF_{mc} = MMP(\mF_i)= max(\mF_i \odot \mA),
\end{equation}
where $\mF_{mc}$ indicates masked class token, pooled by global max pooling along dimension $N_i$. And $\hat{f_i}$ means the locked image encoder with masked max pooling.

Then back to Eq. \ref{img_porj}, we update it as:
\begin{equation}
    \hat{f_i}(x_i) = \{\mF_{mc}, \mF_i\},
\end{equation}
Where MMP is included in $\hat{f_i}$ with new image projection layer $\hat{\phi_i}$, and MMP doesn't influence the output image features $\mF_i$ ($\mA$ is not applied to $\mF_i$). Specifically, this attention mask is only used to guide the training and not applied in test phase.

Then we optimize the new pair of linear projection layers by the contrastive loss $\mathcal{L}$ as below:
\begin{equation}
    \mathcal{L}(x_i, x_t) =  \mathcal{L}(\frac{\mF_{mc} \cdot \hat{\phi_i}}{|| \mF_{mc} \cdot \hat{\phi_i} ||_2}, \  \frac{\mF_t \cdot \hat{\phi_i}}{|| \mF_t \cdot \hat{\phi_i} ||_2}).
\end{equation}

To conclude, MMP is designed to learn $\hat{\phi_i}$ and $\hat{\phi_t}$ for reasonable localization results, without interference to the recognition performance.

\subsection{ECLIP and RCLIP}

\begin{figure*}[t]
\centering
\includegraphics[width=0.9\textwidth]{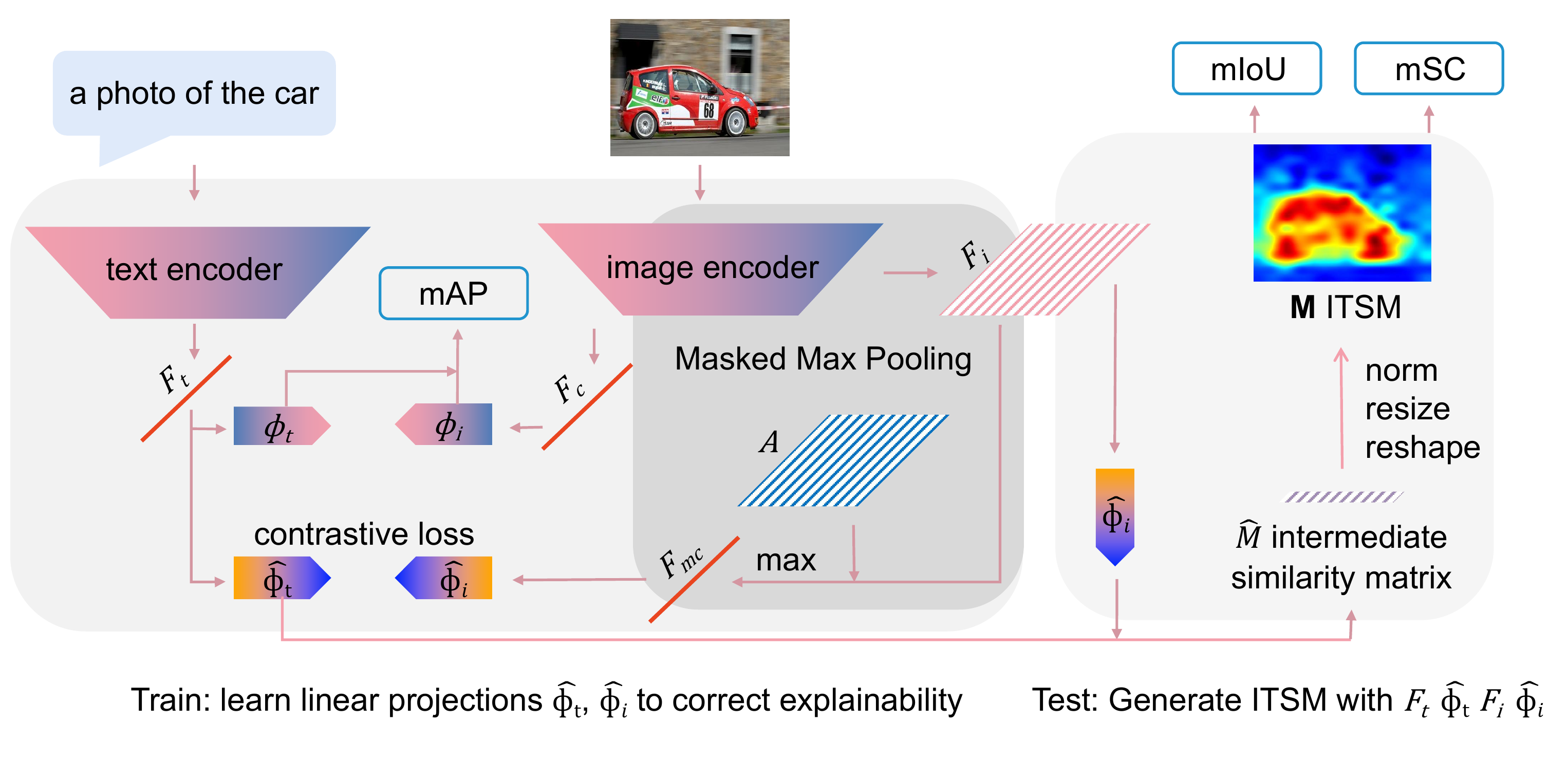}

\caption{Illustration of ECLIP for single image-text pair. \textbf{Middle}: the image encoder return features of class token $\mF_c$, image tokens $\mF_i$. The masked class token $\mF_{mc}$ comes from $\mF_i$ with offline attention map $\mA$ (class agnostic). \textbf{Left}: Only new linear projections $\phi_i,\hat{\phi_i}$ are optimized by contrastive loss, and other parameters are locked without interference to recognition performance. \textbf{Right}: The generation of ITSM, the image features $\mF_i$ are projected by $\hat{\phi_i}$ to get intermediate similarity matrix $\hat{\mM}$ with $\mF_i,\ \hat{\phi_t}$. After reshape, resize and min-max normalization, the ITSM is evaluated by mIoU and mSC. Recognition task is evaluated by mAP with original weights.}
\label{framework}
\end{figure*}

As shown in Fig. \ref{framework}, Explainable CLIP (ECLIP) is consisted of three parts. The left part indicates a new pair of linear projections are used to correct the explainability, and the recognition task is evaluated with original parameters without optimization. Because of limited computational resources, we just correct its explainability with a small dataset compared to the original CLIP. This design allows the original recognition ability is not damaged under a small dataset. The middle part is the proposed masked max pooling. Specifically, max pooling locates the confident foregrounds, and offline attention mask guides the alignment during training only as Eq. \ref{correct}. Besides, we draw the ITSM with $\hat{\phi_i}$ and $\hat{\phi_t}$ as the right part. This part is described from Eq. \ref{text_porj} to Eq. \ref{itsm}.

Besides ECLIP, we also propose a simple version, Reversed CLIP (RCLIP), to correct the explainability without training. Our motivation is to \textbf{make good use of the mistake}. As shown in Fig. \ref{problem}, CLIP shows opposite visualization results and prefers the background more. Analyzed in Fig. \ref{reason_shift}, this problem is owing to semantic shift. So, one simple idea is to make the best of a mistake by reversing the ITSM to get predictions $\tM_r$ of RCLIP as:
\begin{equation}
    \tM_r = Abs(1-\tM),
\label{reverse}
\end{equation}
where $Abs$ gets the absolute value for $1 - \tM$.

\section{Experiments}
\subsection{Experimental Setup}
\textbf{Dataset.} The original CLIP \cite{radford2021learning} uses 400 millions data. However, this dataset is not available, also too large to reproduce. \textbf{Because of limited computational resources}, we use the dataset of Google Conceptual Captions 3 millions (GCC3M) \cite{sharma2018conceptual} to train the model for the moderate quantity. For the evaluation, we use three datasets with pixel-level annotations, which support the measurement of localization ability well. Specifically, we use the validation sets of Pascal VOC 2012 \cite{everingham2010pascal}, MS COCO 2017 \cite{lin2014microsoft} and ImageNet-Segmentation-50 (ImageNet-S50) \cite{gao2022large}, whose quantities are 1449, 5000 and 752, respectively. Note that there are 80 foreground categories in COCO, and it's more complex and difficult than VOC whose class number is 20. ImageNet-S50 has 50 foregrounds, while it's single label dataset, thus it's simpler than above two datasets.

\begin{table*}[hb]
\centering
\setlength\tabcolsep{7.5pt}
\begin{tabular}{cccccccccc}
\hline 
  &  & \multicolumn{2}{c}{VOC 2012} &\multicolumn{2}{c}{COCO 2017} & \multicolumn{2}{c}{ImageNet-S50} & \multicolumn{2}{c}{\textbf{Average}}\\
Method & Network & mIoU $\uparrow$ & mSC $\uparrow$ & mIoU $\uparrow$ & mSC $\uparrow$ & mIoU $\uparrow$ & mSC $\uparrow$ & mIoU $\uparrow$ & mSC $\uparrow$ \\
\hline
CLIP & ResNet50 & 17.72 & -27.07 & 10.50 & -19.21 & 28.15 & -25.46 & 18.79 & -23.91\\
\textbf{ECLIP} & ResNet50 & 49.30 & 29.16 & 26.98 & 22.78 & 61.07 & 28.64 & $\textbf{45.78}_{+\textbf{26.99}}$ & $\textbf{26.86}_{+\textbf{50.77}}$\\
\hdashline[0.5pt/5pt]
CLIP & ResNet101 & 17.99 & -23.95 & 10.63 & -18.05 & 28.10 & -23.13 & 18.91 & -21.71 \\
\textbf{ECLIP} & ResNet101 & 49.33 & 28.93 & 26.95 & 22.39 & 62.22 & 30.01 & $\textbf{46.17}_{+\textbf{27.26}}$ & $\textbf{27.11}_{+\textbf{48.82}}$\\
\hdashline[0.5pt/5pt]
CLIP & ViT-B/32 & 17.52 & -25.01 & 10.14 & -22.54 & 27.89 & -22.83 & 18.52 & -23.46\\
\textbf{ECLIP} & ViT-B/32 & 48.39 & 34.74 & 25.78 & 25.00 & 62.37 & 35.94 & $\textbf{45.51}_{+\textbf{26.99}}$ & $\textbf{31.89}_{+\textbf{55.35}}$\\
\hdashline[0.5pt/5pt]
CLIP & ViT-B/16 & 17.46 & -18.55 & 9.80 & -22.30 & 27.94 & -16.54 & 18.40 & -19.13\\
\textbf{ECLIP} & ViT-B/16 & 45.22 & 26.16 & 23.39 & 17.77 & 54.94 & 23.75 & $\textbf{41.18}_{+\textbf{22.78}}$ & $\textbf{22.56}_{+\textbf{41.69}}$\\
\hdashline[0.5pt/5pt]
CLIP & ViT-L/14 & 17.26 & -21.59 & 9.67 & -26.46 & 27.99 & -16.35 & 18.31 & -21.47 \\
\textbf{ECLIP} & ViT-L/14 & 46.11 & 26.03 & 27.20 & 22.45 & 59.69 & 27.24 & $\textbf{44.33}_{+\textbf{26.02}}$ & $\textbf{25.24}_{+\textbf{46.71}}$\\
\hline 
\end{tabular}

\caption{\label{res_all_net}Results of three datasets for varied networks, with their average results. mIoU (\%) measures the mask-level explainability, and mSC (\%) ranges from -100\% to 100\%, reflecting the score contrast between foreground and background (lower than 0 indicates the model prefer the background more). Our ECLIP surpasses original CLIP at large margins.}
\end{table*}

\begin{figure}[h]
\centering
\begin{subfigure}{.16\textwidth}
\centering
  \includegraphics[width=2.7cm,height=1.8cm]{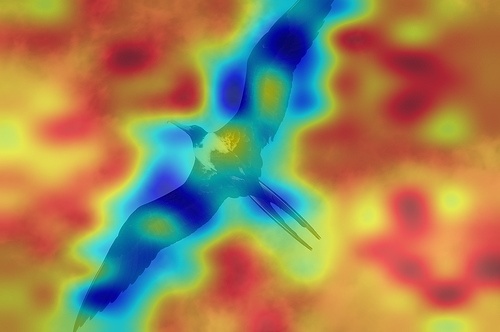}
  \caption{mSC \textless 0\centering}
\end{subfigure}%
\begin{subfigure}{.16\textwidth}
\centering
  \includegraphics[width=2.7cm,height=1.8cm]{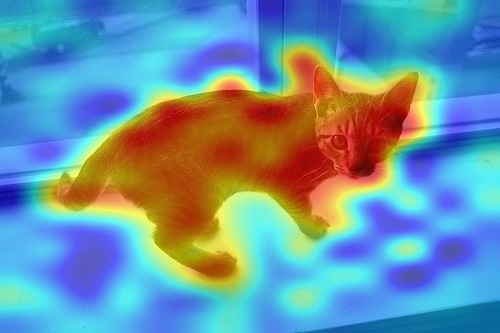}
  \caption{high mSC\centering}
\end{subfigure}%
\begin{subfigure}{.16\textwidth}
\centering
  \includegraphics[width=2.7cm,height=1.8cm]{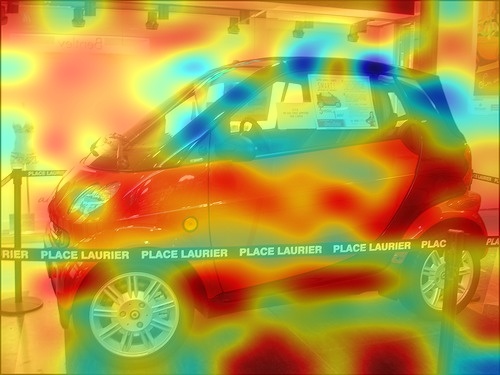}
  \caption{low mSC\centering}
\end{subfigure}%

\caption{New insights from mSC, mSC are able to measure opposite visualization results (a), while mIoU cannot. And mSC reflects the color contrast between foregrounds and backgrounds (b vs. c). Predicted foreground is colored in red.}
\label{msc}
\end{figure}

\textbf{Evaluation metric.} As the evaluated datasets have multiple label of single image, we use \textbf{mAP} to measure the recognition performance, which is widely used in multi-label classification. For explainability, we evaluate it with the pixel-level annotations by \textbf{mIoU} (mean Intersection of Union). Since the threshold for foreground mask influences mIoU a lot, to avoid the interference of manual threshold, we apply grid search to ITSM $\tM$ for each image at step 0.01 to match the appropriate foreground mask. Thus, each image for all models are fairly compared without influence of thresholds. Besides, we evaluate the existed foreground categories to avoid the interference of varied classification performance. We also propose a metric \textbf{mSC} (\textbf{mean Score Contrast}, range from 100\% to -100\% for the ITSM) to measure the explainability in the level of score contrast. Specifically, we use the \emph{mean score of foregrounds to minus the mean score of backgrounds}, then count the mean score over categories. This metric is necessary in two aspects as Shown in Fig. \ref{msc}: (1) mSC can measure the phenomenon of opposite visualization, when the score is lower than 0, while mIoU cannot. (2) If the mSC is high, the color contrast between foreground and background is obvious, which presents better visualization results than low score contrast.

\textbf{Settings.} For the CLIP, we use the official models trained from 400 millions private data, without fine-tuning the original parameters. And our models are all trained with GCC3M to update $\hat{\phi_i}$ and $\hat{\phi_t}$ only. The output side of ResNet50, ResNet101, ViT-B/32 are 7 from input image size 224 without center crop. And the sizes of ViT-B/16 and ViT-L/14 are 14 and 16, respectively. The text prompt is "a photo of the". For the offline attention map, we try DINO \cite{caron2021emerging} and MoCoV3 \cite{chen2021empirical} which are pre-trained from ImageNet \cite{deng2009imagenet} without label. We extract the self-attention map from the last transformer layer, and select DINO (ViT-B/16) for the main experiments finally. The weights of new projections are updated by AdamW \cite{loshchilov2017decoupled}, at learning rate 1e-4, total batch size 1024, weight decay 0.05 for 10 epochs. Other training settings including augmentation, scheduler are followed by \cite{touvron2021training}.

\subsection{Results}

\textbf{Significant improvements compared with original CLIP.} We list the results on three datasets as shown in Tab. \ref{res_all_net}. It's clear that our ECLIP increases the explainability performances at large margins compared with the original CLIP. And these improvements are universal for all networks, regardless of vision transformer or convolutional networks. Look at the average results, our mean mIoU ranges from 41.18\% to 46.17\%, which beyond CLIP 27.26\% at most. For the mSC, ViT-B/32 achieves the highest result at 31.89\% and surpasses CLIP by 55.35\%. Among these networks, ViT-B/32 shows the highest explainability performance, while ViT-B/16 performs worst. As shown in Fig. \ref{problem} higher output resolution is more detailed, but the noise is more obvious, so patch size 16 performs worse.

\textbf{Compared with conventional explainability methods.} We pick ViT-B/32 and ViT-B/16 as backbones, because they perform the best and worst, respectively, in Tab. \ref{res_all_net}. Then we apply the most used gradient based method, Grad-CAM \cite{selvaraju2017grad}, and the latest explainability method \cite{chefer2021generic} for bi-model (based on \cite{chefer2021transformer} for transformer). Note, other methods for ViT (e.g. rollout \cite{abnar2020quantifying}) are not compared, because it's class agnostic only. Besides, we compare the ITSM of CLIP with our ECLIP and RCLIP (Eq. \ref{reverse}). These three methods are based on the raw feature map without backpropagation. So our methods allow batch operation for fast and straight forward visualization. As shown in Tab. \ref{res_methods}, Grad-CAM shows the same behavior as CLIP, which prefers the background more measured by mSC. The latest explainability method \cite{chefer2021generic} works better than Grad-CAM, but the mean mSC is still low at 6.53\% for ViT-B/16. It suggests its performance on dense prediction is bad, which limits its applicability. Lock at RCLIP, it's the simplest method without training or backpropagation, with performances obviously beyond the well-designed Bi-Model. For the ECLIP, it almost ranks first at all datasets and networks, and shows greatly improvements compared with other methods, e.g. 12.84\% average improvement at mIoU of ViT-B/32 and 16.03\% average improvement at mSC on ViT-B/16.

\begin{table*}
\setlength{\tabcolsep}{3mm}
\centering
\begin{tabular}{cccccccccc}
\hline 
  &  & \multicolumn{2}{c}{VOC 2012} &\multicolumn{2}{c}{COCO 2017} & \multicolumn{2}{c}{ImageNet-S50} & \multicolumn{2}{c}{\textbf{Average}}\\
Method & BP & mIoU $\uparrow$ & mSC $\uparrow$ & mIoU $\uparrow$ & mSC $\uparrow$ & mIoU $\uparrow$ & mSC $\uparrow$ & mIoU $\uparrow$ & mSC $\uparrow$ \\
\hline
\multicolumn{10}{c}{ViT-B/16} \\
\hdashline[0.5pt/5pt]
CLIP \cite{radford2021learning} & \XSolidBrush & 17.46 & -18.55 & 9.80 & -22.30 & 27.94 & -16.54 & 18.40 & -19.13\\
Grad-CAM \cite{selvaraju2017grad} & \checkmark & 18.12 & -12.30 & 10.04 & -16.66 & 29.10 & -7.81 & 19.09 & -12.26 \\
Bi-Model\cite{chefer2021generic} & \checkmark & 29.32 & 6.02 & 17.42 & 7.28 & 41.85 & 6.30 & 29.53 & 6.53 \\
\textbf{RCLIP} (ours) & \XSolidBrush & \underline{36.32} & \underline{18.56} & \textbf{24.80} & \textbf{22.30} & \underline{47.38} & \underline{16.54} & \underline{36.13} & \underline{19.13}\\
\textbf{ECLIP} (ours) & \XSolidBrush & \textbf{45.22} & \textbf{26.16} & \underline{23.39} & \underline{17.77} & \textbf{54.94} & \textbf{23.75} & \textbf{41.18} & \textbf{22.56}\\
\hdashline[0.5pt/5pt]
\multicolumn{10}{c}{ViT-B/32} \\
\hdashline[0.5pt/5pt]
CLIP \cite{radford2021learning} & \XSolidBrush & 17.52 & -25.01 & 10.14 & -22.54 & 27.89 & -22.83 & 18.52 & -23.46\\
Grad-CAM \cite{selvaraju2017grad} & \checkmark & 18.87 & -10.38 & 10.57 & -13.91 & 30.48 & -4.96 & 19.97 & -9.75\\
Bi-Model\cite{chefer2021generic} & \checkmark & 32.61 & 16.80 & 18.72 & 15.50 & 46.67 & 16.24 & 32.67 & 16.18 \\
\textbf{RCLIP} (ours) & \XSolidBrush & \underline{39.79} & \underline{25.01} & \underline{23.20} & \underline{22.54} & \underline{50.95} & \underline{22.83} & \underline{37.98} & \underline{23.46}\\
\textbf{ECLIP} (ours) & \XSolidBrush & \textbf{48.39} & \textbf{34.74} & \textbf{25.78} & \textbf{25.00} & \textbf{62.37} & \textbf{35.94} & \textbf{45.51} & \textbf{31.89}\\
\hline 
\end{tabular}

\caption{\label{res_methods}Results compared with conventional explainability methods, with average results. BP indicates backpropagation, which limits batch inference, and requires backpropagation multiple times for each image. mIoU (\%) measures the mask-level explainability, and mSC (\%) ranges from -100\% to 100\%, reflecting the score contrast between foreground and background. Best results are marked bold and second results are taged with underline. Note, Bi-Model \cite{chefer2021generic} is based on the latest explainability method \cite{chefer2021transformer} for ViT. Our RCLIP (Eq. \ref{reverse}, without training) and ECLIP (with MMP) surpass other methods at large margins without complex operations like backpropagation.}
\end{table*}

\subsection{Ablation Study}



\textbf{Effectiveness of masked max pooling.} From Tab. \ref{mmp_ablation}, we can see that max pooling correct the opposite visualization results from the original attention pooling (baseline), with improvements by 17.55\% and 30.53\% at mIoU and mSC, respectively. Based on max pooling, we further add attention map to form the masked max pooling in the bottom part. The MoCoV3 \cite{chen2021empirical} and DINO \cite{caron2021emerging} both improves the performance, via the guidance during training to emphasize the salient region. This operation requires no annotation, and doesn't bring cost to the inference phase.

\begin{table}[h]
\centering
\setlength\tabcolsep{1pt}
\begin{tabular}{cccc}
\hline 
Pooling & Attention Mask & mIoU (\%) $\uparrow$ & mSC (\%) $\uparrow$ \\
\hline
attntion (baseline) & no & 17.90 & -13.86 \\
average & no & 19.14 & -4.11 \\
max & no & \textbf{35.45$_{+17.55}$} & \textbf{16.67$_{+30.53}$} \\
\hline
max & no (baseline)& 35.45 & 16.67 \\
max & MoCoV3 \cite{chen2021empirical} & 44.70 & 25.03 \\
max & DINO \cite{caron2021emerging} & \textbf{45.52$_{+10.07}$} & \textbf{26.16$_{+9.49}$} \\
\hline
\end{tabular}
\caption{\label{mmp_ablation} Effectiveness of masked max pooling. Max is the key factor of performance improvement, and attention mask (class agnostic) further boost it. Results are obtained at the same settings}
\end{table}

\textbf{Necessity of new linear projections.} In ECLIP, we add a new pair of linear projections to explain the model for two reasons. (1) To remain the recognition performance without interference. In Tab. \ref{res_cls}, directly training on 3 million data from random weights leads to serious performance drop. Compared with it, ECLIP remain original performances, because original parameters are locked. And RCLIP doesn't require training. (2) Classification and interpretability suit different pool methods. In general experience, max pooling doesn't suit the classification task. If we use max pooling for CLIP to train from scratch, the recognition task must be sacrificed, even its interpretability is corrected. Thus, adding a new pair of linear projections is essential for both task, without interference to each other.

\begin{table}
\centering
\setlength\tabcolsep{2pt}
\begin{tabular}{ccccc}
\hline 
Method & Data & VOC12 & COCO17 & ImageNet-S50 \\
\hline
CLIP & - & 80.31 & 53.07 & 90.54 \\
CLIP & 3M & 46.78 & 19.37 & 35.86 \\
RCLIP & - & 80.31 & 53.07 & 90.54 \\
ECLIP & 3M & 80.31 & 53.07 & 90.54 \\
\hline
\end{tabular}
\caption{\label{res_cls} The recognition performances of ECLIP and RCLIP are the same to CLIP. Data indicates the training data quantity (million). The results are measured by mAP (\%) on CLIP ViT-B/16.}
\end{table}

\subsection{Visualization}
\begin{figure*}
\centering
  \rotatebox{90}{\small{\ \ \ VOC}} 
  \includegraphics[width=2.056cm,height=1.5cm]{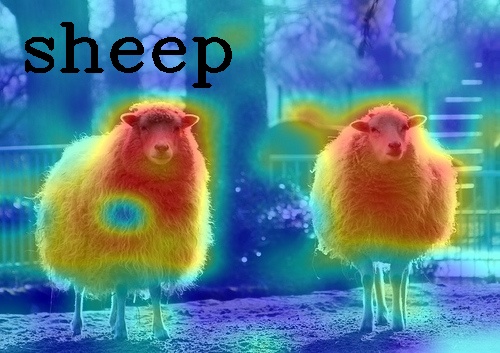} \includegraphics[width=2.056cm,height=1.5cm]{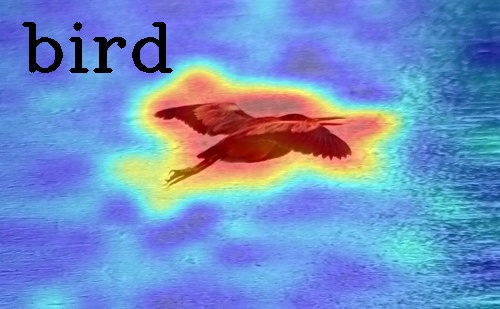} 
  \includegraphics[width=2.056cm,height=1.5cm]{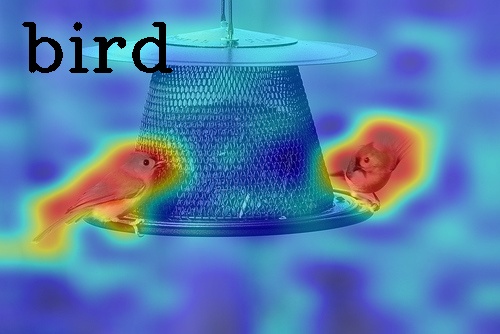}
    \includegraphics[width=2.056cm,height=1.5cm]{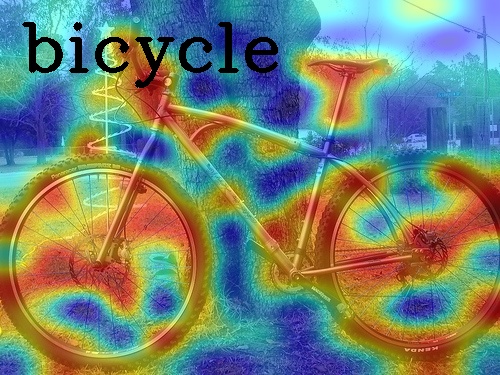}
  \includegraphics[width=2.056cm,height=1.5cm]{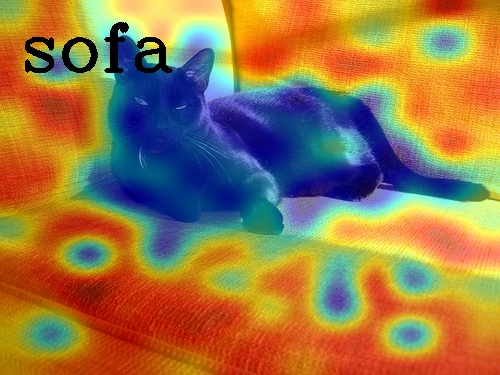}
  \includegraphics[width=2.056cm,height=1.5cm]{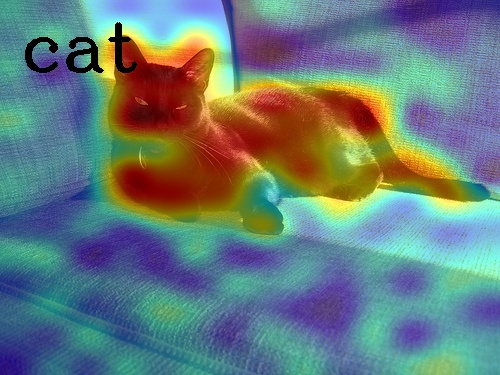}
  \includegraphics[width=2.056cm,height=1.5cm]{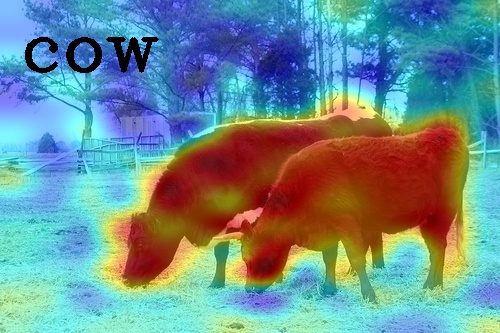}
  \includegraphics[width=2.056cm,height=1.5cm]{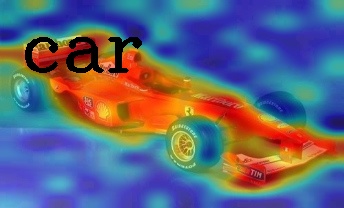}
  
  \rotatebox{90}{\small{\ \ COCO}} 
  \includegraphics[width=2.056cm,height=1.5cm]{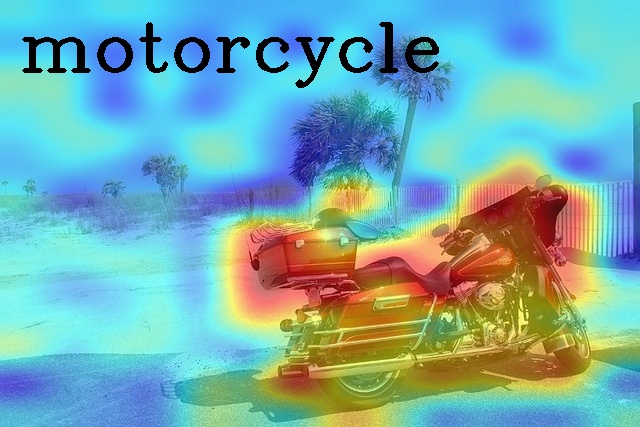} \includegraphics[width=2.056cm,height=1.5cm]{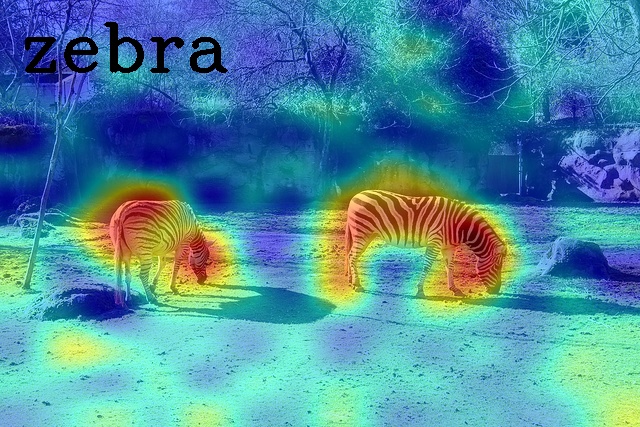} 
  \includegraphics[width=2.056cm,height=1.5cm]{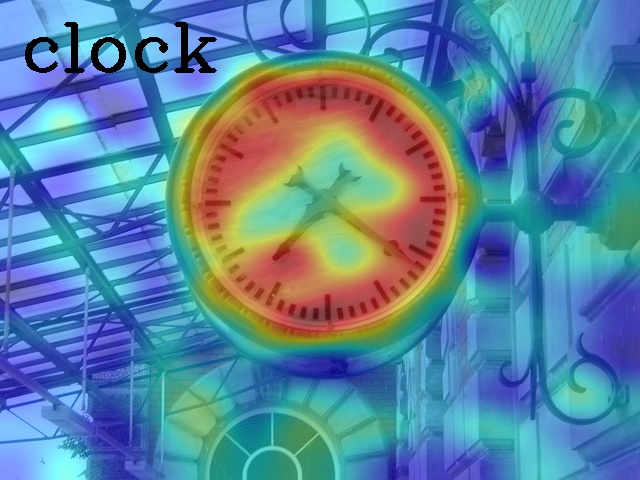}
    \includegraphics[width=2.056cm,height=1.5cm]{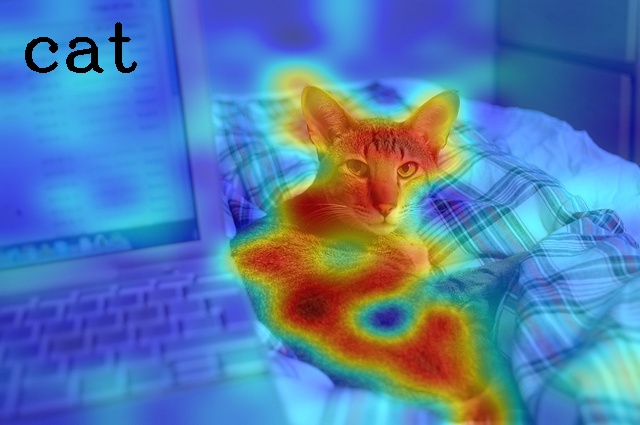}
  \includegraphics[width=2.056cm,height=1.5cm]{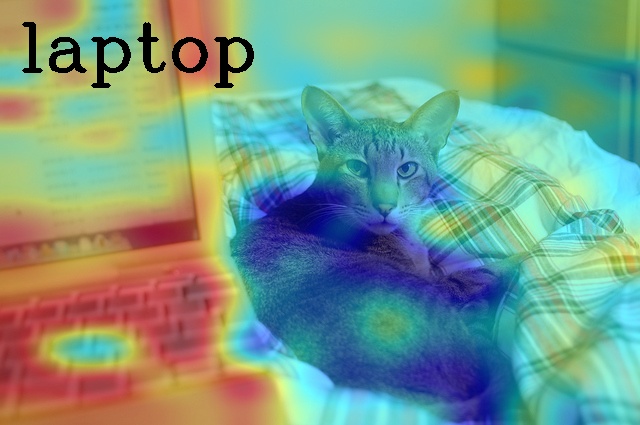}
  \includegraphics[width=2.056cm,height=1.5cm]{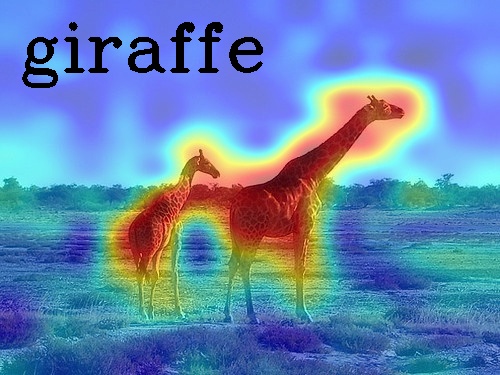}
  \includegraphics[width=2.056cm,height=1.5cm]{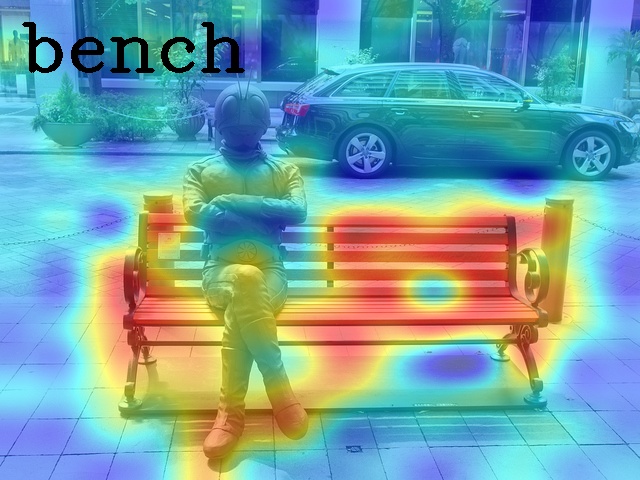}
  \includegraphics[width=2.056cm,height=1.5cm]{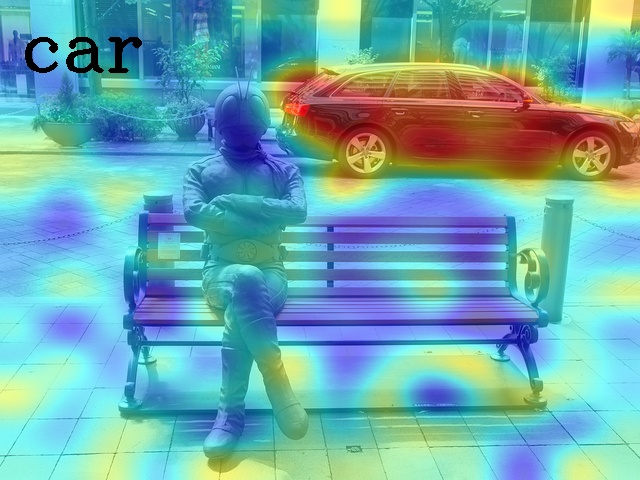}
  
    \hspace{-1.45mm}
  \rotatebox{90}{\small{\ ImageNet}}
  \includegraphics[width=2.056cm,height=1.5cm]{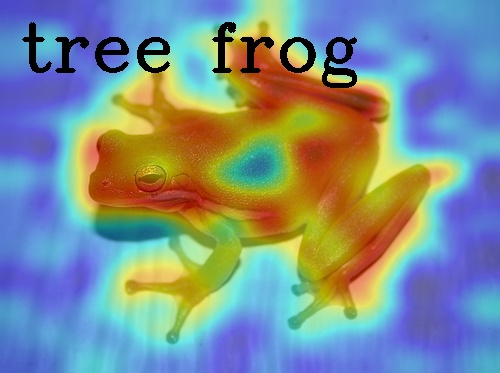} \includegraphics[width=2.056cm,height=1.5cm]{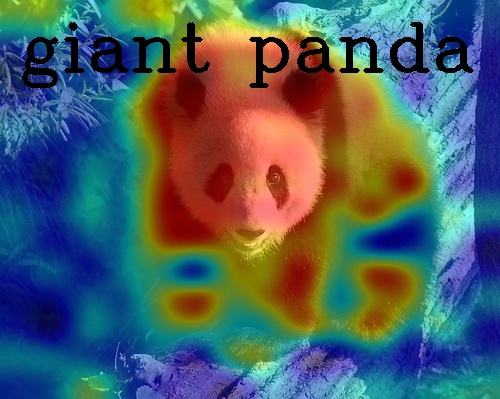} 
  \includegraphics[width=2.056cm,height=1.5cm]{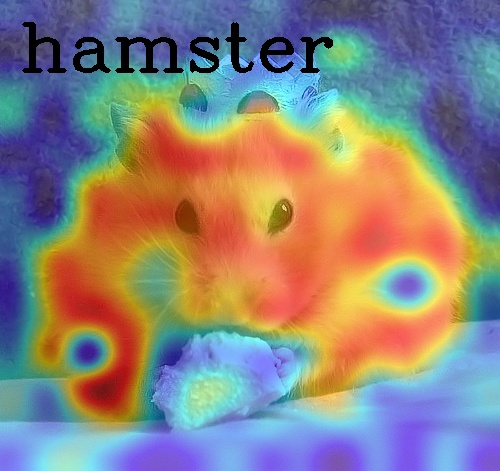}
    \includegraphics[width=2.056cm,height=1.5cm]{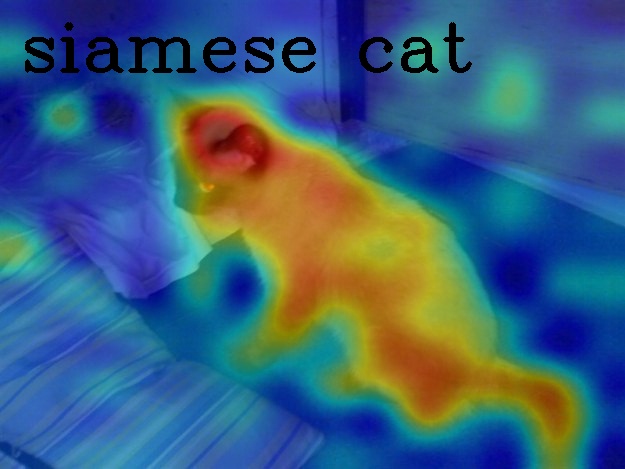}
  \includegraphics[width=2.056cm,height=1.5cm]{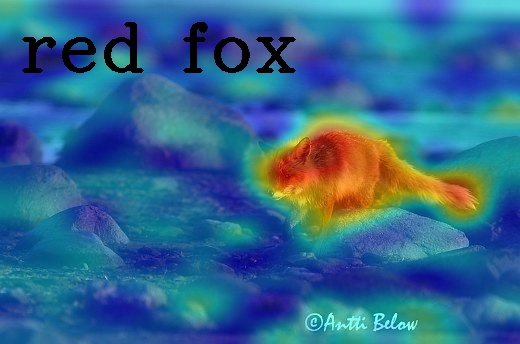}
  \includegraphics[width=2.056cm,height=1.5cm]{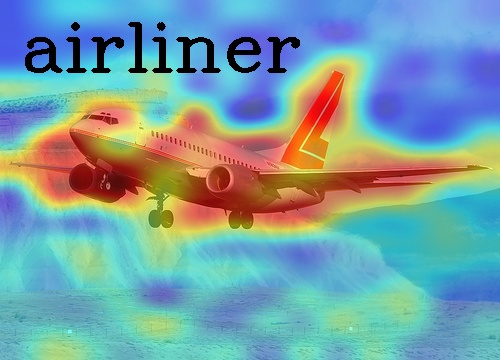}
  \includegraphics[width=2.056cm,height=1.5cm]{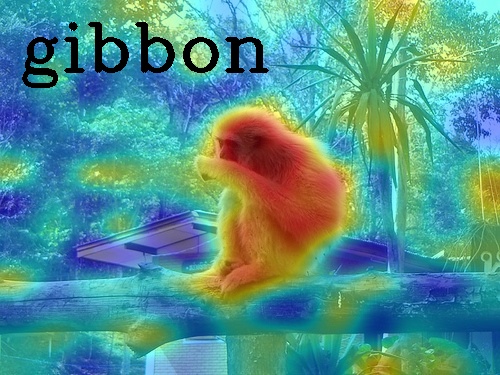}
  \includegraphics[width=2.056cm,height=1.5cm]{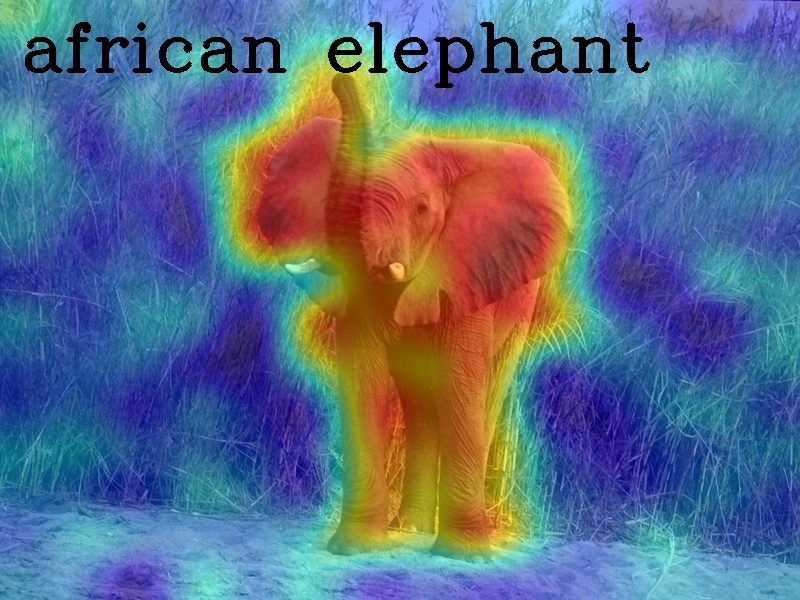}
\caption{Qualitative visualization results with correspondent texts from our ECLIP ViT-B/16 on three datasets.}
\label{vis}
\end{figure*}

\begin{figure}[h]
\centering
\begin{subfigure}{.096\textwidth}
  \includegraphics[width=1.64cm]{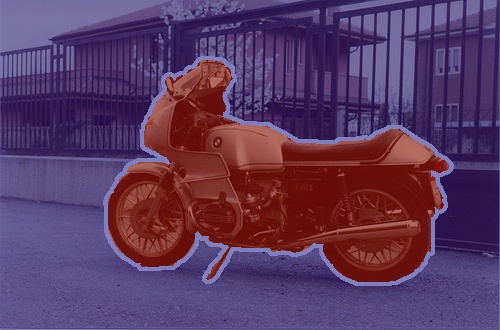} \\ 
  \includegraphics[width=1.64cm]{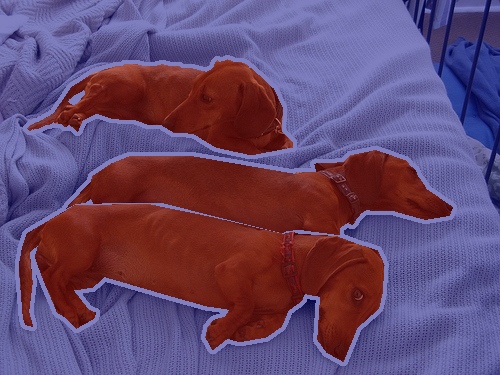} \\
  \includegraphics[width=1.64cm]{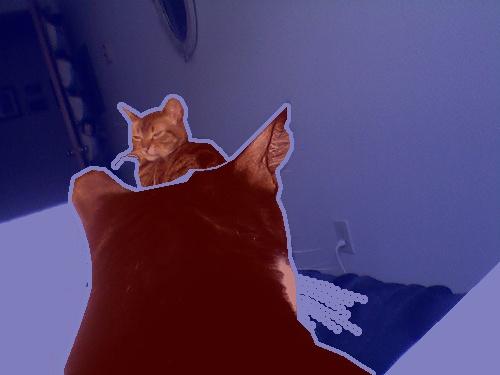} \\
  \includegraphics[width=1.64cm,height=1.9cm]{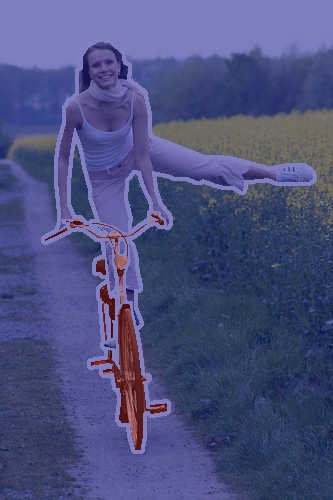} \\
  \includegraphics[width=1.64cm,height=1.9cm]{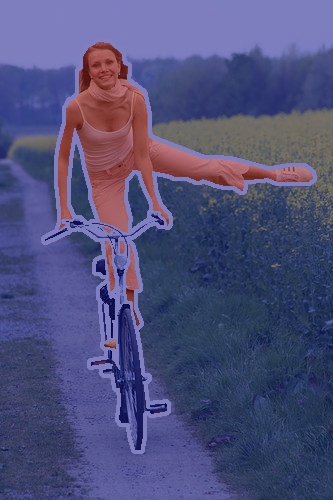} \\
  \includegraphics[width=1.64cm,height=1.85cm]{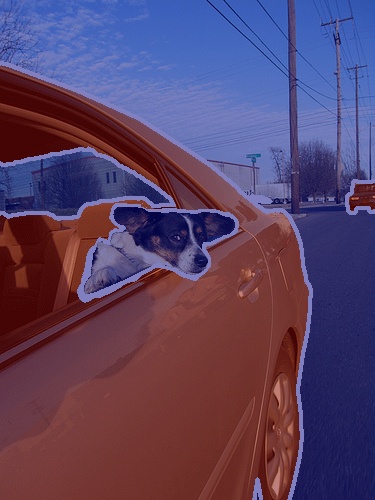} \\
  \includegraphics[width=1.64cm,height=1.85cm]{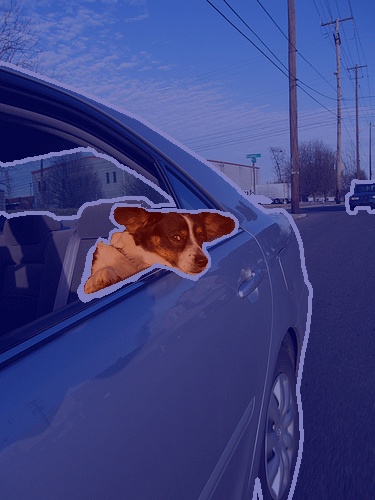} \\
  \caption{GT \centering}
\end{subfigure}%
\begin{subfigure}{.096\textwidth}
  \includegraphics[width=1.64cm]{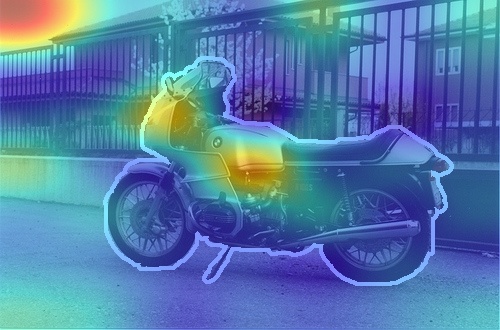} \\ 
  \includegraphics[width=1.64cm]{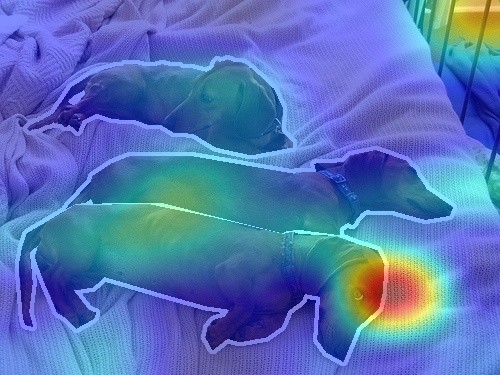} \\
  \includegraphics[width=1.64cm]{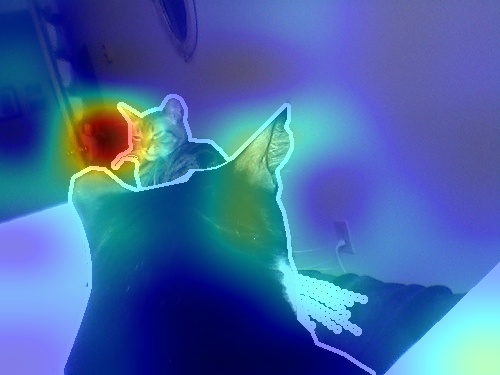} \\
  \includegraphics[width=1.64cm,height=1.9cm]{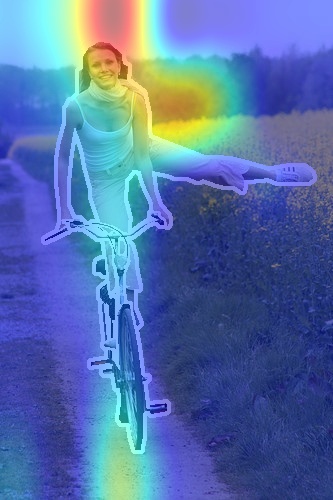} \\
  \includegraphics[width=1.64cm,height=1.9cm]{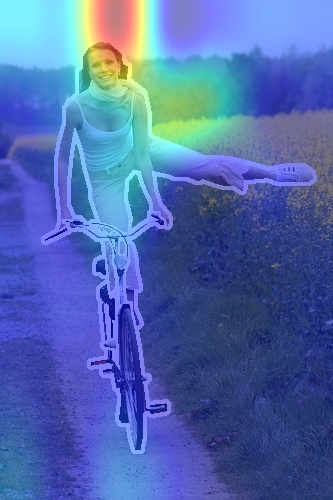} \\
  \includegraphics[width=1.64cm,height=1.85cm]{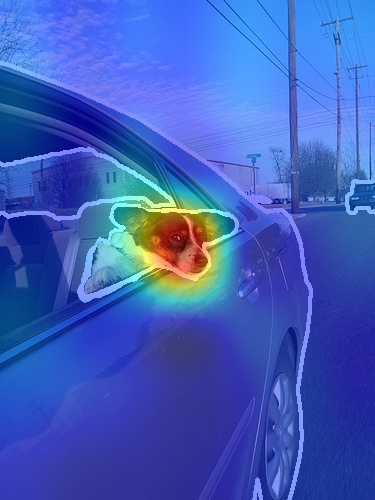} \\
  \includegraphics[width=1.64cm,height=1.85cm]{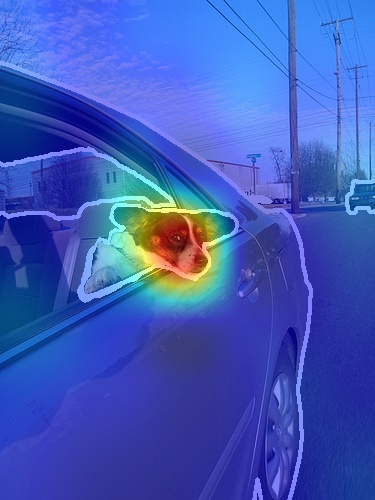} \\
  \caption{\cite{chefer2021generic} \centering}
\end{subfigure}%
\begin{subfigure}{.096\textwidth}
  \includegraphics[width=1.64cm]{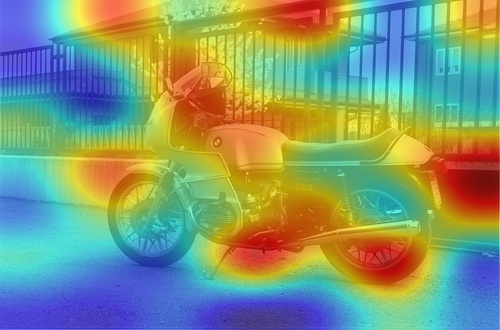} \\ 
  \includegraphics[width=1.64cm]{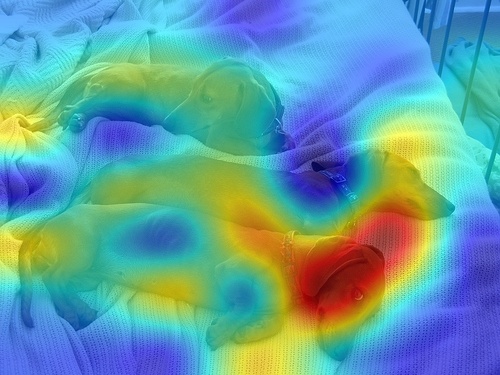} \\
  \includegraphics[width=1.64cm]{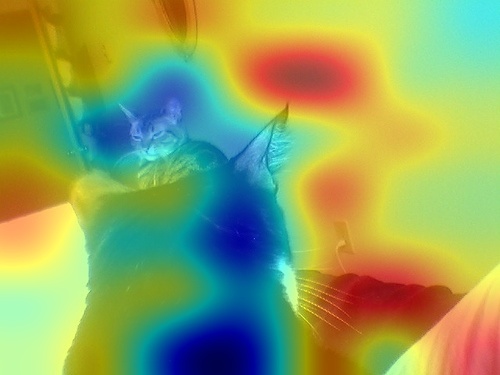} \\
  \includegraphics[width=1.64cm,height=1.9cm]{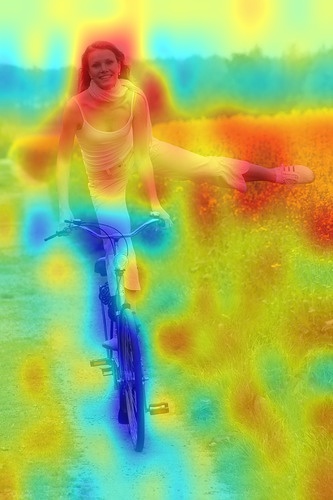} \\
  \includegraphics[width=1.64cm,height=1.9cm]{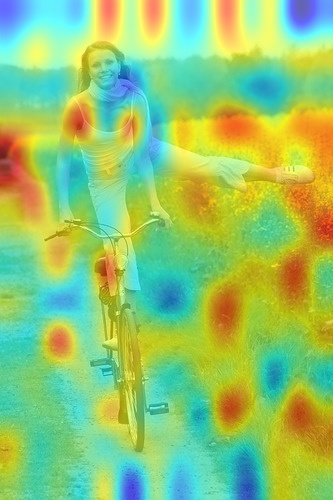} \\
  \includegraphics[width=1.64cm,height=1.85cm]{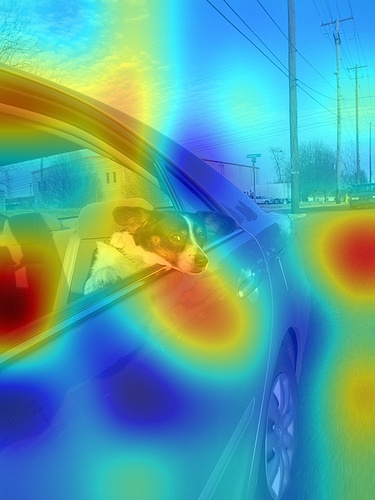} \\
  \includegraphics[width=1.64cm,height=1.85cm]{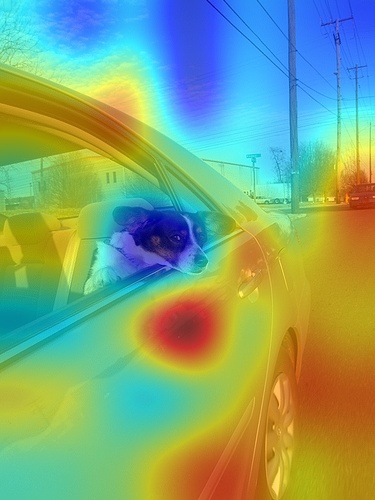} \\
  \caption{\cite{selvaraju2017grad} \centering}
\end{subfigure}%
\begin{subfigure}{.096\textwidth}
  \includegraphics[width=1.64cm]{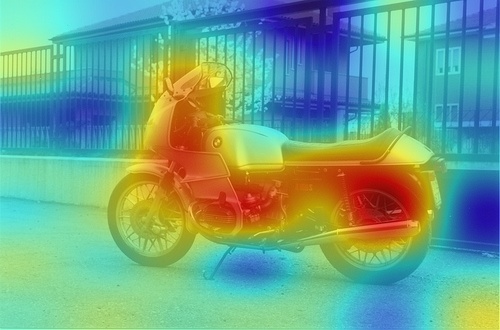} \\ 
  \includegraphics[width=1.64cm]{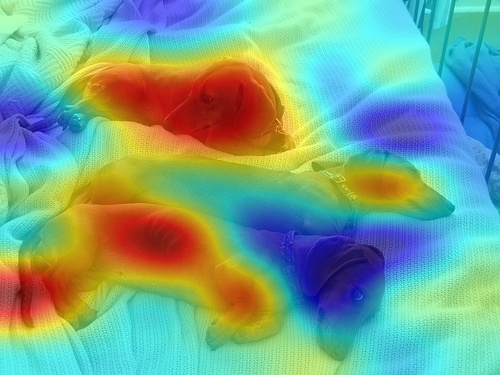} \\
  \includegraphics[width=1.64cm]{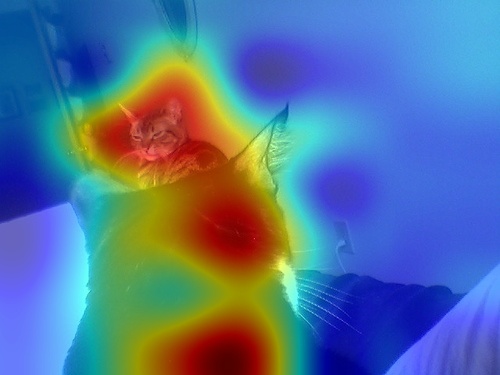} \\
  \includegraphics[width=1.64cm,height=1.9cm]{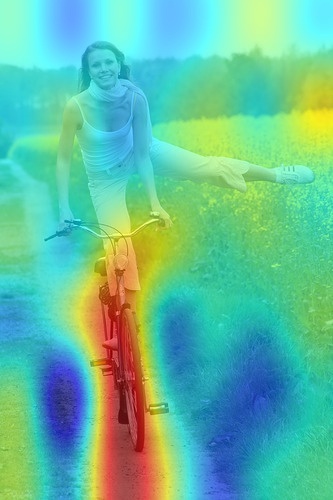} \\
  \includegraphics[width=1.64cm,height=1.9cm]{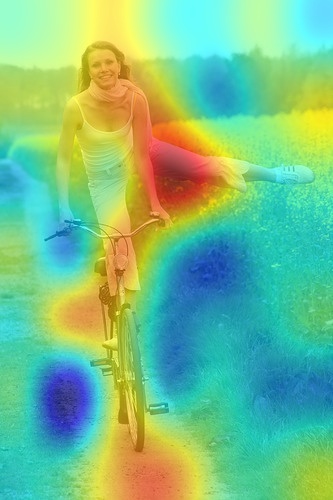} \\
  \includegraphics[width=1.64cm,height=1.85cm]{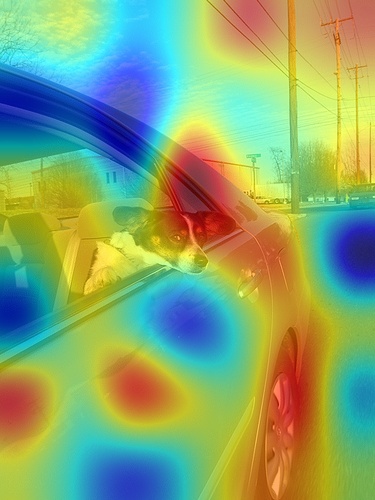} \\
  \includegraphics[width=1.64cm,height=1.85cm]{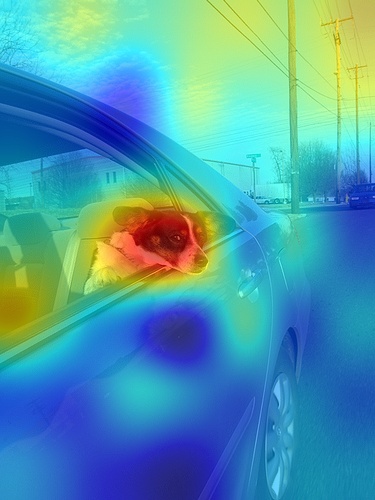} \\
  \caption{RCLIP \centering}
\end{subfigure}%
\begin{subfigure}{.096\textwidth}
  \includegraphics[width=1.64cm]{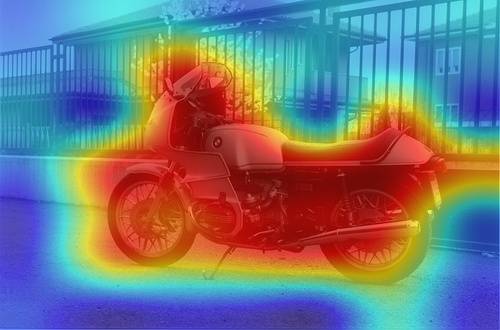} \\ 
  \includegraphics[width=1.64cm]{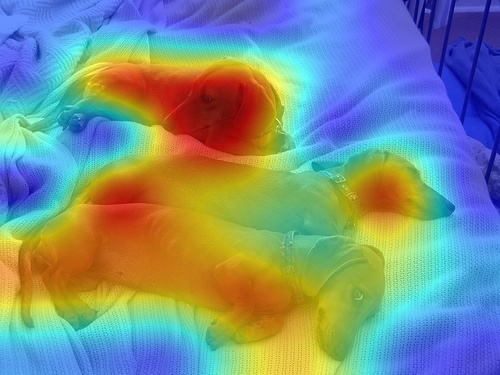} \\
  \includegraphics[width=1.64cm]{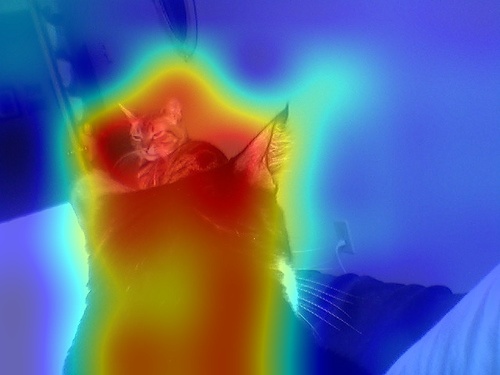} \\
  \includegraphics[width=1.64cm,height=1.9cm]{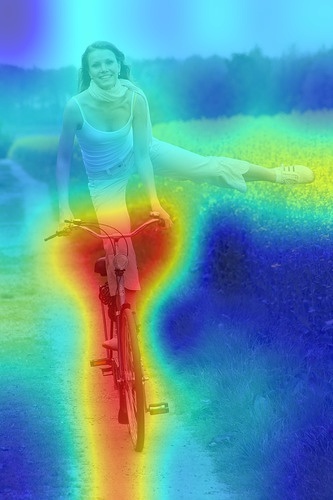} \\
  \includegraphics[width=1.64cm,height=1.9cm]{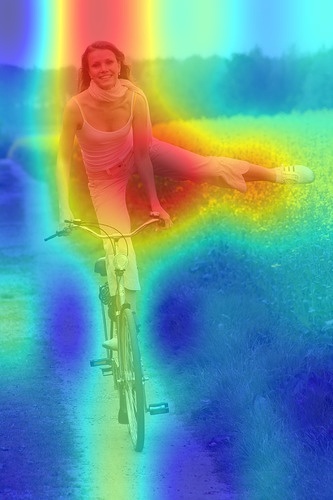} \\
  \includegraphics[width=1.64cm,height=1.85cm]{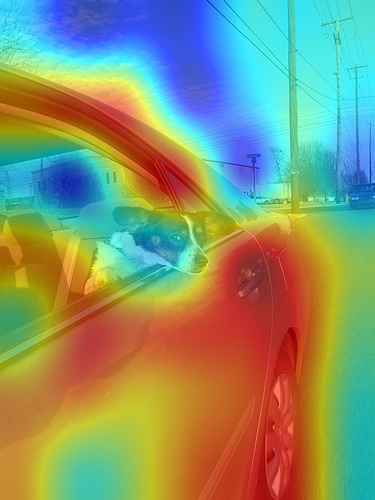} \\
  \includegraphics[width=1.64cm,height=1.85cm]{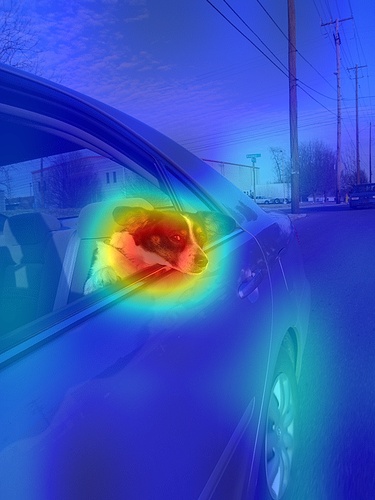} \\
  \caption{ECLIP \centering}
\end{subfigure}%

\caption{Visualization comparison between conventional methods and ours. \cite{chefer2021generic} (Bi-Model) is the latest method based on \cite{chefer2021transformer} for transformer, and \cite{selvaraju2017grad} (Grad-CAM) is the most used gradient based method. Our RCLIP is simpler and better than these well-designed methods, and ECLIP show the best explainability. Note, \cite{chefer2021generic} applies class agnostic self-attention in visualization, thus confused at multi-label sometimes, while ECLIP uses attention during training only. All methods are based on CLIP ViT-B/32. Predicted foreground is close to red.}
\label{vis_grad_method}
\end{figure}

As shown in Fig.\ref{vis}, we draw the qualitative visualization results on VOC12, COCO17 and ImageNet. From these qualitative results, we believe the proposed method is able to explain which parts influence the predictions most, and help us to understand the model for better credibility. Also, these results show the potentiality for tasks like segmentation and localization. Then we compare the conventional methods with ours as Fig. \ref{vis_grad_method}. The latest method \cite{chefer2021generic} for transformer is confused at multiple objects and locate partial regions, because it uses the class agnostic self-attention in bad quality during visualization. And Grad-CAM \cite{selvaraju2017grad}, the most used gradient based method, behaves like CLIP in Fig. \ref{problem} and often focus on the backgrounds. Our RCLIP is simpler than these well-designed methods, also performs better. And ECLIP has the best explainability over others.

\section{Conclusion}
In summary, we visually explain the Contrastive Language-Image Pre-training (CLIP) from its raw prediction map. And we observe that CLIP is prone to focus on the background, presenting opposite visualization results. Then, the reason is located as the pooling module, where features are shifted to opposite semantic regions. To solve the problem, the masked max pooling is proposed to correct the explainability, with free attention mask. Then we propose the Explainable  CLIP (ECLIP) based on this pooling, and a simple version, reversed CLIP (RCLIP). Our methods achieve nontrivial explainability improvements over original CLIP and conventional methods at large margins.

\clearpage

{\small
\bibliographystyle{ieee_fullname}
\bibliography{egbib}
}

\clearpage
\appendix

\section{Analyze of semantic shift.}
To support semantic shift, the explanation for the problem of opposite visualization in CLIP \cite{radford2021learning}, we list the details of of semantic shift in Tab. \ref{num_shift}. For different foreground sizes, the ratio of B2F (confident features shifted from background to foreground) and F2B (foreground features shifted to background) varies. For the size (0.25,0.75), the shifted feature takes about half (248.3 out of 512 channels after linear projection), which suggests the feature shift between opposite semantic regions is universal, and takes a large ratio. Then, these features at opposite semantic regions forms an erroneous prediction map.

\begin{table}[h]
\centering
\setlength\tabcolsep{10pt}
\begin{tabular}{cccc}
\hline 
Size & B2F & F2B & B2B/F2F \\
\hline
(0, 0.25] & 78.0 & 45.2 & 388.7 \\
(0.25, 0.75] & 111.8 & 136.5 & 263.7 \\
(0.75, 1] & 44.7 & 97.0 & 370.3 \\
(0, 1] & 78.8 & 114.7 & 318.5 \\
\hline 
\end{tabular}
\caption{\label{num_shift} Statistics of mean shifted channel number at varied foreground sizes on VOC12 \cite{everingham2010pascal} for CLIP ViT-B/16. B2F means discriminative features shifted from background to foreground, F2B is opposite to it, and B2B/F2F indicates the number of features without semantic shift. The shifted features take a large ratio and constitute finial predictions, which suggests semantic shift is the reason of opposite results.}
\end{table}

\begin{figure}[t]
\centering
\begin{subfigure}{.24\textwidth}
  \includegraphics[width=4.1cm,height=3.8cm]{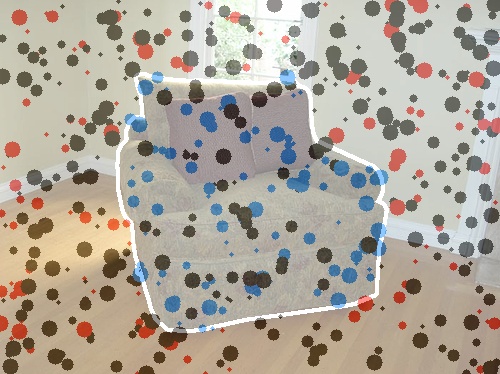} \\
  \includegraphics[width=4.1cm,height=3.8cm]{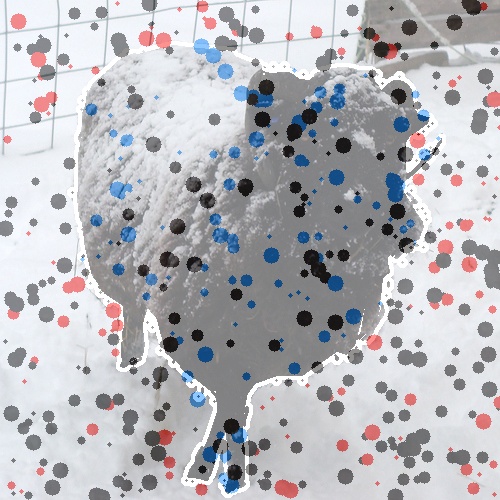} \\
  \includegraphics[width=4.1cm,height=3.8cm]{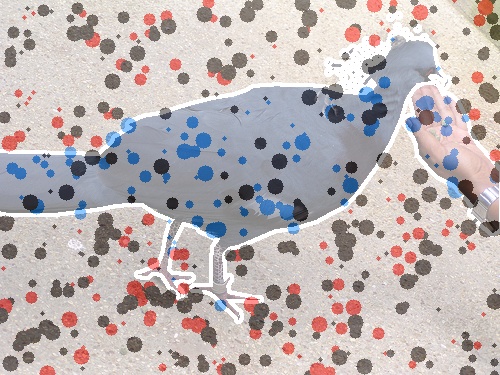} \\
  \caption{\centering Semantic Shift}
  \label{shift}
\end{subfigure}%
\begin{subfigure}{.24\textwidth}
  \includegraphics[width=4.1cm,height=3.8cm]{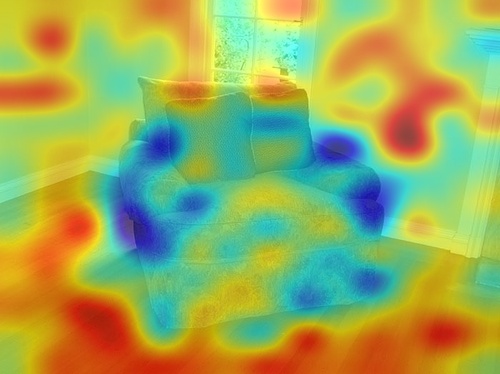} \\
  \includegraphics[width=4.1cm,height=3.8cm]{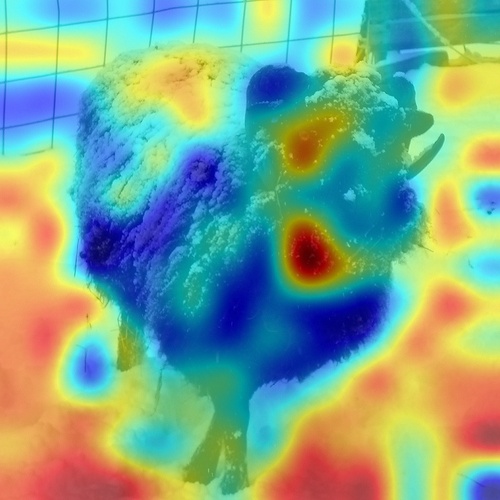} \\
  \includegraphics[width=4.1cm,height=3.8cm]{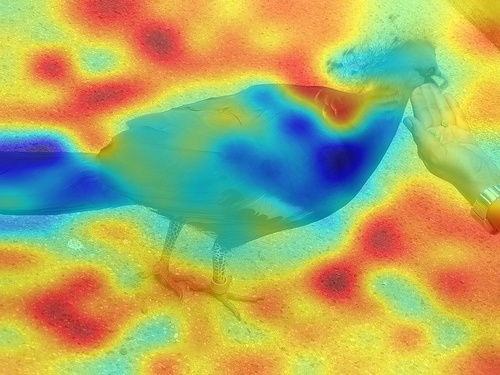} \\

  \caption{\centering ITSM of original CLIP}
  \label{_itsm}
\end{subfigure}%
\caption{Illustration of feature shift with ITSM on ViT-B/16. we draw one point for each channel on the image. And the location of a point is same to the pixel, whose score is most close to the pooled score. Note, larger points means higher pooled value. Blue points indicate features shifted from background to foreground, and red points are opposite. For ITSM, predicted foreground is colored in red. The semantic shift is compared between max pooling (max localization ability) and original attention pooling.} 
\label{shift_vs_itsm}
\end{figure}

Then, we list more examples of semantic shift with corresponding Image-Text Similarity Map (ITSM) on CLIP ViT-B/16 and ViT-L/14 as \ref{shift_vs_itsm}. From these images, we can see that shifted features in blue (background to foreground) and red (foreground to background) take a large promotion. Besides, we can find obvious gray points on these examples, where opposite visualization doesn't happen on ITSM (red regions on foreground).

\ \\
\ \\
\ \\
\ \\
\ \\
\ \\
\ \\
\ \\
\ \\
\ \\
\ \\
\ \\
\ \\
\ \\
\ \\
\ \\
\ \\
\ \\
\ \\
\ \\
\ \\
\ \\
\ \\
\ \\
\ \\
\ \\
\ \\
\ \\
\ \\

\section{Opposite Visualization is Universal for CLIP}
\begin{figure*}[ht]
\leftline{ \ \ \ \ \ \ Ground Truth   \ \ \ \ \ \ \ \ \ \ \ \ ResNet50  \ \ \ \ \ \ \ \ \ \ \ \ \ \ \ \ ResNet101 \ \ \ \ \ \ \ \ \ \ \ \ \ \ \ ViT-B/32  \ \ \ \ \ \ \ \ \ \ \ \ \ \ \ \ \ ViT-B/16  \ \ \ \ \ \ \ \ \ \ \ \ \ \ \ \ ViT-L/14 }
\vspace{2mm}
\centering
  \includegraphics[width=2.8cm,height=2.5cm]{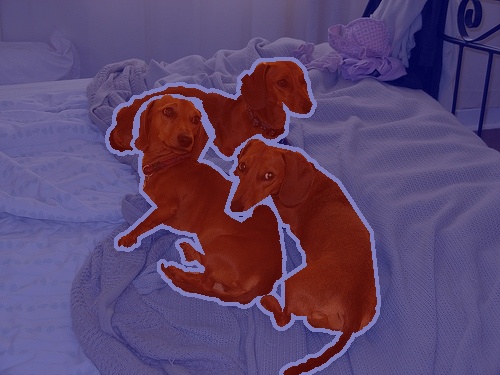} \includegraphics[width=2.8cm,height=2.5cm]{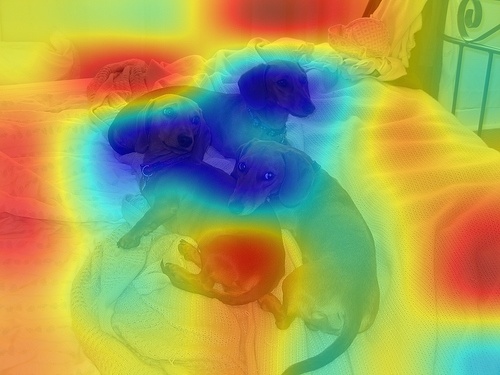}   \includegraphics[width=2.8cm,height=2.5cm]{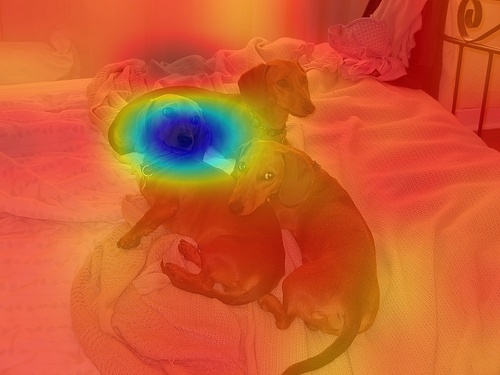} \includegraphics[width=2.8cm,height=2.5cm]{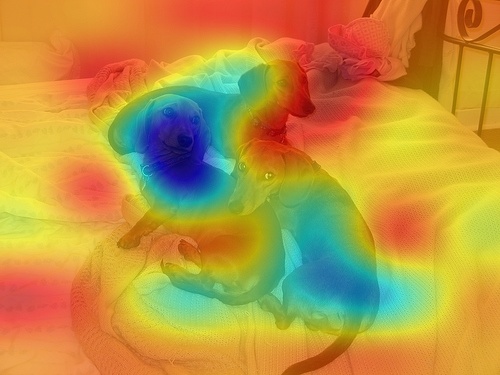} \includegraphics[width=2.8cm,height=2.5cm]{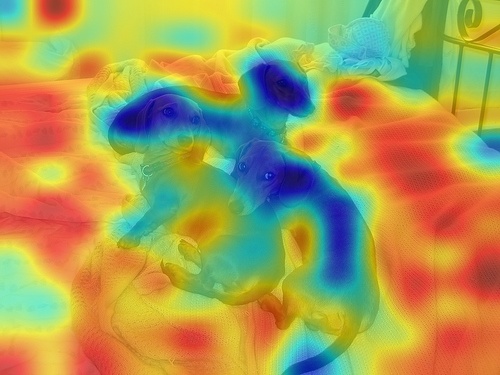} \includegraphics[width=2.8cm,height=2.5cm]{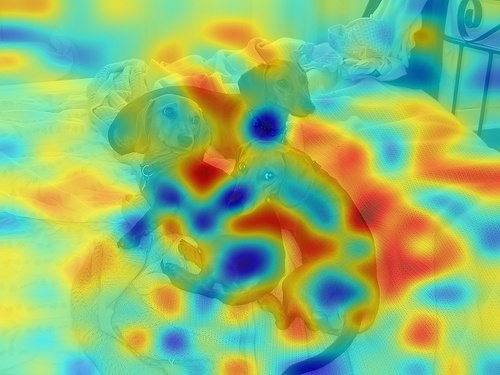} \\

  \includegraphics[width=2.8cm,height=2.5cm]{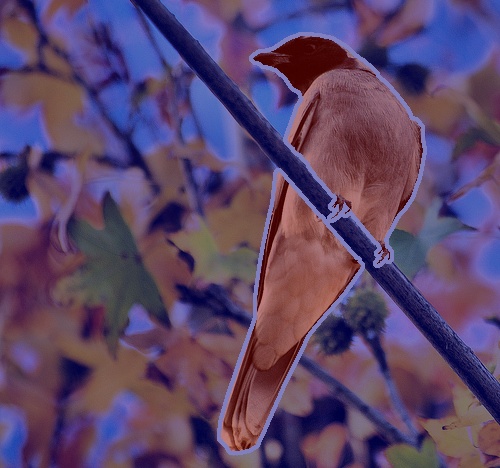} \includegraphics[width=2.8cm,height=2.5cm]{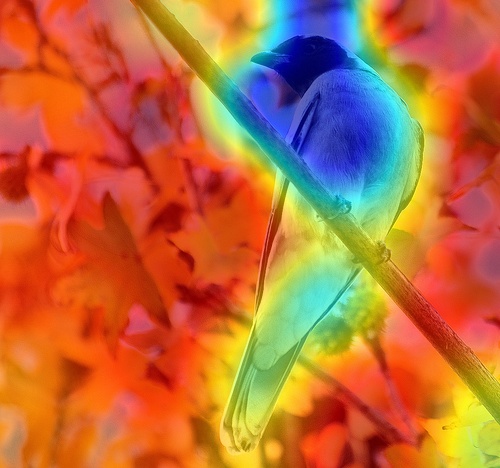}   \includegraphics[width=2.8cm,height=2.5cm]{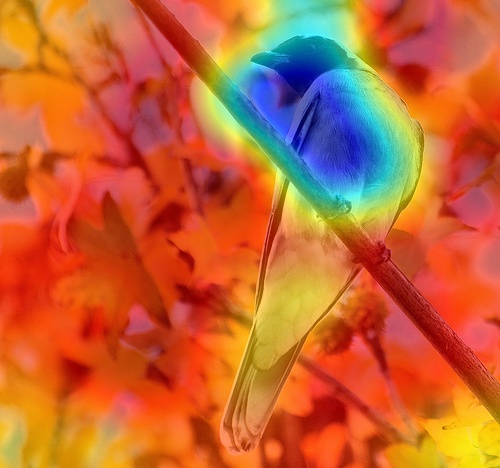} \includegraphics[width=2.8cm,height=2.5cm]{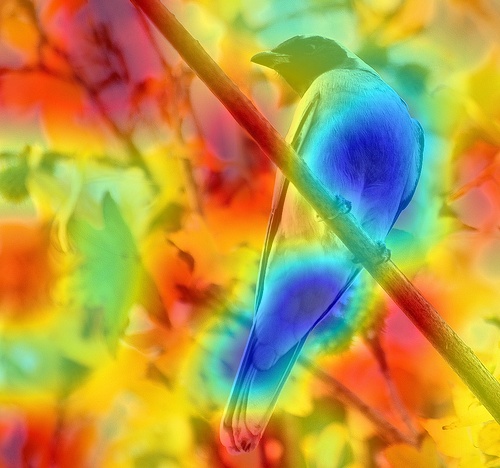} \includegraphics[width=2.8cm,height=2.5cm]{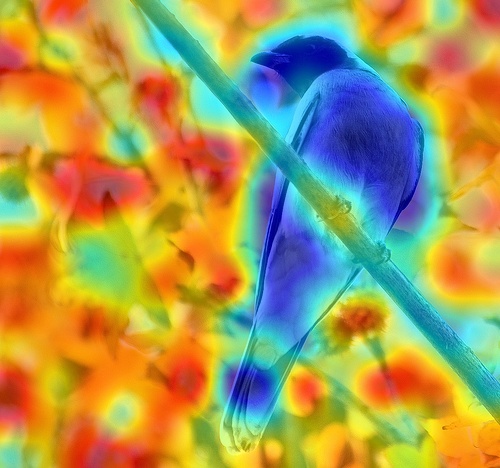} \includegraphics[width=2.8cm,height=2.5cm]{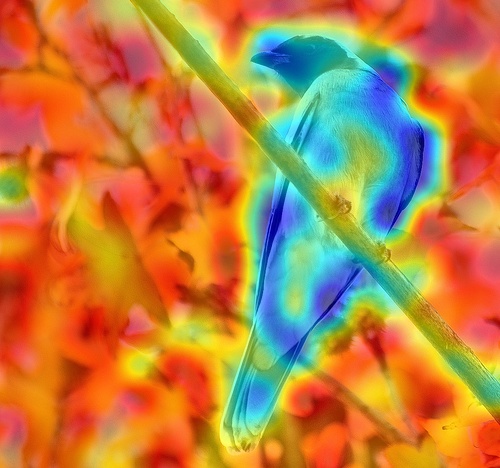} \\
  
  \includegraphics[width=2.8cm,height=2.5cm]{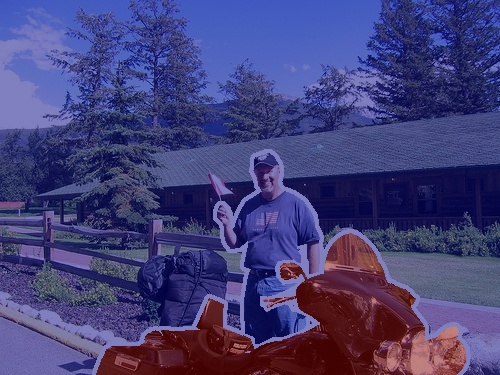} \includegraphics[width=2.8cm,height=2.5cm]{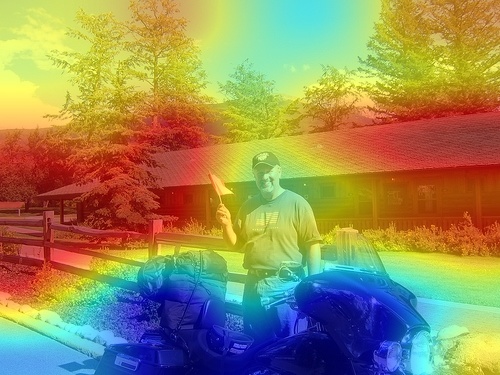}   \includegraphics[width=2.8cm,height=2.5cm]{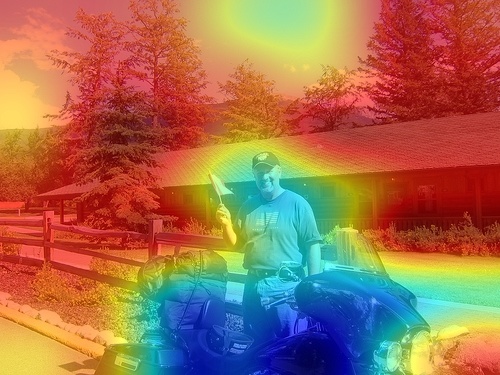} \includegraphics[width=2.8cm,height=2.5cm]{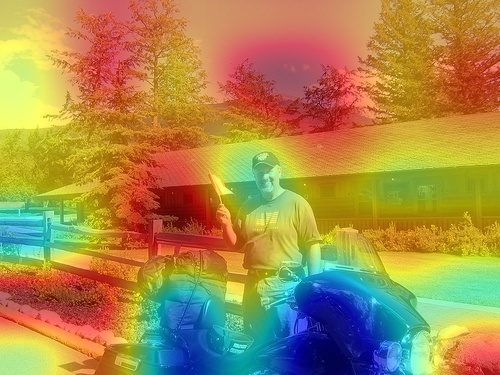} \includegraphics[width=2.8cm,height=2.5cm]{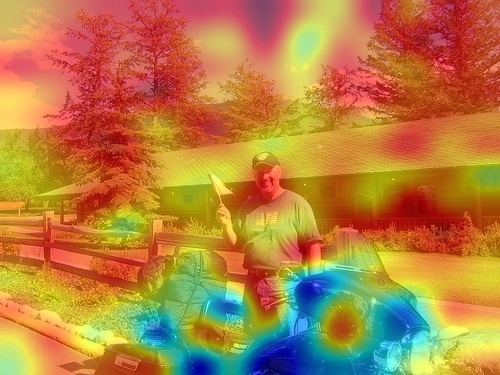} \includegraphics[width=2.8cm,height=2.5cm]{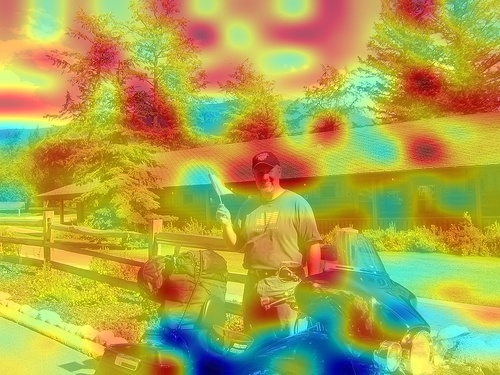} \\
  
  \includegraphics[width=2.8cm,height=2.5cm]{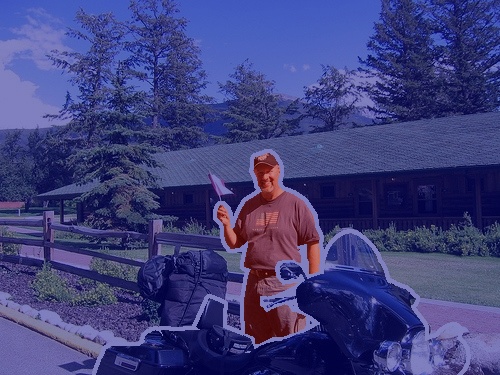} \includegraphics[width=2.8cm,height=2.5cm]{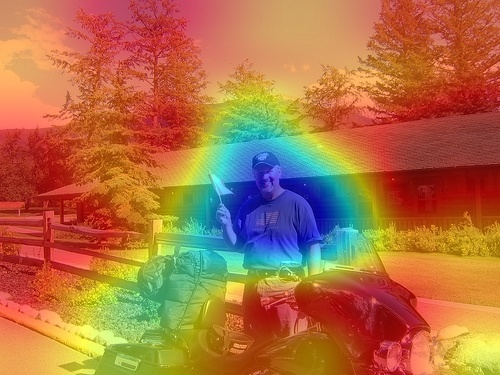}   \includegraphics[width=2.8cm,height=2.5cm]{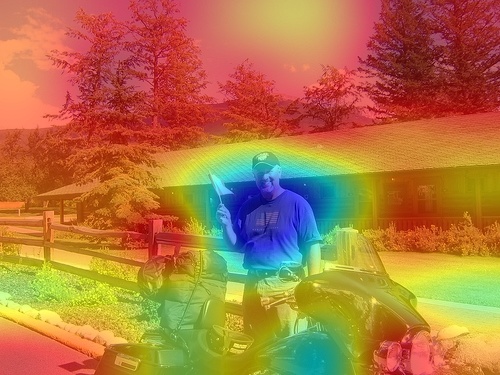} \includegraphics[width=2.8cm,height=2.5cm]{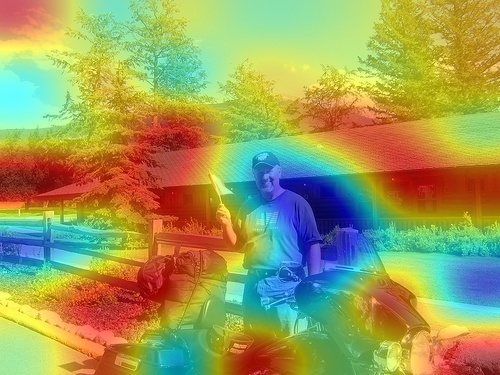} \includegraphics[width=2.8cm,height=2.5cm]{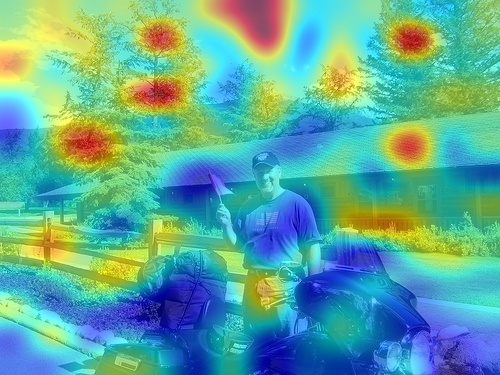} \includegraphics[width=2.8cm,height=2.5cm]{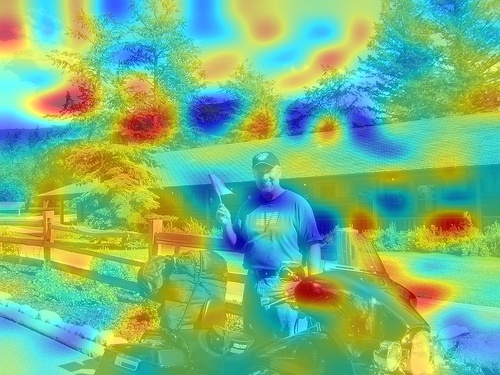} \\
  
  \includegraphics[width=2.8cm,height=2.5cm]{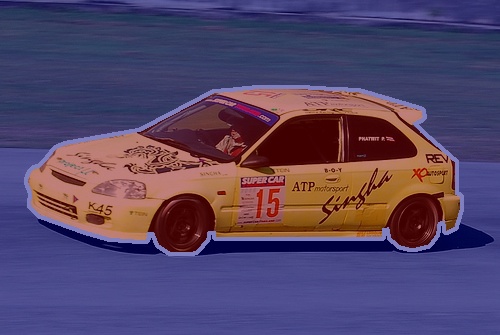} \includegraphics[width=2.8cm,height=2.5cm]{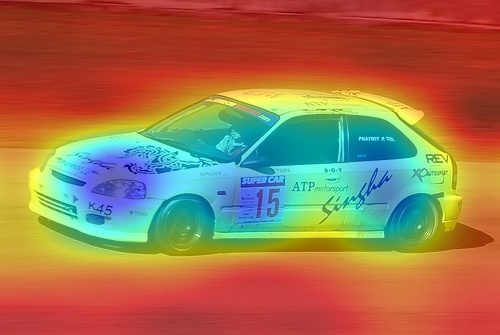}   \includegraphics[width=2.8cm,height=2.5cm]{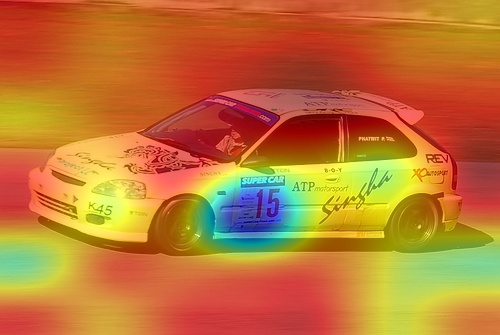} \includegraphics[width=2.8cm,height=2.5cm]{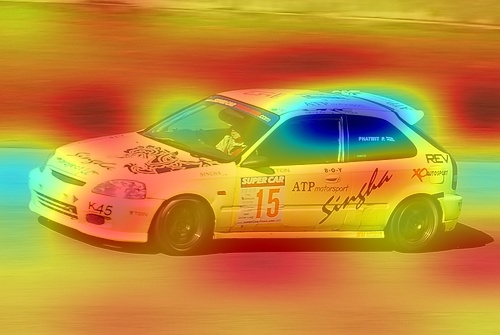} \includegraphics[width=2.8cm,height=2.5cm]{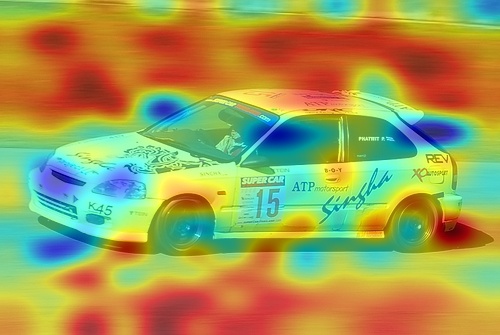} \includegraphics[width=2.8cm,height=2.5cm]{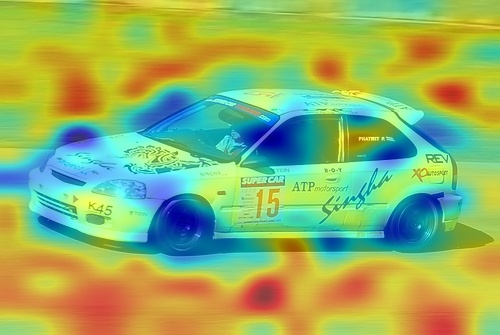} \\
  
  \includegraphics[width=2.8cm,height=2.5cm]{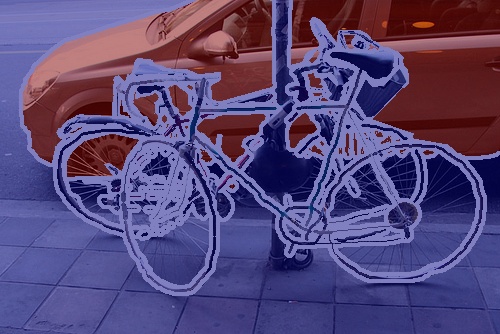} \includegraphics[width=2.8cm,height=2.5cm]{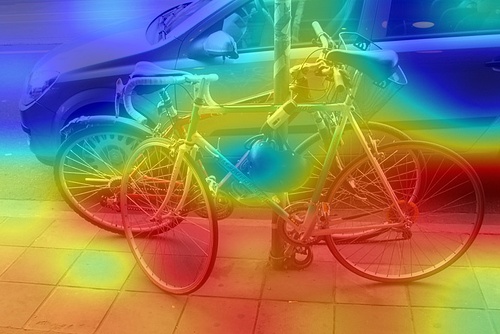}   \includegraphics[width=2.8cm,height=2.5cm]{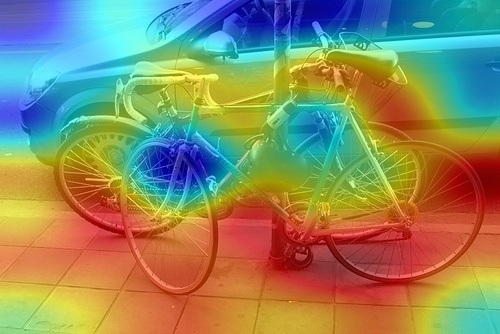} \includegraphics[width=2.8cm,height=2.5cm]{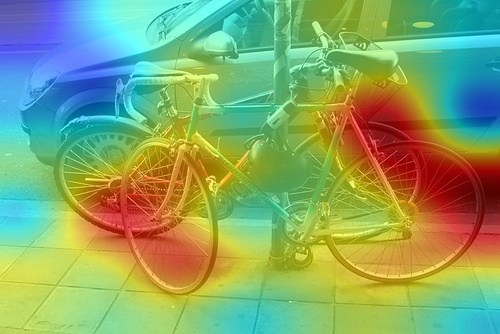} \includegraphics[width=2.8cm,height=2.5cm]{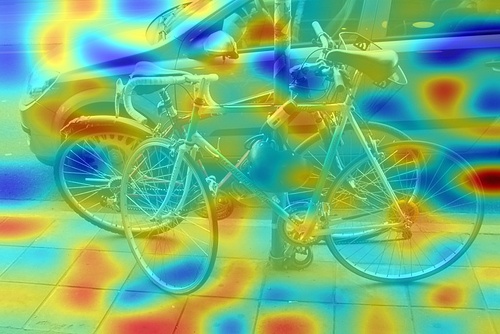} \includegraphics[width=2.8cm,height=2.5cm]{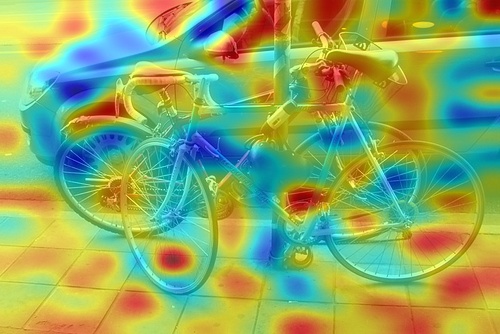} \\

  \includegraphics[width=2.8cm,height=2.5cm]{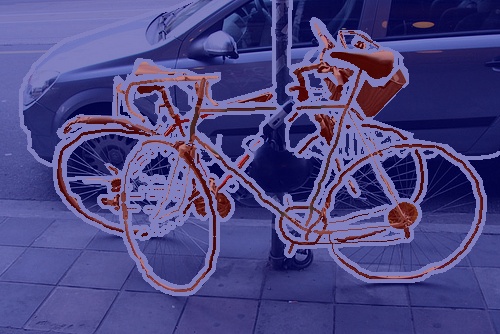} \includegraphics[width=2.8cm,height=2.5cm]{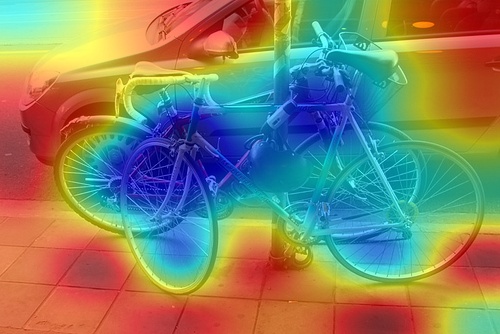}   \includegraphics[width=2.8cm,height=2.5cm]{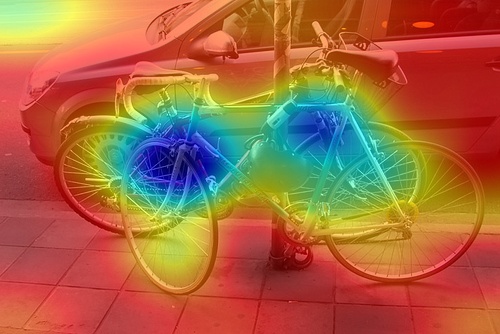} \includegraphics[width=2.8cm,height=2.5cm]{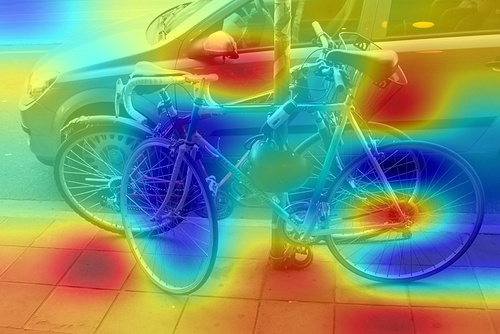} \includegraphics[width=2.8cm,height=2.5cm]{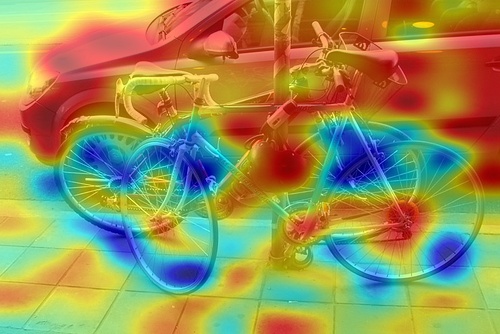} \includegraphics[width=2.8cm,height=2.5cm]{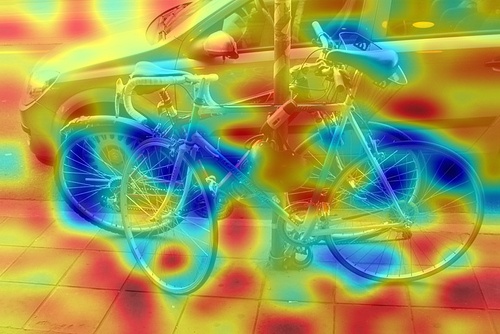} \\

\caption{Visualization of Image-Text Similarity Map for CLIP on varied backbones. Note, foreground is close to red and background is color in blue. The opposite visualization results are universal for convolutional networks \cite{he2016deep} and vision transformers \cite{dosovitskiy2020image}.}
\label{vis_net}
\end{figure*}

As shown in Fig. \ref{vis_net}, the opposite visualization results are universal for CLIP, regardless of convolutional networks or vision transformers. Besides, lower output size (e.g. 7 $\times$ 7 for ResNets \cite{he2016deep} and ViT-B/16 \cite{dosovitskiy2020image}) show smoother results than larger output (14 $\times$ 14 of ViT-B/16 and 16 $\times$ 16 of ViT-L/14). Thus, lower output side seems better, while networks at larger output sides provide detailed results. Out of these differences, the opposite explainability of CLIP is common.

\section{Break the Limitation of Output Size}
\begin{figure*}[t]
\leftline{\ \ \ \ \ \ \ \ \ \ \ \ Ground Truth   \ \ \ \ \ \ \ \ \ \ \  Bi-Model ($\times$7) \cite{chefer2021generic}  \ \ \ \ \ \ \  Bi-Model ($\times$14) \cite{chefer2021generic} \ \ \ \ \ \ \ \ \ \ \ ECLIP ($\times$7)  \ \ \ \ \ \ \ \ \ \ \ \ \ \ \ \ ECLIP ($\times$14)}
\vspace{2mm}
\centering
  \includegraphics[width=3.3cm]{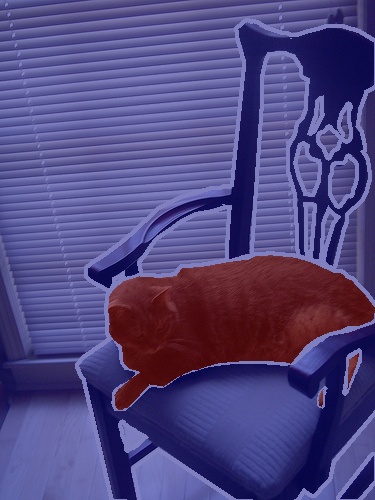} \includegraphics[width=3.3cm]{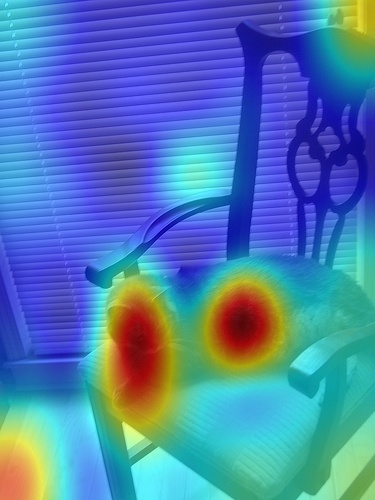}   \includegraphics[width=3.3cm]{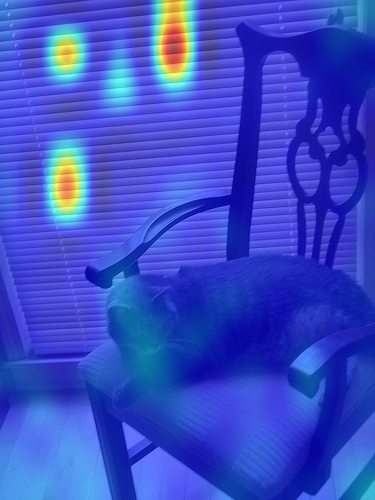} \includegraphics[width=3.3cm]{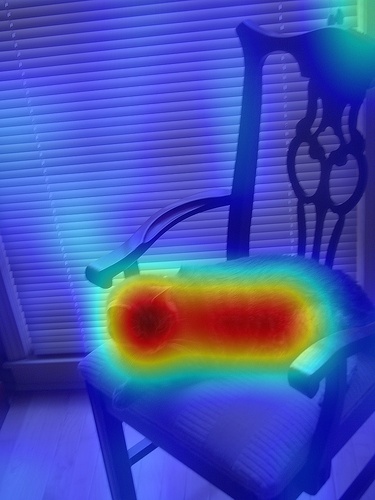} \includegraphics[width=3.3cm]{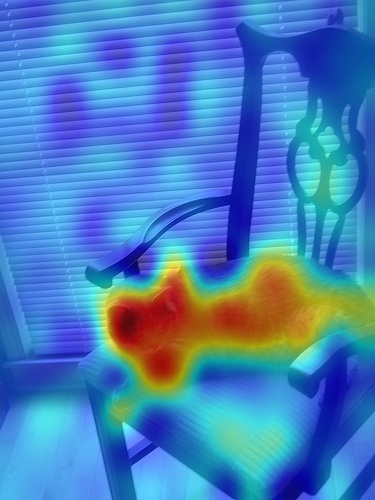} \\
  
  \includegraphics[width=3.3cm]{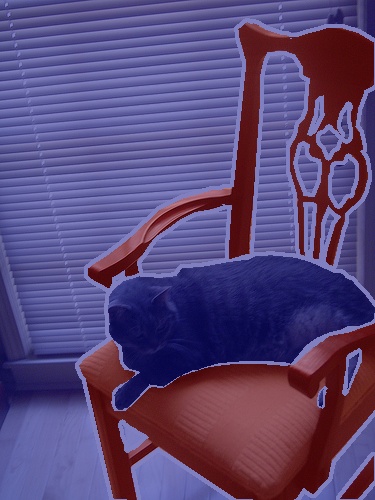} \includegraphics[width=3.3cm]{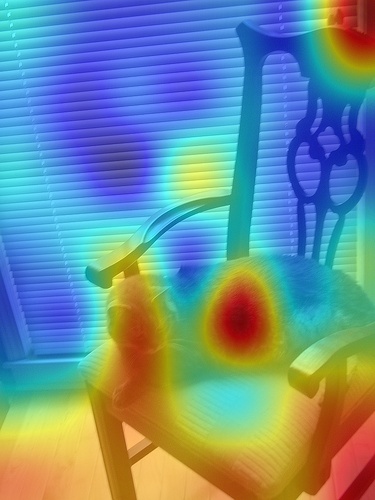}   \includegraphics[width=3.3cm]{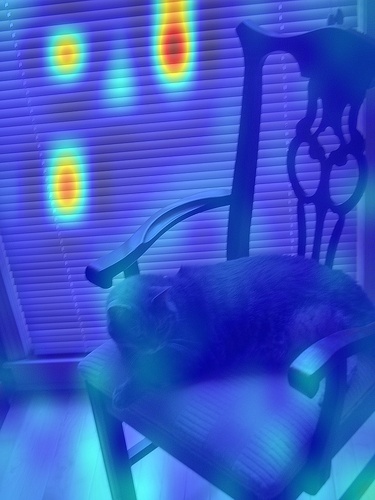} \includegraphics[width=3.3cm]{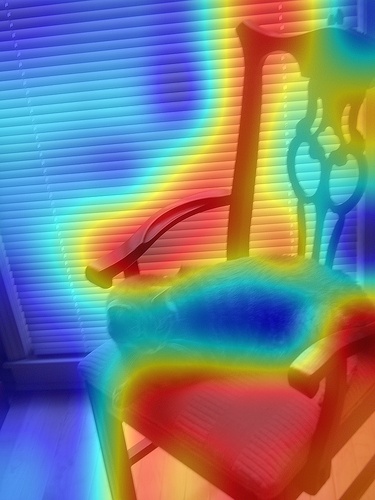} \includegraphics[width=3.3cm]{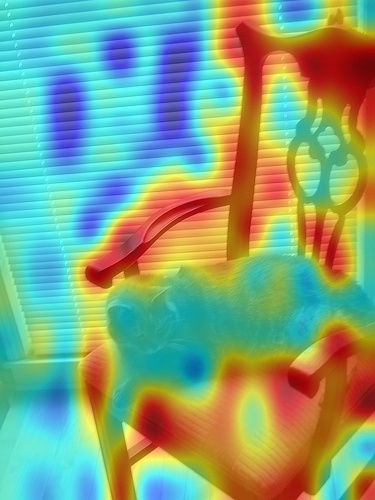} \\

  \includegraphics[width=3.3cm]{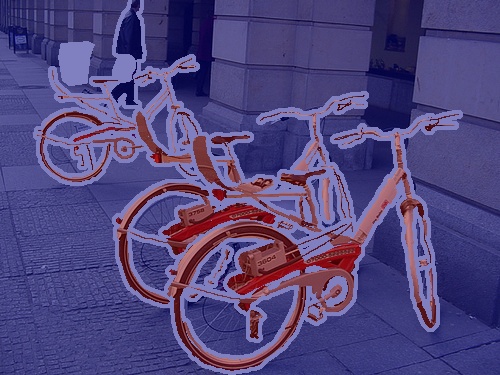} \includegraphics[width=3.3cm]{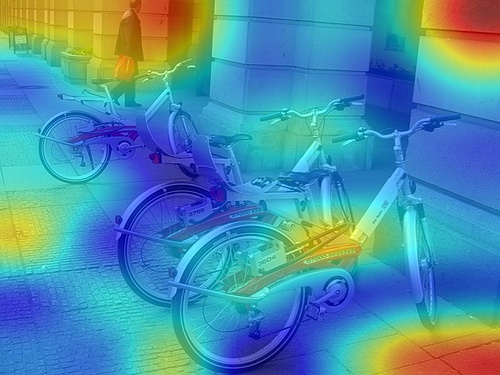}   \includegraphics[width=3.3cm]{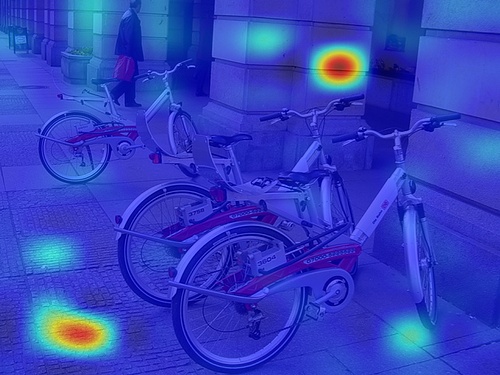} \includegraphics[width=3.3cm]{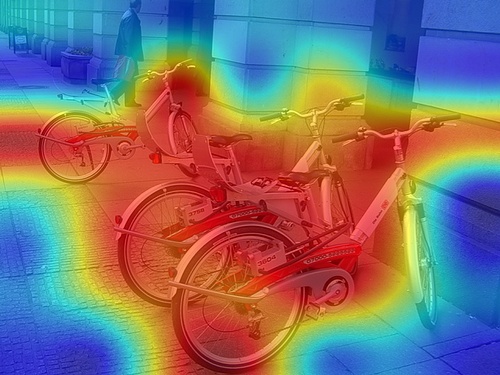} \includegraphics[width=3.3cm]{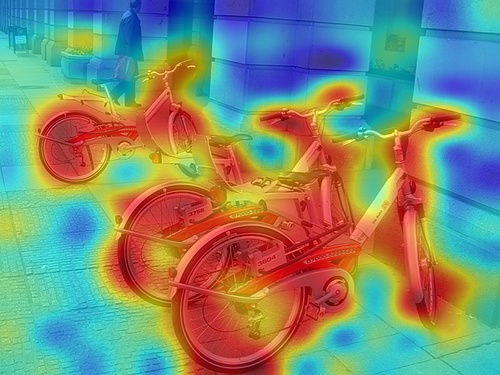} \\
  
  \includegraphics[width=3.3cm]{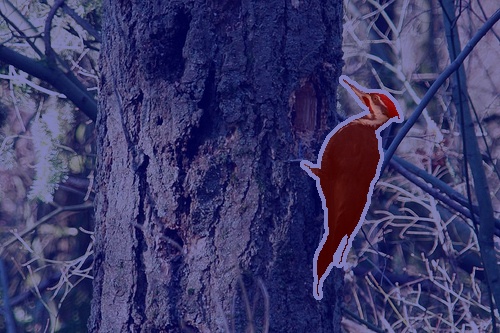} \includegraphics[width=3.3cm]{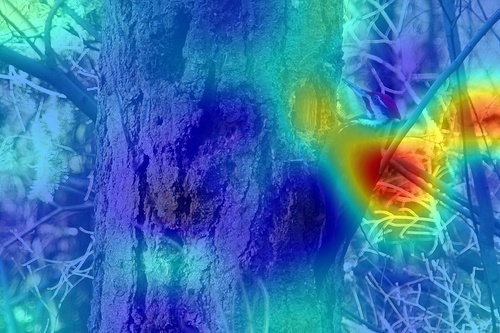}   \includegraphics[width=3.3cm]{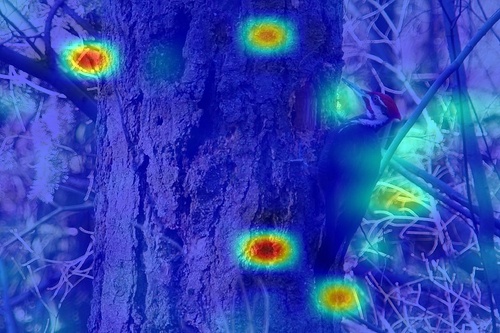} \includegraphics[width=3.3cm]{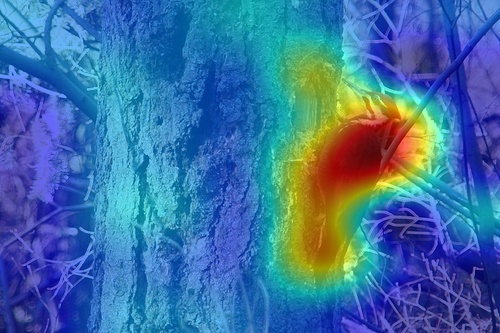} \includegraphics[width=3.3cm]{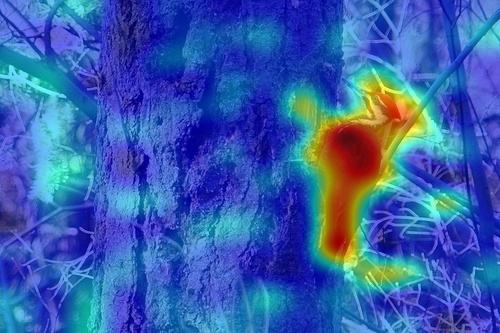} \\
  
  \includegraphics[width=3.3cm,height=3.3cm]{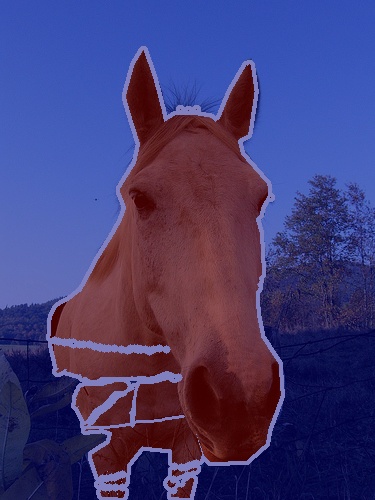} \includegraphics[width=3.3cm,height=3.3cm]{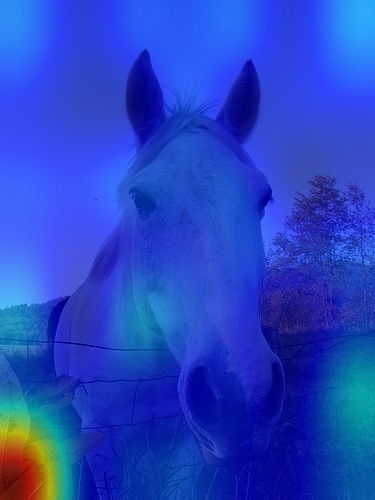}   \includegraphics[width=3.3cm,height=3.3cm]{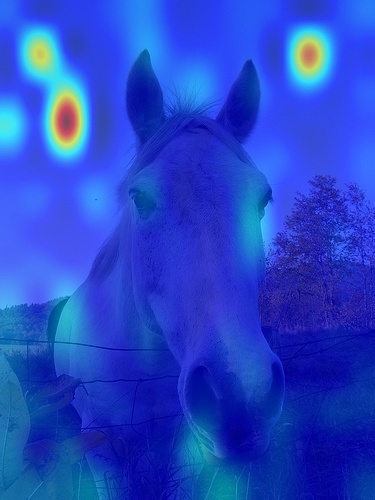} \includegraphics[width=3.3cm,height=3.3cm]{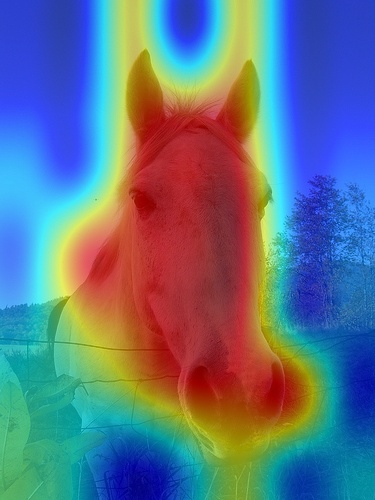} \includegraphics[width=3.3cm,height=3.3cm]{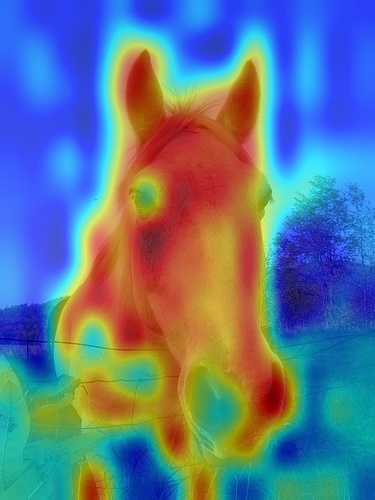} \\
  
\caption{Visualization examples on VOC 2012 dataset \cite{everingham2010pascal}. Bi-Model \cite{chefer2021generic} is highly limited at output side, and our ECLIP breaks this limitation. Note, output side $\times$7 comes from ViT-B/32 and $\times$14 from ViT-B/16, besides, predicted foreground is colored in red.}
\label{vis_side}
\end{figure*}

As discussed above, the output side is related to the visualization result. And we find the impact of output side is much serious in Bi-Model \cite{chefer2021generic}. As shown in Fig. \ref{vis_side}, our ECLIP is much less influenced by the output side, and larger network (ViT-B/16) show detailed results and beyond ViT-B/32 for some cases. But Bi-Model meets serious performance damage, and shows punctate heatmap. Besides, Bi-Model is confused at multiple classes, especially for larger output side. The visualization results are almost the same for different categories, when the output side is scaled up, which shows the limitation of output side. And our method breaks this limitation.

\section{Additional Visualization Results}

\begin{figure*}[t]
\leftline{\ \ \ \ \ \ \ \ Ground Truth   \ \ \ \ \ \ \ \ \  \ \ \ \ \ \ CLIP \ \ \ \ \ \ \ \ \ \ \ \ \ \ \ Grad-CAM \cite{selvaraju2017grad} \ \ \ \ \ \ \ \ \ \  Bi-Model\cite{chefer2021generic} \ \ \ \ \ \ \ \ \ \ \ \ \ \ RCLIP   \ \ \ \ \ \ \ \ \ \ \ \ \ \ \ \ \ \ \ ECLIP}
\vspace{2mm}
\centering
  \includegraphics[width=2.75cm]{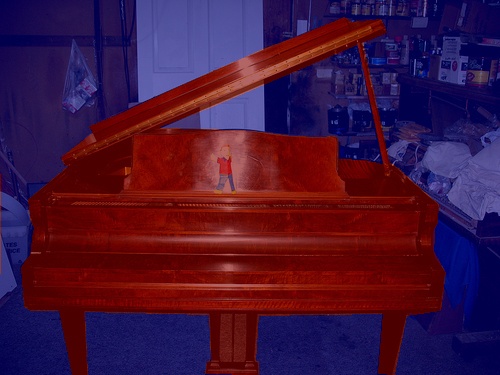} \includegraphics[width=2.75cm]{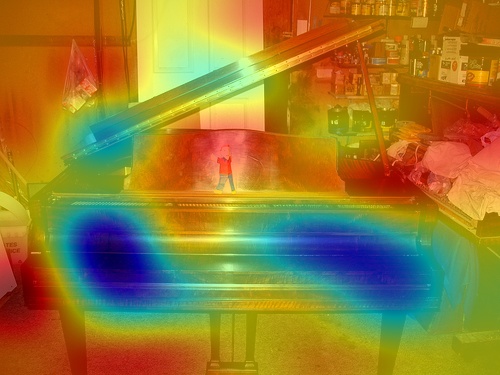}   \includegraphics[width=2.75Cm]{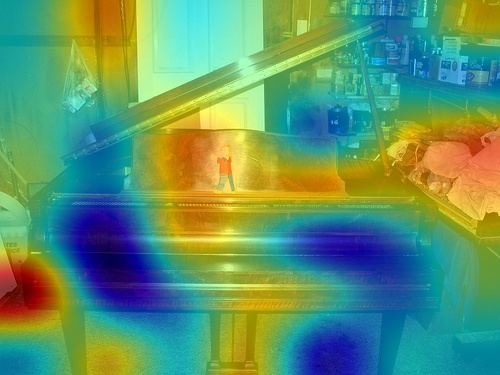} \includegraphics[width=2.75cm]{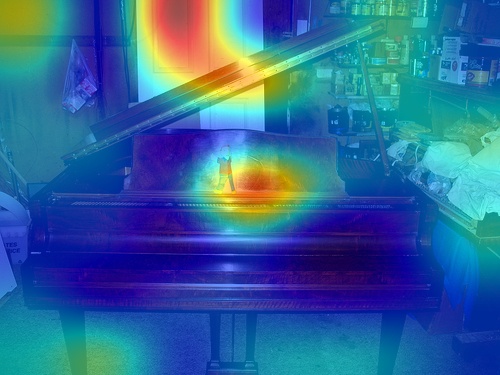} \includegraphics[width=2.75cm]{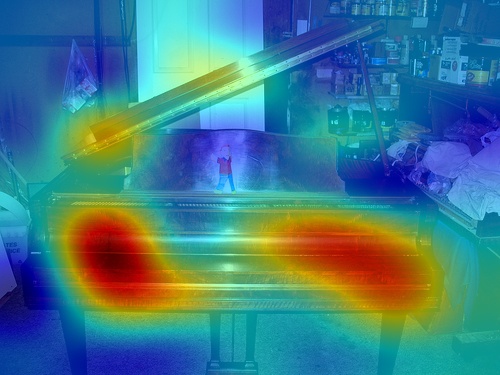} \includegraphics[width=2.75cm]{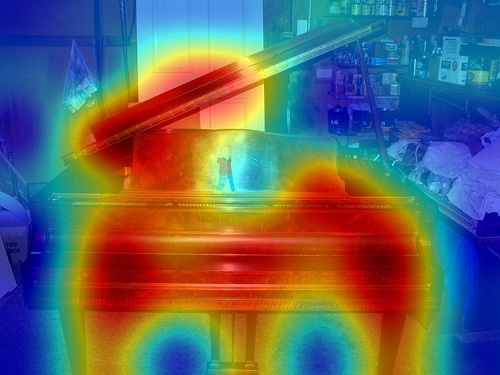} \\
  
  \includegraphics[width=2.75cm]{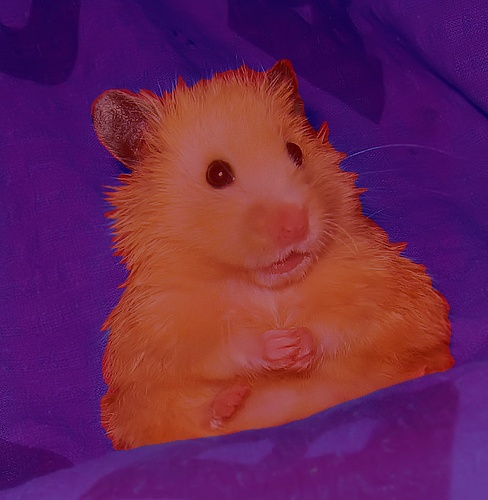} \includegraphics[width=2.75cm]{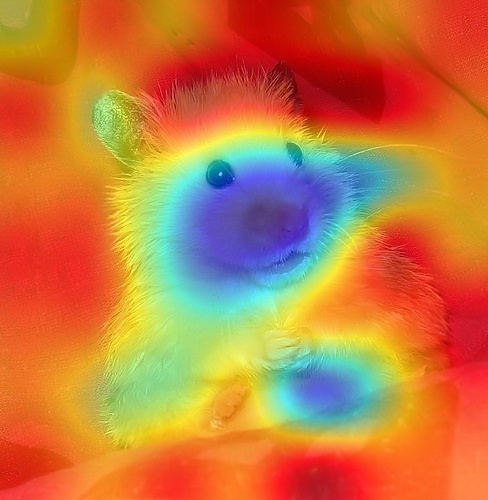}   \includegraphics[width=2.75Cm]{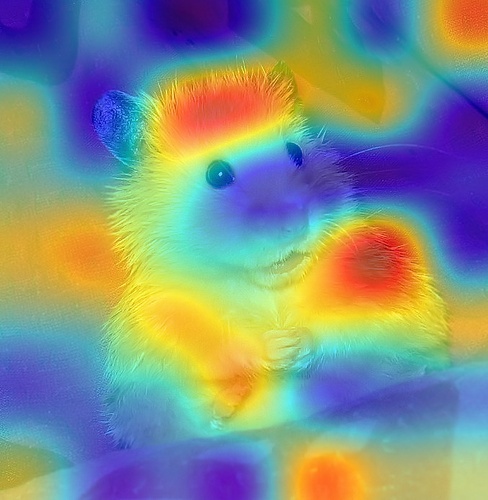} \includegraphics[width=2.75cm]{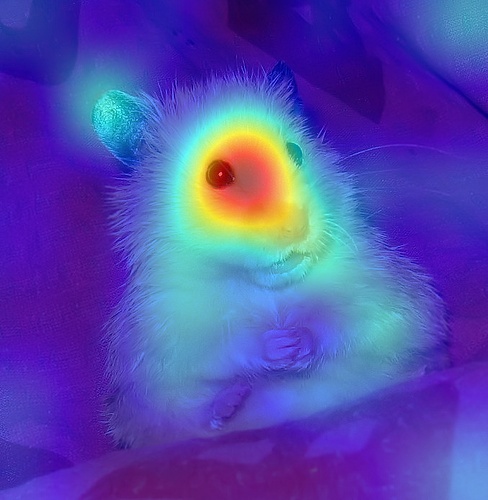} \includegraphics[width=2.75cm]{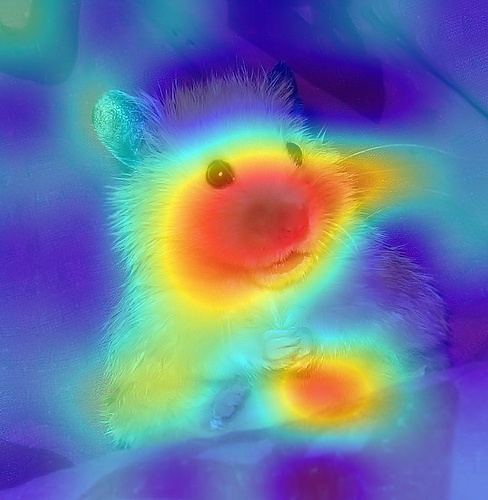} \includegraphics[width=2.75cm]{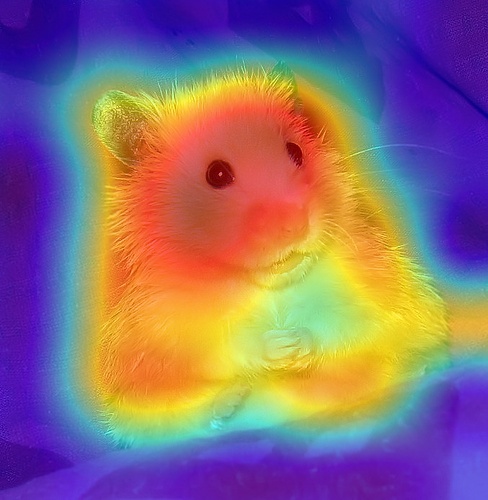} \\
  
  \includegraphics[width=2.75cm]{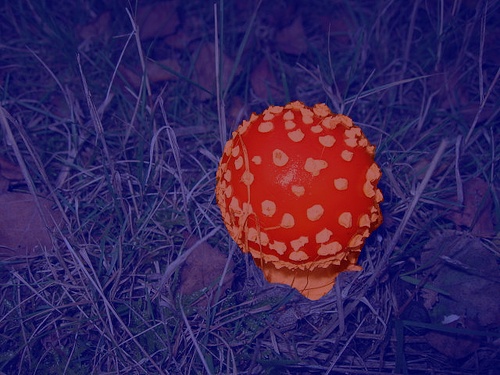} \includegraphics[width=2.75cm]{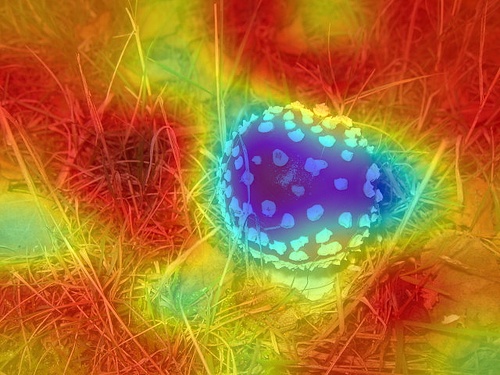}   \includegraphics[width=2.75Cm]{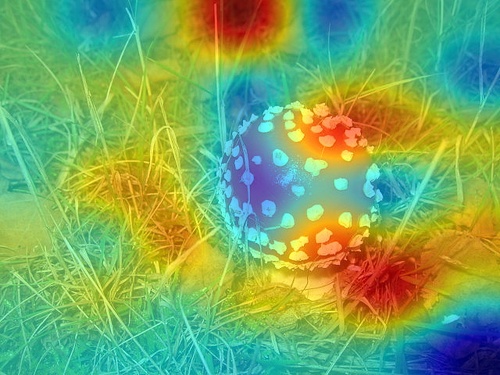} \includegraphics[width=2.75cm]{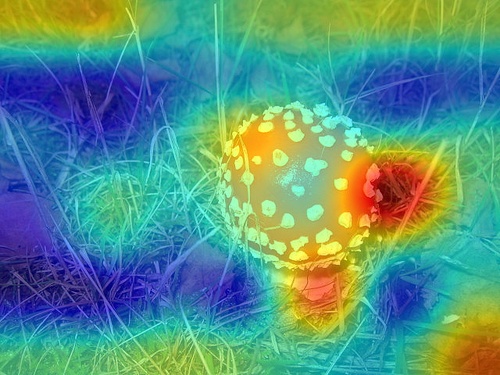} \includegraphics[width=2.75cm]{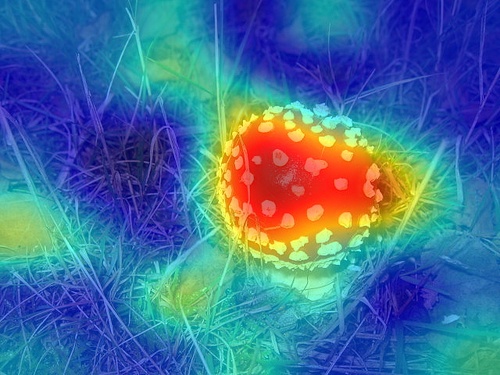} \includegraphics[width=2.75cm]{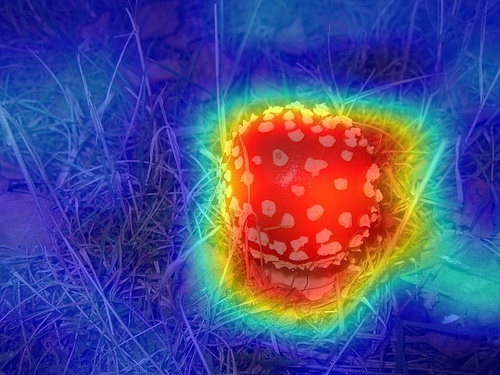} \\

  \includegraphics[width=2.75cm]{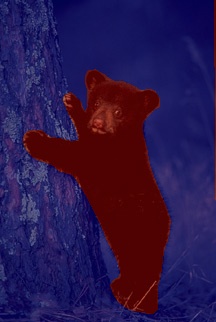} \includegraphics[width=2.75cm]{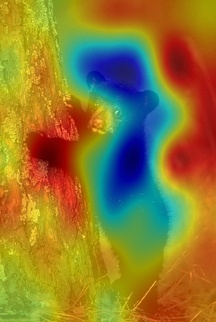}   \includegraphics[width=2.75Cm]{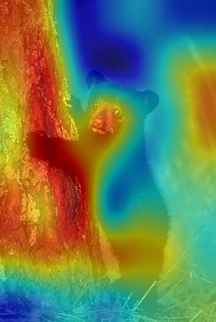} \includegraphics[width=2.75cm]{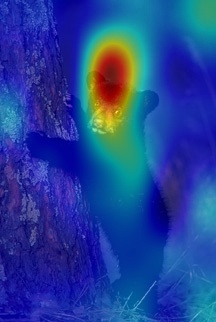} \includegraphics[width=2.75cm]{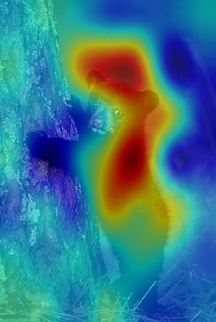} \includegraphics[width=2.75cm]{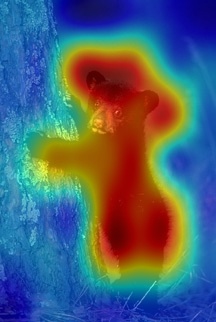} \\
  
  \includegraphics[width=2.75cm]{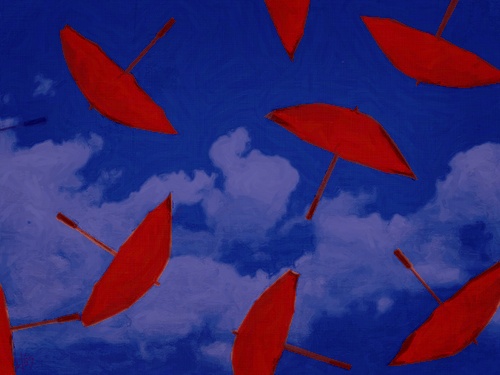} \includegraphics[width=2.75cm]{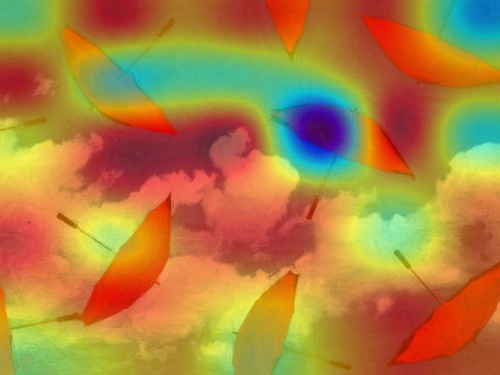}   \includegraphics[width=2.75Cm]{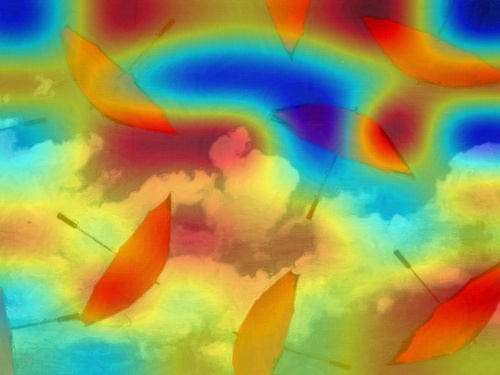} \includegraphics[width=2.75cm]{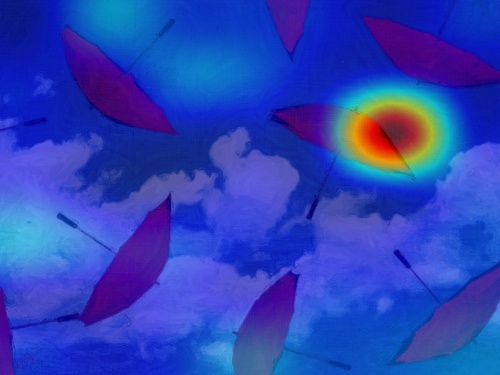} \includegraphics[width=2.75cm]{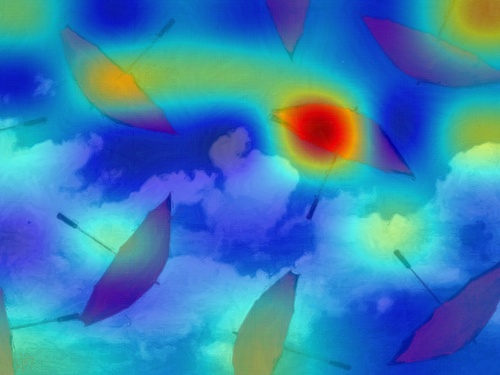} \includegraphics[width=2.75cm]{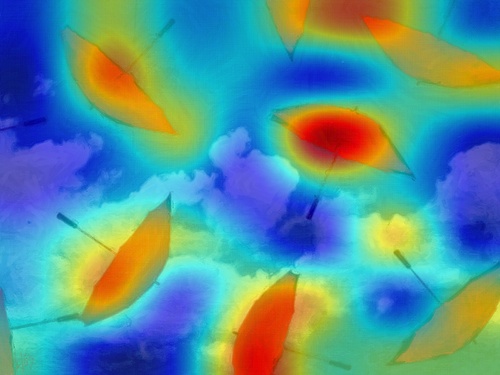} \\
  
  \includegraphics[width=2.75cm]{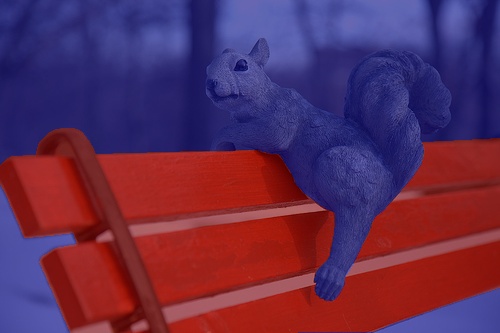} \includegraphics[width=2.75cm]{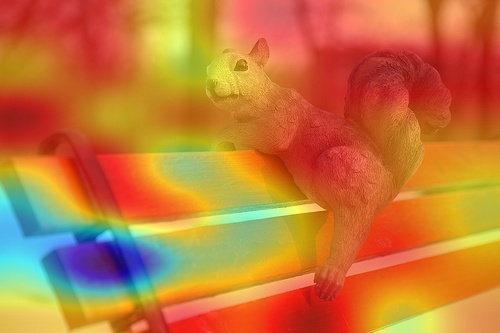}   \includegraphics[width=2.75Cm]{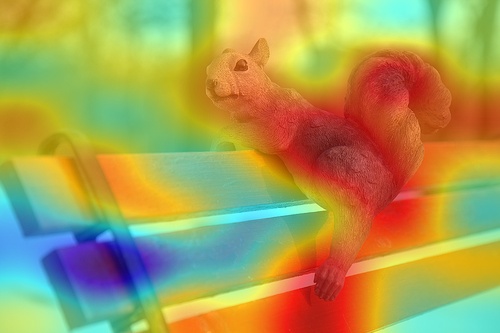} \includegraphics[width=2.75cm]{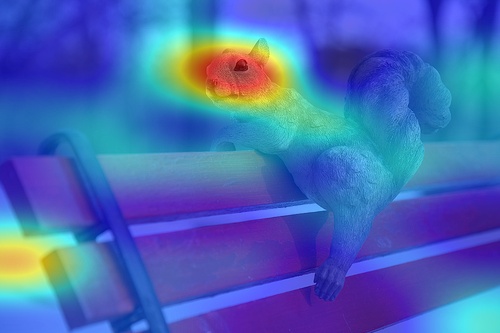} \includegraphics[width=2.75cm]{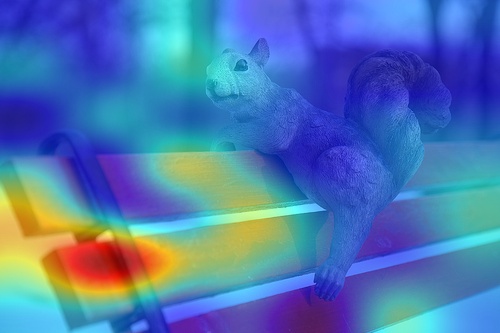} \includegraphics[width=2.75cm]{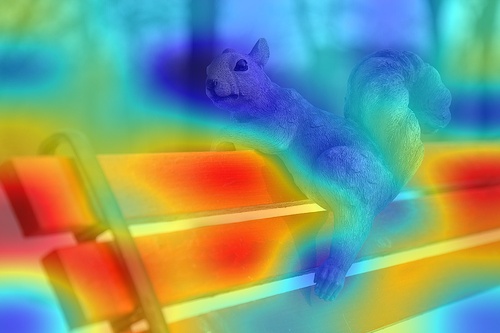} \\
  
  \includegraphics[width=2.75cm]{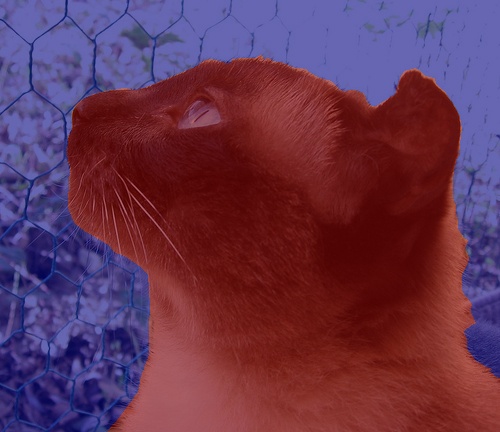} \includegraphics[width=2.75cm]{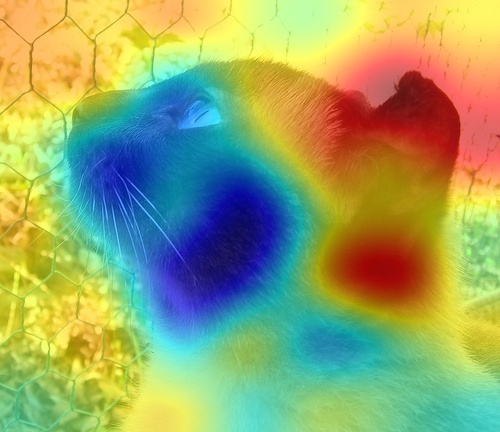}   \includegraphics[width=2.75Cm]{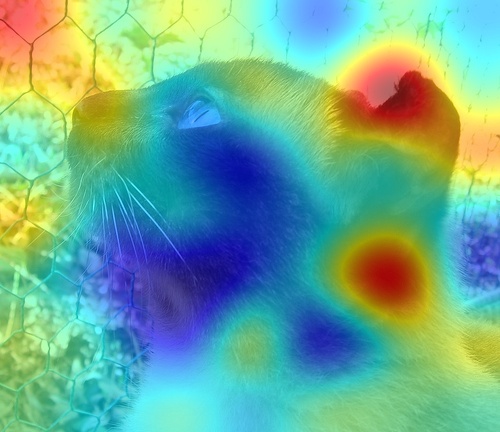} \includegraphics[width=2.75cm]{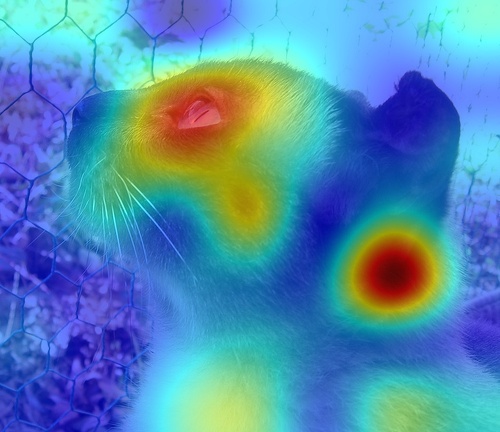} \includegraphics[width=2.75cm]{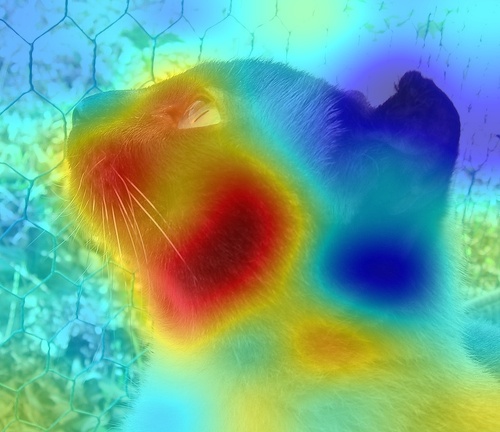} \includegraphics[width=2.75cm]{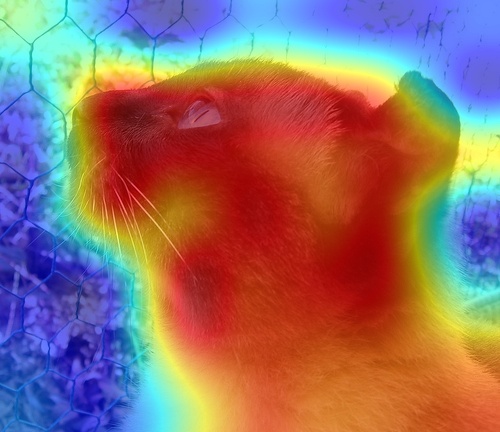} \\
\caption{Visualization comparison between conventional methods and ours (RCLIP, ECLIP) on ImageNetS-50 \cite{gao2022large} with ViT-B/32.}
\label{vis_imagenet}
\end{figure*}

\begin{figure*}[!t]
\leftline{\ \ \ \ \ \ \ \ Ground Truth   \ \ \ \ \ \ \ \ \  \ \ \ \ \ \ CLIP \ \ \ \ \ \ \ \ \ \ \ \ \ \ \ Grad-CAM \cite{selvaraju2017grad} \ \ \ \ \ \ \ \ \ \  Bi-Model\cite{chefer2021generic} \ \ \ \ \ \ \ \ \ \ \ \ \ \ RCLIP   \ \ \ \ \ \ \ \ \ \ \ \ \ \ \ \ \ \ \ ECLIP}
\vspace{2mm}
\centering
  \includegraphics[width=2.75cm]{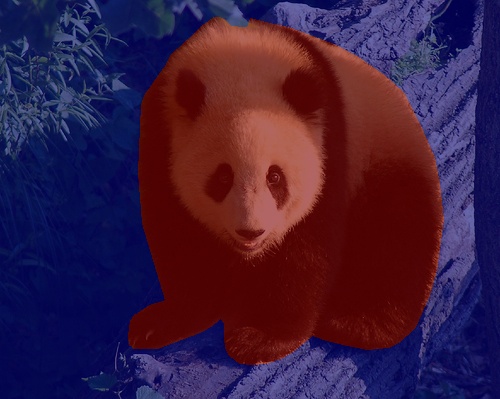} \includegraphics[width=2.75cm]{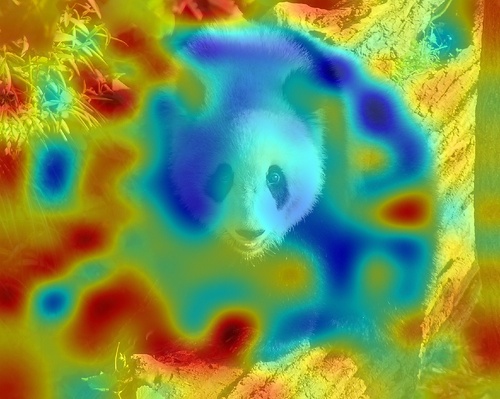}   \includegraphics[width=2.75Cm]{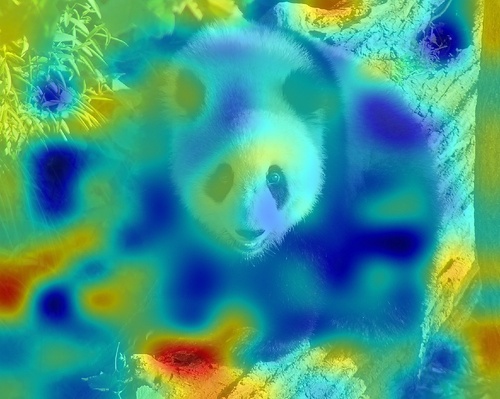} \includegraphics[width=2.75cm]{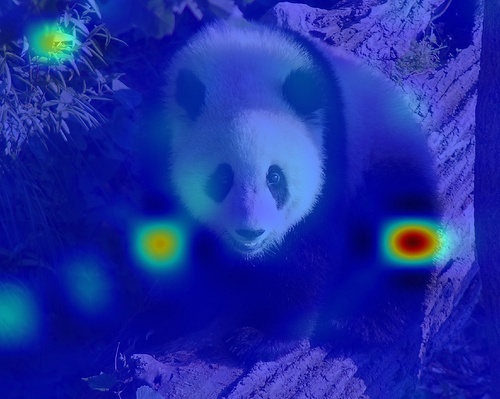} \includegraphics[width=2.75cm]{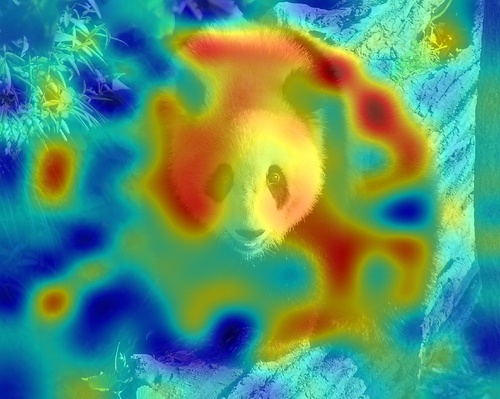} \includegraphics[width=2.75cm]{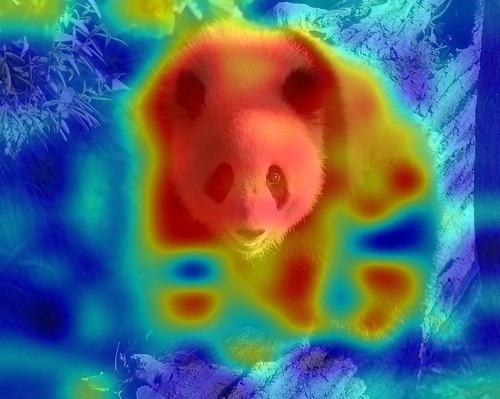} \\
  
  \includegraphics[width=2.75cm]{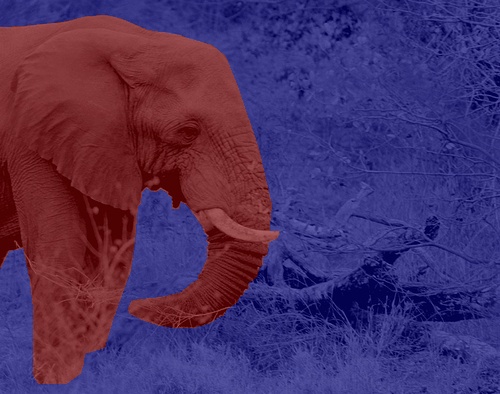} \includegraphics[width=2.75cm]{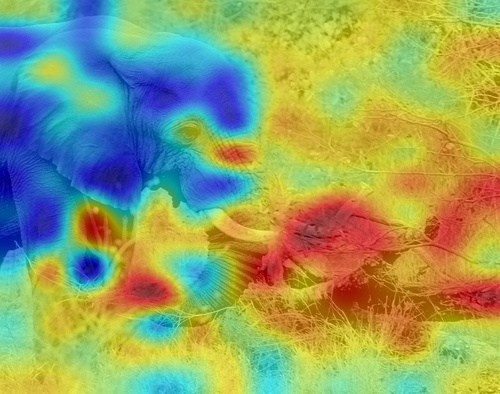}   \includegraphics[width=2.75Cm]{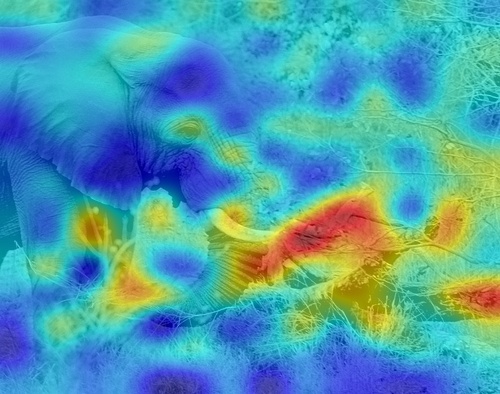} \includegraphics[width=2.75cm]{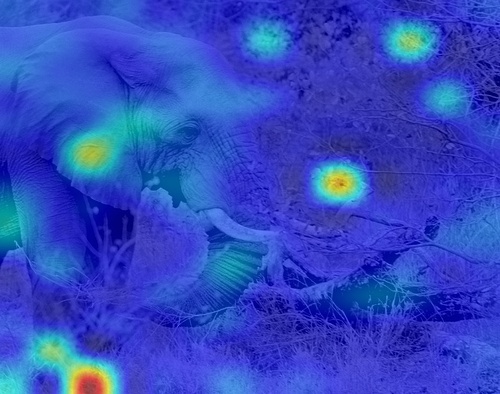} \includegraphics[width=2.75cm]{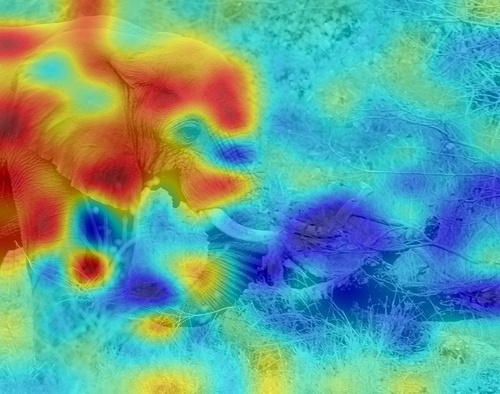} \includegraphics[width=2.75cm]{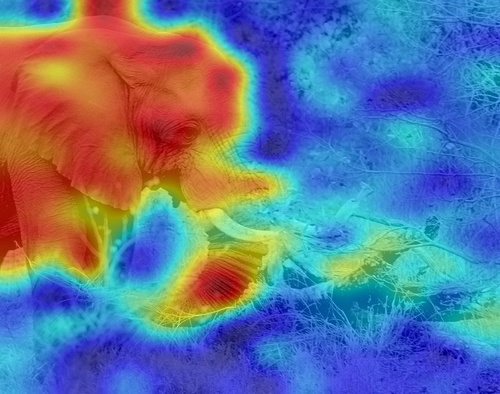} \\
  
  \includegraphics[width=2.75cm]{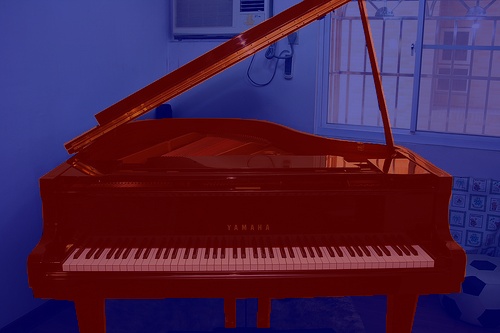} \includegraphics[width=2.75cm]{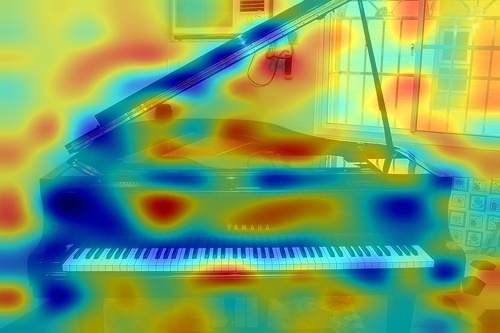}   \includegraphics[width=2.75Cm]{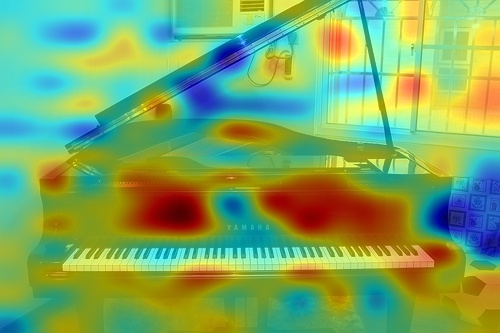} \includegraphics[width=2.75cm]{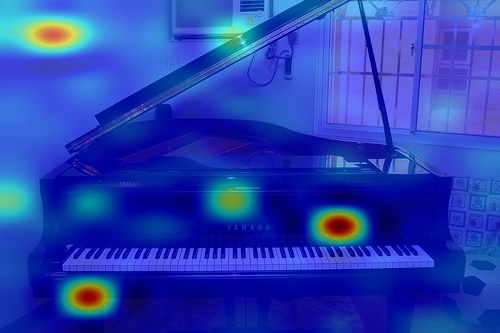} \includegraphics[width=2.75cm]{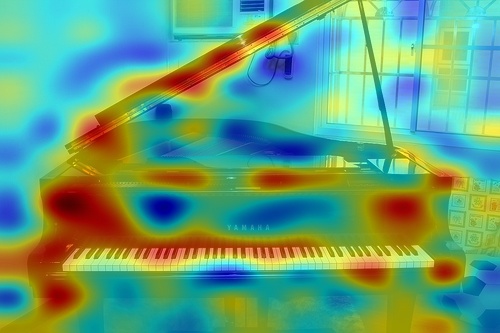} \includegraphics[width=2.75cm]{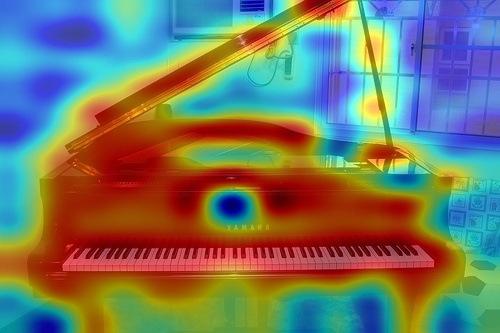} \\

  \includegraphics[width=2.75cm]{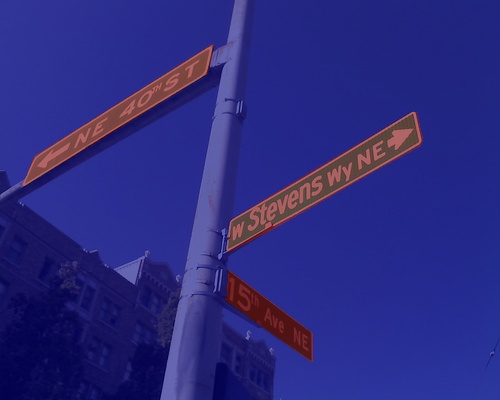} \includegraphics[width=2.75cm]{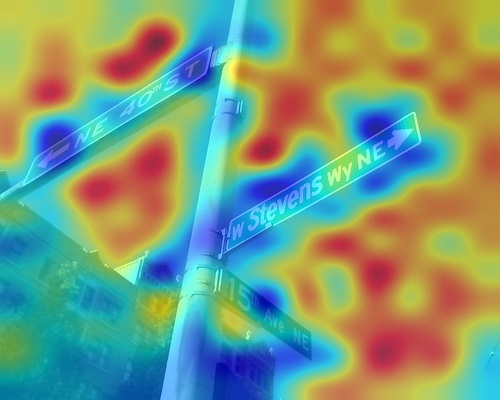}   \includegraphics[width=2.75Cm]{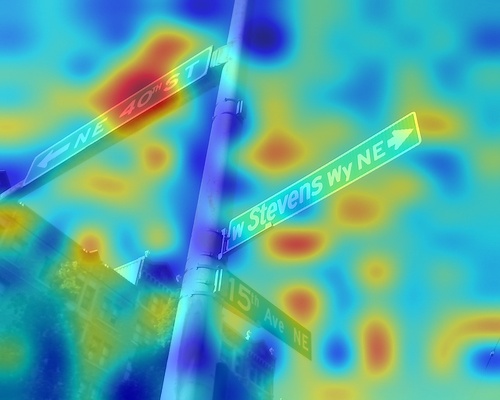} \includegraphics[width=2.75cm]{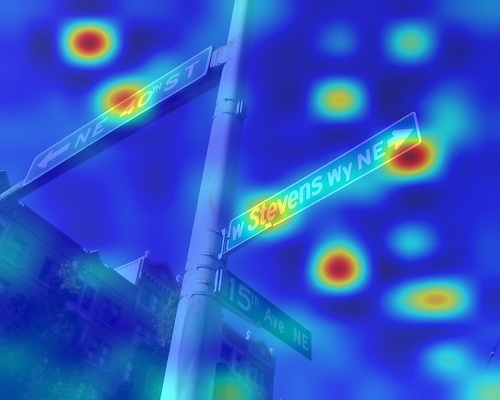} \includegraphics[width=2.75cm]{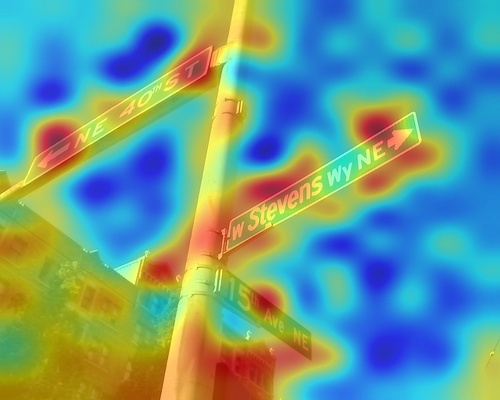} \includegraphics[width=2.75cm]{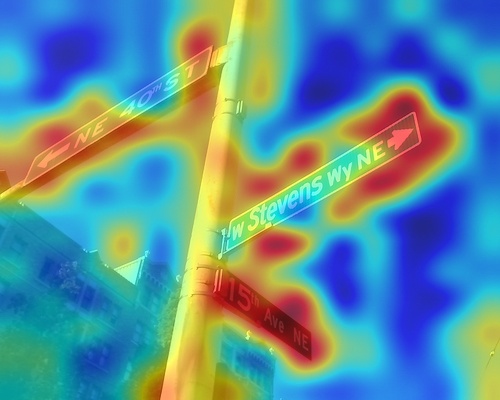} \\
  
  \includegraphics[width=2.75cm]{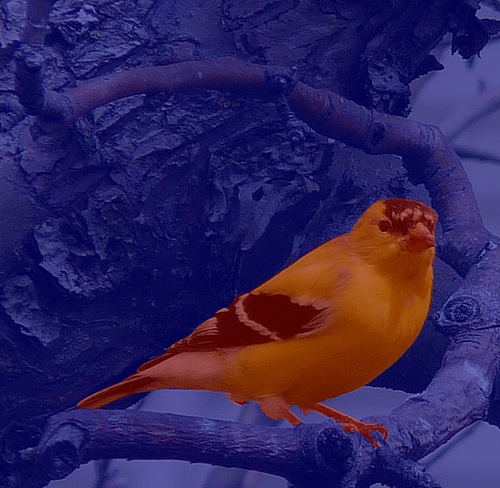} \includegraphics[width=2.75cm]{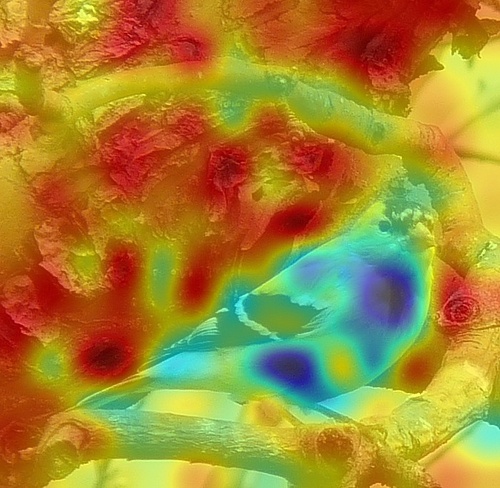}   \includegraphics[width=2.75Cm]{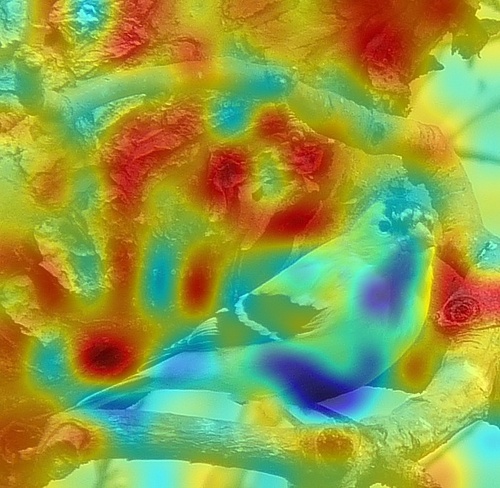} \includegraphics[width=2.75cm]{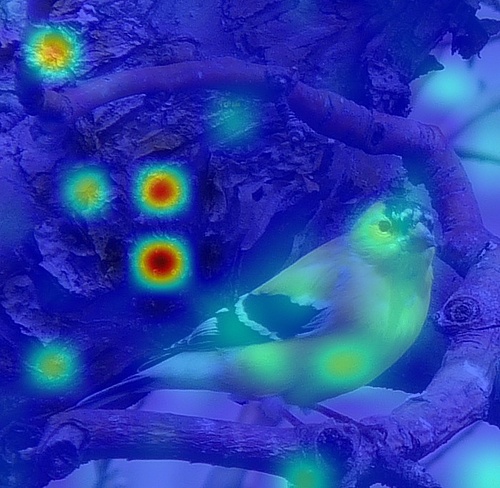} \includegraphics[width=2.75cm]{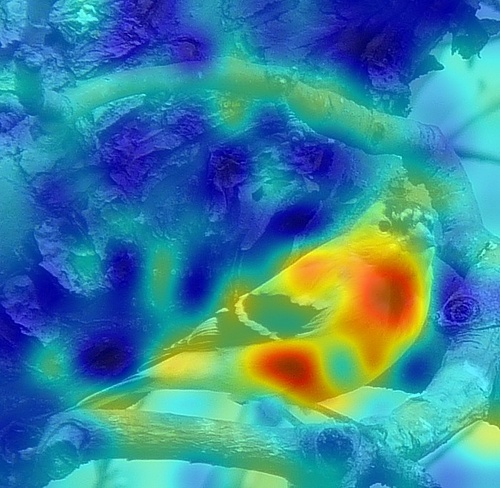} \includegraphics[width=2.75cm]{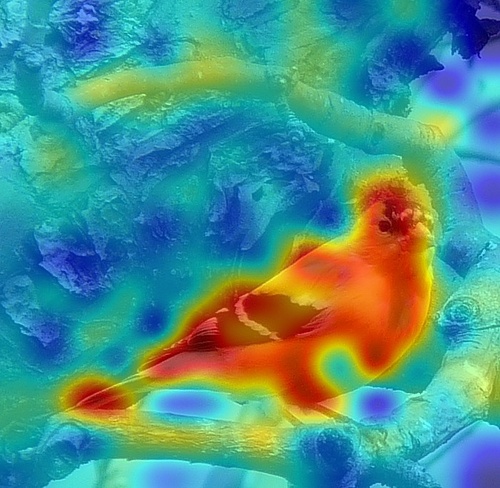} \\
  
  \includegraphics[width=2.75cm]{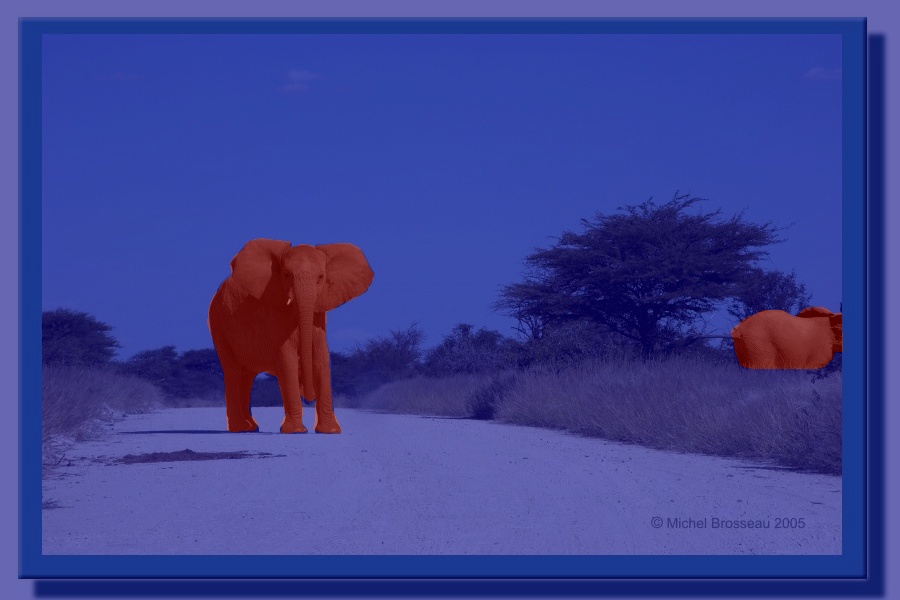} \includegraphics[width=2.75cm]{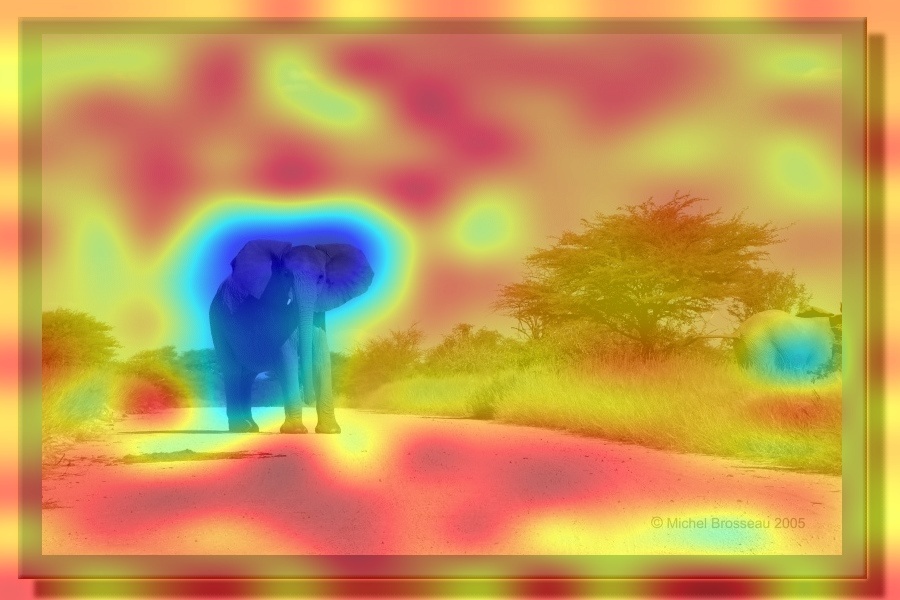}   \includegraphics[width=2.75Cm]{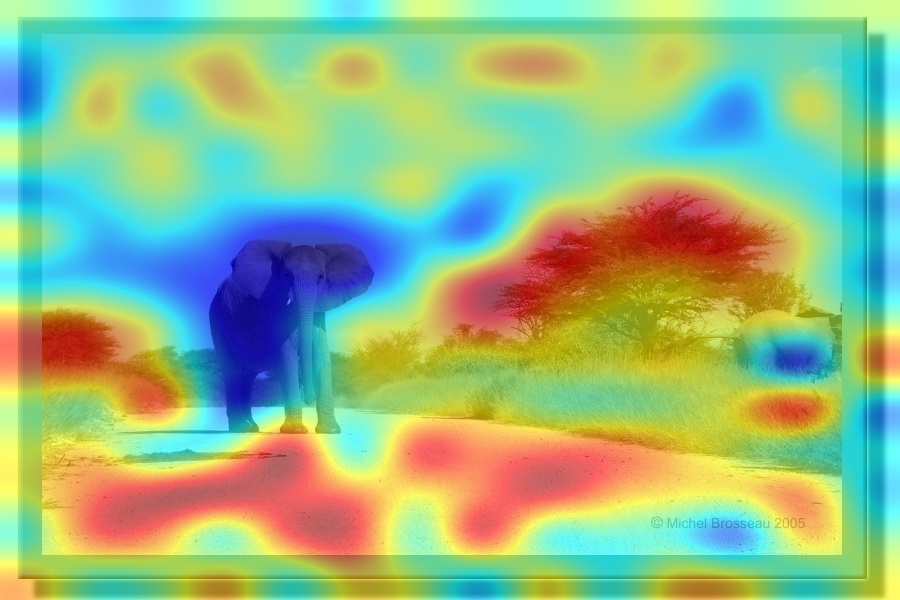} \includegraphics[width=2.75cm]{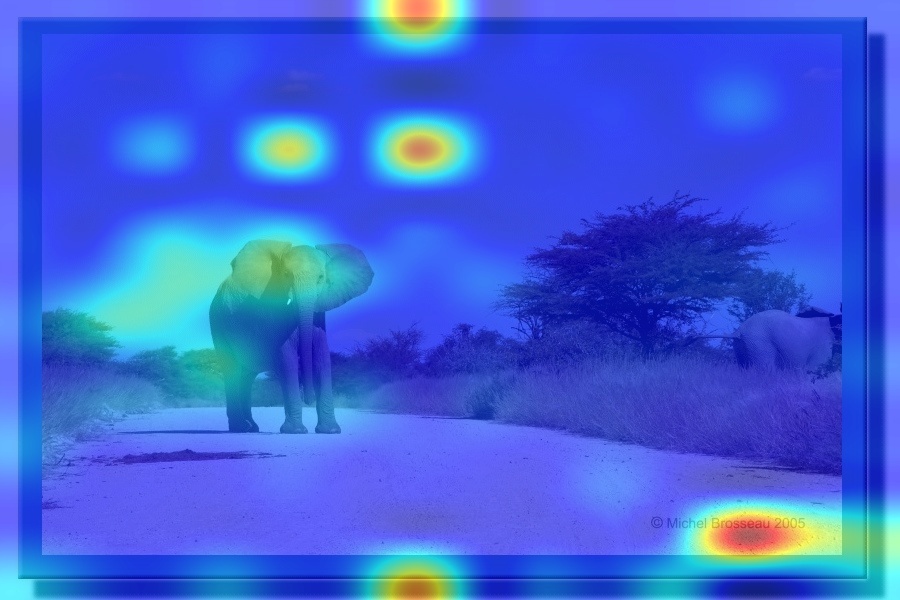} \includegraphics[width=2.75cm]{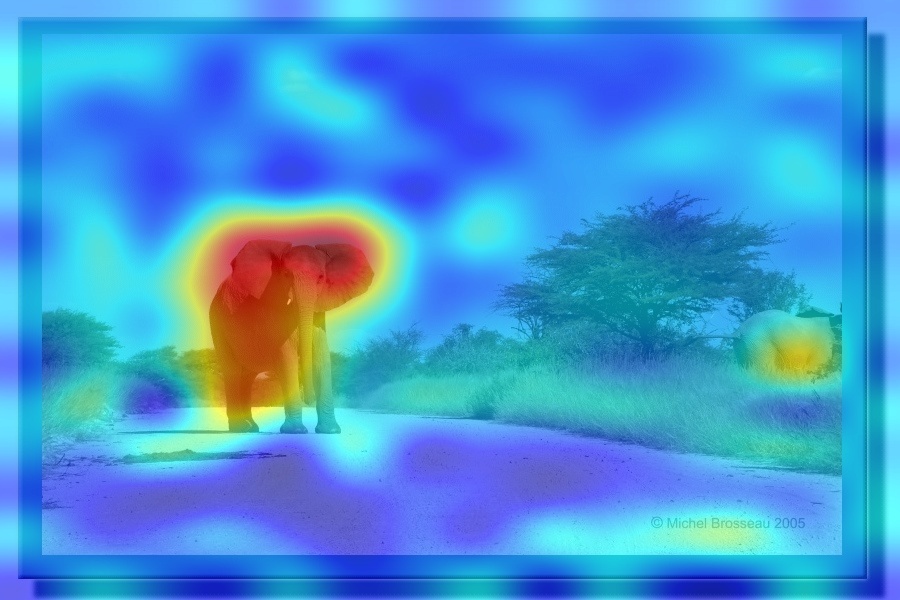} \includegraphics[width=2.75cm]{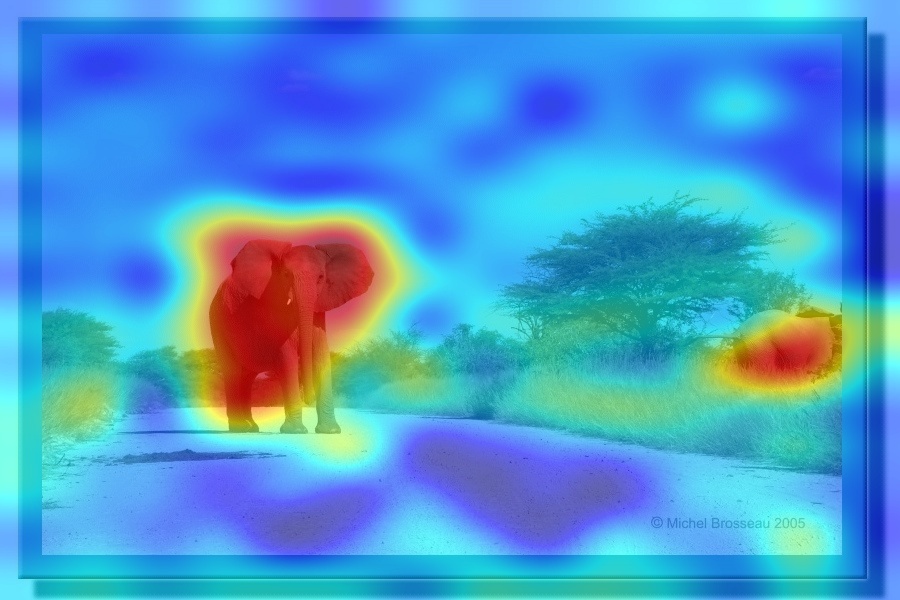} \\
  
\caption{Visualization comparison between conventional methods and ours (RCLIP, ECLIP) on ImageNetS-50 \cite{gao2022large} for ViT-B/16.}
\label{vis_imagenet_16}
\end{figure*}

Besides the visual comparison on VOC \cite{everingham2010pascal} in the main paper, we set additional comparison between conventional methods and ours as Fig. \ref{vis_imagenet} on ImageNetS-50 (ImageNet-Segmentation-50) \cite{gao2022large}. These results are based on ViT-B/32 at output size $\times7$, which suits Bi-Model best \cite{chefer2021generic}. Even so, the latest transformer based explanation methods for CLIP, Bi-Model, performs much worse than our RCLIP and ECLIP. And the most used gradient based method, Grad-CAM \cite{selvaraju2017grad}, behaves like the original CLIP, shown opposite visualization results. Then we draw the results on ViT-B/16 in Fig. \ref{vis_imagenet_16}. In the condition of dense prediction, our methods outperform conventional methods with larger margins.

\ \\
\ \\
\ \\
\ \\
\ \\
\ \\
\ \\
\ \\
\ \\
\ \\
\ \\
\ \\
\ \\
\ \\
\ \\
\ \\
\ \\
\ \\
\ \\
\ \\

\end{document}

%% file: math.tex
\usepackage{amsmath,amsfonts,bm}

\def\eqref#1{equation~\ref{#1}}

\def\1{\bm{1}}

\def\mA{{\bm{A}}}

\def\mF{{\bm{F}}}

\def\mM{{\bm{M}}}

\DeclareMathAlphabet{\mathsfit}{\encodingdefault}{\sfdefault}{m}{sl}
\SetMathAlphabet{\mathsfit}{bold}{\encodingdefault}{\sfdefault}{bx}{n}
\newcommand{\tens}[1]{\bm{\mathsfit{#1}}}

\def\tM{{\tens{M}}}

\newcommand{\R}{\mathbb{R}}



%% file: arXiv.bbl
\begin{thebibliography}{10}\itemsep=-1pt

\bibitem{abnar2020quantifying}
Samira Abnar and Willem Zuidema.
\newblock Quantifying attention flow in transformers.
\newblock {\em arXiv preprint arXiv:2005.00928}, 2020.

\bibitem{antol2015vqa}
Stanislaw Antol, Aishwarya Agrawal, Jiasen Lu, Margaret Mitchell, Dhruv Batra,
  C~Lawrence Zitnick, and Devi Parikh.
\newblock Vqa: Visual question answering.
\newblock In {\em Proceedings of the IEEE international conference on computer
  vision}, pages 2425--2433, 2015.

\bibitem{azizi2021big}
Shekoofeh Azizi, Basil Mustafa, Fiona Ryan, Zachary Beaver, Jan Freyberg,
  Jonathan Deaton, Aaron Loh, Alan Karthikesalingam, Simon Kornblith, Ting
  Chen, et~al.
\newblock Big self-supervised models advance medical image classification.
\newblock In {\em Proceedings of the IEEE/CVF International Conference on
  Computer Vision}, pages 3478--3488, 2021.

\bibitem{bao2021beit}
Hangbo Bao, Li Dong, and Furu Wei.
\newblock Beit: Bert pre-training of image transformers.
\newblock {\em arXiv preprint arXiv:2106.08254}, 2021.

\bibitem{caron2021emerging}
Mathilde Caron, Hugo Touvron, Ishan Misra, Herv{\'e} J{\'e}gou, Julien Mairal,
  Piotr Bojanowski, and Armand Joulin.
\newblock Emerging properties in self-supervised vision transformers.
\newblock In {\em Proceedings of the IEEE/CVF International Conference on
  Computer Vision}, pages 9650--9660, 2021.

\bibitem{cha2022domain}
Junbum Cha, Kyungjae Lee, Sungrae Park, and Sanghyuk Chun.
\newblock Domain generalization by mutual-information regularization with
  pre-trained models.
\newblock {\em arXiv preprint arXiv:2203.10789}, 2022.

\bibitem{changpinyo2021conceptual}
Soravit Changpinyo, Piyush Sharma, Nan Ding, and Radu Soricut.
\newblock Conceptual 12m: Pushing web-scale image-text pre-training to
  recognize long-tail visual concepts.
\newblock In {\em Proceedings of the IEEE/CVF Conference on Computer Vision and
  Pattern Recognition}, pages 3558--3568, 2021.

\bibitem{chattopadhay2018grad}
Aditya Chattopadhay, Anirban Sarkar, Prantik Howlader, and Vineeth~N
  Balasubramanian.
\newblock Grad-cam++: Generalized gradient-based visual explanations for deep
  convolutional networks.
\newblock In {\em 2018 IEEE winter conference on applications of computer
  vision (WACV)}, pages 839--847. IEEE, 2018.

\bibitem{chefer2021generic}
Hila Chefer, Shir Gur, and Lior Wolf.
\newblock Generic attention-model explainability for interpreting bi-modal and
  encoder-decoder transformers.
\newblock In {\em Proceedings of the IEEE/CVF International Conference on
  Computer Vision}, pages 397--406, 2021.

\bibitem{chefer2021transformer}
Hila Chefer, Shir Gur, and Lior Wolf.
\newblock Transformer interpretability beyond attention visualization.
\newblock In {\em Proceedings of the IEEE/CVF Conference on Computer Vision and
  Pattern Recognition}, pages 782--791, 2021.

\bibitem{chen2021empirical}
Xinlei Chen, Saining Xie, and Kaiming He.
\newblock An empirical study of training self-supervised vision transformers.
\newblock In {\em Proceedings of the IEEE/CVF International Conference on
  Computer Vision}, pages 9640--9649, 2021.

\bibitem{deng2009imagenet}
Jia Deng, Wei Dong, Richard Socher, Li-Jia Li, Kai Li, and Li Fei-Fei.
\newblock Imagenet: A large-scale hierarchical image database.
\newblock In {\em 2009 IEEE conference on computer vision and pattern
  recognition}, pages 248--255. Ieee, 2009.

\bibitem{doersch2015unsupervised}
Carl Doersch, Abhinav Gupta, and Alexei~A Efros.
\newblock Unsupervised visual representation learning by context prediction.
\newblock In {\em Proceedings of the IEEE international conference on computer
  vision}, pages 1422--1430, 2015.

\bibitem{dosovitskiy2020image}
Alexey Dosovitskiy, Lucas Beyer, Alexander Kolesnikov, Dirk Weissenborn,
  Xiaohua Zhai, Thomas Unterthiner, Mostafa Dehghani, Matthias Minderer, Georg
  Heigold, Sylvain Gelly, et~al.
\newblock An image is worth 16x16 words: Transformers for image recognition at
  scale.
\newblock {\em arXiv preprint arXiv:2010.11929}, 2020.

\bibitem{everingham2010pascal}
Mark Everingham, Luc Van~Gool, Christopher~KI Williams, John Winn, and Andrew
  Zisserman.
\newblock The pascal visual object classes (voc) challenge.
\newblock {\em International journal of computer vision}, 88(2):303--338, 2010.

\bibitem{gao2021clip}
Peng Gao, Shijie Geng, Renrui Zhang, Teli Ma, Rongyao Fang, Yongfeng Zhang,
  Hongsheng Li, and Yu Qiao.
\newblock Clip-adapter: Better vision-language models with feature adapters.
\newblock {\em arXiv preprint arXiv:2110.04544}, 2021.

\bibitem{gao2022large}
Shanghua Gao, Zhong-Yu Li, Ming-Hsuan Yang, Ming-Ming Cheng, Junwei Han, and
  Philip Torr.
\newblock Large-scale unsupervised semantic segmentation.
\newblock {\em IEEE Transactions on Pattern Analysis and Machine Intelligence},
  2022.

\bibitem{goh2021multimodal}
Gabriel Goh, Nick Cammarata, Chelsea Voss, Shan Carter, Michael Petrov, Ludwig
  Schubert, Alec Radford, and Chris Olah.
\newblock Multimodal neurons in artificial neural networks.
\newblock {\em Distill}, 6(3):e30, 2021.

\bibitem{gu2022wukong}
Jiaxi Gu, Xiaojun Meng, Guansong Lu, Lu Hou, Minzhe Niu, Hang Xu, Xiaodan
  Liang, Wei Zhang, Xin Jiang, and Chunjing Xu.
\newblock Wukong: 100 million large-scale chinese cross-modal pre-training
  dataset and a foundation framework.
\newblock {\em arXiv preprint arXiv:2202.06767}, 2022.

\bibitem{he2016deep}
Kaiming He, Xiangyu Zhang, Shaoqing Ren, and Jian Sun.
\newblock Deep residual learning for image recognition.
\newblock In {\em Proceedings of the IEEE conference on computer vision and
  pattern recognition}, pages 770--778, 2016.

\bibitem{jaiswal2020survey}
Ashish Jaiswal, Ashwin~Ramesh Babu, Mohammad~Zaki Zadeh, Debapriya Banerjee,
  and Fillia Makedon.
\newblock A survey on contrastive self-supervised learning.
\newblock {\em Technologies}, 9(1):2, 2020.

\bibitem{lei2021less}
Jie Lei, Linjie Li, Luowei Zhou, Zhe Gan, Tamara~L Berg, Mohit Bansal, and
  Jingjing Liu.
\newblock Less is more: Clipbert for video-and-language learning via sparse
  sampling.
\newblock In {\em Proceedings of the IEEE/CVF Conference on Computer Vision and
  Pattern Recognition}, pages 7331--7341, 2021.

\bibitem{li2022language}
Boyi Li, Kilian~Q Weinberger, Serge Belongie, Vladlen Koltun, and Ren{\'e}
  Ranftl.
\newblock Language-driven semantic segmentation.
\newblock {\em arXiv preprint arXiv:2201.03546}, 2022.

\bibitem{li2019analysis}
Hengduo Li, Bharat Singh, Mahyar Najibi, Zuxuan Wu, and Larry~S Davis.
\newblock An analysis of pre-training on object detection.
\newblock {\em arXiv preprint arXiv:1904.05871}, 2019.

\bibitem{lin2014microsoft}
Tsung-Yi Lin, Michael Maire, Serge Belongie, James Hays, Pietro Perona, Deva
  Ramanan, Piotr Doll{\'a}r, and C~Lawrence Zitnick.
\newblock Microsoft coco: Common objects in context.
\newblock In {\em European conference on computer vision}, pages 740--755.
  Springer, 2014.

\bibitem{loshchilov2017decoupled}
Ilya Loshchilov and Frank Hutter.
\newblock Decoupled weight decay regularization.
\newblock {\em arXiv preprint arXiv:1711.05101}, 2017.

\bibitem{luo2021clip4clip}
Huaishao Luo, Lei Ji, Ming Zhong, Yang Chen, Wen Lei, Nan Duan, and Tianrui Li.
\newblock Clip4clip: An empirical study of clip for end to end video clip
  retrieval.
\newblock {\em arXiv preprint arXiv:2104.08860}, 2021.

\bibitem{ma2021simple}
Teli Ma, Shijie Geng, Mengmeng Wang, Jing Shao, Jiasen Lu, Hongsheng Li, Peng
  Gao, and Yu Qiao.
\newblock A simple long-tailed recognition baseline via vision-language model.
\newblock {\em arXiv preprint arXiv:2111.14745}, 2021.

\bibitem{mahajan2018exploring}
Dhruv Mahajan, Ross Girshick, Vignesh Ramanathan, Kaiming He, Manohar Paluri,
  Yixuan Li, Ashwin Bharambe, and Laurens Van Der~Maaten.
\newblock Exploring the limits of weakly supervised pretraining.
\newblock In {\em Proceedings of the European conference on computer vision
  (ECCV)}, pages 181--196, 2018.

\bibitem{ni2022expanding}
Bolin Ni, Houwen Peng, Minghao Chen, Songyang Zhang, Gaofeng Meng, Jianlong Fu,
  Shiming Xiang, and Haibin Ling.
\newblock Expanding language-image pretrained models for general video
  recognition.
\newblock {\em arXiv preprint arXiv:2208.02816}, 2022.

\bibitem{radford2021learning}
Alec Radford, Jong~Wook Kim, Chris Hallacy, Aditya Ramesh, Gabriel Goh,
  Sandhini Agarwal, Girish Sastry, Amanda Askell, Pamela Mishkin, Jack Clark,
  et~al.
\newblock Learning transferable visual models from natural language
  supervision.
\newblock In {\em International Conference on Machine Learning}, pages
  8748--8763. PMLR, 2021.

\bibitem{Rao_2022_CVPR}
Yongming Rao, Wenliang Zhao, Guangyi Chen, Yansong Tang, Zheng Zhu, Guan Huang,
  Jie Zhou, and Jiwen Lu.
\newblock Denseclip: Language-guided dense prediction with context-aware
  prompting.
\newblock In {\em Proceedings of the IEEE/CVF Conference on Computer Vision and
  Pattern Recognition (CVPR)}, pages 18082--18091, June 2022.

\bibitem{selvaraju2017grad}
Ramprasaath~R Selvaraju, Michael Cogswell, Abhishek Das, Ramakrishna Vedantam,
  Devi Parikh, and Dhruv Batra.
\newblock Grad-cam: Visual explanations from deep networks via gradient-based
  localization.
\newblock In {\em Proceedings of the IEEE international conference on computer
  vision}, pages 618--626, 2017.

\bibitem{sharma2018conceptual}
Piyush Sharma, Nan Ding, Sebastian Goodman, and Radu Soricut.
\newblock Conceptual captions: A cleaned, hypernymed, image alt-text dataset
  for automatic image captioning.
\newblock In {\em Proceedings of the 56th Annual Meeting of the Association for
  Computational Linguistics (Volume 1: Long Papers)}, pages 2556--2565, 2018.

\bibitem{touvron2021training}
Hugo Touvron, Matthieu Cord, Matthijs Douze, Francisco Massa, Alexandre
  Sablayrolles, and Herv{\'e} J{\'e}gou.
\newblock Training data-efficient image transformers \& distillation through
  attention.
\newblock In {\em International Conference on Machine Learning}, pages
  10347--10357. PMLR, 2021.

\bibitem{wang2020score}
Haofan Wang, Zifan Wang, Mengnan Du, Fan Yang, Zijian Zhang, Sirui Ding, Piotr
  Mardziel, and Xia Hu.
\newblock Score-cam: Score-weighted visual explanations for convolutional
  neural networks.
\newblock In {\em Proceedings of the IEEE/CVF conference on computer vision and
  pattern recognition workshops}, pages 24--25, 2020.

\bibitem{wang2019camp}
Zihao Wang, Xihui Liu, Hongsheng Li, Lu Sheng, Junjie Yan, Xiaogang Wang, and
  Jing Shao.
\newblock Camp: Cross-modal adaptive message passing for text-image retrieval.
\newblock In {\em Proceedings of the IEEE/CVF International Conference on
  Computer Vision}, pages 5764--5773, 2019.

\bibitem{wang2022cris}
Zhaoqing Wang, Yu Lu, Qiang Li, Xunqiang Tao, Yandong Guo, Mingming Gong, and
  Tongliang Liu.
\newblock Cris: Clip-driven referring image segmentation.
\newblock In {\em CVPR}, pages 11686--11695, 2022.

\bibitem{xu2022groupvit}
Jiarui Xu, Shalini De~Mello, Sifei Liu, Wonmin Byeon, Thomas Breuel, Jan Kautz,
  and Xiaolong Wang.
\newblock Groupvit: Semantic segmentation emerges from text supervision.
\newblock In {\em Proceedings of the IEEE/CVF Conference on Computer Vision and
  Pattern Recognition}, pages 18134--18144, 2022.

\bibitem{xu2021simple}
Mengde Xu, Zheng Zhang, Fangyun Wei, Yutong Lin, Yue Cao, Han Hu, and Xiang
  Bai.
\newblock A simple baseline for zero-shot semantic segmentation with
  pre-trained vision-language model.
\newblock {\em arXiv preprint arXiv:2112.14757}, 2021.

\bibitem{zeiler2014visualizing}
Matthew~D Zeiler and Rob Fergus.
\newblock Visualizing and understanding convolutional networks.
\newblock In {\em European conference on computer vision}, pages 818--833.
  Springer, 2014.

\bibitem{zhai2022lit}
Xiaohua Zhai, Xiao Wang, Basil Mustafa, Andreas Steiner, Daniel Keysers,
  Alexander Kolesnikov, and Lucas Beyer.
\newblock Lit: Zero-shot transfer with locked-image text tuning.
\newblock In {\em Proceedings of the IEEE/CVF Conference on Computer Vision and
  Pattern Recognition}, pages 18123--18133, 2022.

\bibitem{zhong2022regionclip}
Yiwu Zhong, Jianwei Yang, Pengchuan Zhang, Chunyuan Li, Noel Codella,
  Liunian~Harold Li, Luowei Zhou, Xiyang Dai, Lu Yuan, Yin Li, et~al.
\newblock Regionclip: Region-based language-image pretraining.
\newblock In {\em Proceedings of the IEEE/CVF Conference on Computer Vision and
  Pattern Recognition}, pages 16793--16803, 2022.

\bibitem{zhou2016learning}
Bolei Zhou, Aditya Khosla, Agata Lapedriza, Aude Oliva, and Antonio Torralba.
\newblock Learning deep features for discriminative localization.
\newblock In {\em Proceedings of the IEEE conference on computer vision and
  pattern recognition}, pages 2921--2929, 2016.

\bibitem{zhou2022learning}
Kaiyang Zhou, Jingkang Yang, Chen~Change Loy, and Ziwei Liu.
\newblock Learning to prompt for vision-language models.
\newblock {\em International Journal of Computer Vision}, pages 1--12, 2022.

\end{thebibliography}
